\pdfoutput=1
\documentclass[11pt]{article}
\usepackage{acl}

\usepackage[T1]{fontenc}
\usepackage[utf8]{inputenc}
\usepackage{microtype}
\usepackage{newtxtext}
\usepackage{newtxmath}
\usepackage{inconsolata}
\usepackage{CJKutf8}
\usepackage{amsmath}
\usepackage{booktabs}
\usepackage{tabularx}
\usepackage{graphicx}
\usepackage{multirow}
\usepackage{placeins}
\usepackage{listings}
\usepackage{subcaption}
\usepackage{tcolorbox}
\usepackage{siunitx}
\newcommand\equalcontributionfootnote[2]{%
  \begingroup
  \renewcommand\thefootnote{}\footnote{\hspace{-1.5mm}\textsuperscript{#1}~#2}%
  \addtocounter{footnote}{-1}%
  \endgroup
}

\title{%
    Investigating the Representation of Backchannels and Fillers\\
    in Fine-tuned Language Models}

\author{%
    Yu Wang\textsuperscript{1$\lozenge$} \:
    Leyi Lao\textsuperscript{2$\lozenge$} \:
    Langchu Huang\textsuperscript{2}\\
    \textbf{Gabriel Skantze\textsuperscript{3}} \:
    \textbf{Yang Xu\textsuperscript{2$\blacklozenge$}} \:
    \textbf{Hendrik Buschmeier\textsuperscript{1$\blacklozenge$}}\\
    \textsuperscript{1} Bielefeld University, Bielefeld, Germany\\
    \textsuperscript{2} Southern University of Science and Technology, Shenzhen, China\\
    \textsuperscript{3} KTH Royal Institute of Technology, Stockholm, Sweden
}

\begin{document}
\maketitle

\begin{abstract}
   Backchannels and fillers are important linguistic expressions in dialogue, but often treated as ‘noise’ to be bypassed in modern transformer-based language models (LMs). Here, we study how they are represented in LMs using three fine-tuning strategies on three dialogue corpora in English and Japanese, in which backchannels and fillers are both preserved and annotated. This allows us to investigate how fine-tuning can help LMs learn these representations. We first apply clustering analysis to the learnt representation of backchannels and fillers, and find increased silhouette scores in representations from fine-tuned models, which suggests that fine-tuning enables LMs to distinguish the nuanced semantic variation in different backchannel and filler use. We also employ natural language generation metrics and qualitative analyses to verify that utterances produced by fine-tuned LMs resemble those produced by humans more closely. Our findings suggest the potential for transforming general LMs into conversational LMs that can produce human-like language more adequately.
\end{abstract}

\section{Introduction}
\label{sec:introduction}%
\equalcontributionfootnote{$\lozenge/\blacklozenge$}{Equal contribution.}%
In everyday conversations, such as the one below from the Switchboard corpus \citep[3508:7--12]{LDC-swbd1r2},
backchannels (e.g., \emph{uh-huh}) and fillers (e.g., \emph{uh}) are ubiquitous: 
\begin{description}
    \itemsep-.5em 
    \item[A:] Well, we have saved our newspapers for years and years --
    \item[B:] \textit{Uh-huh}
    \item[A:] -- because the, \textit{uh}, Boy Scouts --
    \item[B:] \textit{Uh-huh}
    \item[A:] -- our boys have been involved in have, \textit{uh}, had a huge recycling bin …
\end{description}
They play an important role in managing the flow of conversation and are an important way for interlocutors to negotiate the common ground \citep{clark1996using}.
Since backchannels/fillers usually do not convey information directly and have only pragmatic functions (e.g., expressing affirmation or disagreement to the previous utterance; \citealp{juckeret1998discourse}), they are considered semantically ‘bleached’ elements \citep{FULLER200323} and deemed optional due to their lack concrete meanings \citep{SCHOURUP1999227}.
As a result, the NLP community often treats them similarly to stop words, excluding them during pre-processing as a method to “clean” the data and improve accuracy \citep[e.g., see][]{Sarica2021}.
In dependency parsing for spoken dialogue data, for example, studies report that excluding backchannels/fillers in the Switchboard corpus \citep{LDC-swbd1r2} can significantly improve parsing accuracy such as \citep{charniak-johnson-2001-edit, jorgensen-2007-effects, dobrovoljc-martinc-2018-er}.

However, previous studies have highlighted that backchannels and fillers can contain rich contextual meaning in dialogue: backchannels such as \emph{yeah} or \emph{okay}, for example, serve as feedback in response to preceding speech \citep[pp.~32]{clark1996using}.
Fillers, such as \emph{uh} and \emph{um}, are used as signals of disfluency \citep{rose2015and} and reflect the speaker's cognitive processing when searching for the next word \citep{CLARK200273}, thus indicating the speaker's cognitive load (e.g., see Table~1 of \citealp[]{berthold1999} and \citealp[]{rose2015and}). Fillers can thus be regarded as an important signal to the listener that the speaker needs some time to complete the utterance (and wants to hold the turn; \citealp{ball1975listeners}).
Further studies using qualitative approaches show that backchannels, as a source of feedback, play an important role in the incremental updating of dialogue from a semantic perspective, controlling the flow of information during conversation \citep{bergey2024umyeahproducingpredicting} and improving the mutual understanding between the interlocutors in order to reach a joint goal \citep{HowesEshghi2021}. 

In order to figure out the potential of transforming language models (LMs) into conversational LMs, which can utilise backchannels and fillers, it is essential to establish how effectively they can be learnt and represented.
An obstacle for answering this question is that language data used in NLP is often text-based \cite{liesenfeld-dingemanse-2022-building, dingemanse-liesenfeld-2022-text} and that well-annotated conversation data is often insufficient in size and quality. Consequently, there is limited contextual information on backchannels/fillers for LMs to learn from. This might be one reasons why LM-based automatic speech recognition (ASR) systems perform poorly in recognising turn taking/holding, a practice which is largely moderated by backchannels and fillers. Moreover, due to the absence of backchannels and fillers during the pre-training phase, most LMs have limited knowledge of these linguistic units. Consequently, the content generated by these LMs tends to be text-like and differs in form from natural dialogue\footnote{%
    In our opinion, this case is also applicable to LLM such as LLaMA-3-8B, as evidenced in our observation shown in the examples dialogues 1 to 4 in Figure~\ref{fig:nlg_examples} in the Appendix, during the NLG task, the pre-trained LLM barely generates any backchannels and fillers, which should be considered as an important sign that backchannels/fillers are not well represented in the pre-trained language model.
}, a reason, why NLP failed to “[put] natural in natural language processing” \citep{chrupala-2023-putting}. Consequently LMs lack the ability to act as competent dialogue agents that can, for example, produce (i) appropriate backchannels as feedback to user utterances, and are therefore not considered attentive enough to meet human user expectations \citep{buschmeier2018communicative}, and (ii) natural fillers in their utterances, which are vital for their role in organising speech and their communicative functions in spoken language understanding \citep{dinkar-etal-2022-fillers}. As a result, an LM which ignores backchannels and fillers does not clean the data, it strips the model of the social cues necessary for natural and fluid interaction \citep{EdlundGustafson2008}, and consequently cannot be used for building a competent conversational agent.

In this study, we therefore address \textbf{backchannels/fillers as important linguistic phenomena} that are not well reflected in language models and specifically investigate the issue of learning the representations of backchannels and fillers in LMs. Fine-tuning is an approach that has proven effective in tackling the issue of dialogue phenomena in LMs \citep[see, e.g.,][]{noble-maraev-2021-large}.
Although there has been a great deal of work attempting to answer the question of how the representation of linguistic knowledge in language models, such as BERT \citep{devlin-etal-2019-bert} and GPT2 \cite{radford2019language}, is altered after fine-tuning, we believe that work makes a meaningful contribution, as an initial study trying to answer how and how well LMs can represent backchannels and fillers after fine-tuning. Our general research question (\textbf{RQ1}) and three more specific research questions are:

\begin{description}

    \item[RQ1:] 
        Does fine-tuning improve the representations of backchannels and fillers in modern transformer-based language models, such as BERT, GPT-2 and a larger language model such as LLaMA-3 8B, Qwen-3 8B?
    
    \item[RQ2:] What role does contextual information play when we try to obtain the representations of backchannels/fillers from LMs?

    \item[RQ3:] Which of the studied LMs can benefit more from fine-tuning?
    
    \item[RQ4:] Do various fine-tuning strategies make a difference in learning the representation of backchannels/fillers?
    
\end{description}

\section{Related Work}
\label{sec:related-work}

\paragraph{Backchannels/Fillers as Discourse Markers:}

Backchannels/fillers are considered to be discourse marker \cite{juckeret1998discourse}. They are semantically ‘bleached’ elements in conversation, such as \emph{oh, yeah, uh, uhm} which can be either fillers or backchannels -- depending on whether they are within the dialogue as a sign of disfluency, or stand alone or at the beginning of an utterance, which are then usually taken as feedback to the previous utterance.

As indicated by \citet{FULLER200323}, \citet{fox2010discourse} and \citet{skantze2021turn}, backchannels/fillers have the following two properties: first of all, they indicate the turn relations of utterances and thus play a role in conversation management, e.g., \textit{uh} and \textit{uhm} in speech, as a signal of disfluency, can indicate turn-holding and pause; secondly, they are `optional', i.e., deleting them from the utterance won't change its truth conditional meaning. Although fillers and backchannels are considered semantically bleached elements in dialogue, they are considered important for managing the flow of the dialogue and play an important role in `grounding' processes: reflecting the attentiveness of the listeners during the interaction, confirming listeners' understanding state as well as establishing common ground \citep[]{clark1996using, buschmeier2018communicative}.
Moreover, backchannels/fillers are linguistic universals, as shown in the survey by \citet{dingemanse-liesenfeld-2022-text}, they tend to be the most frequent expressions in spoken language distribution.

Given the rich roles backchannels/fillers play in dialogue, some studies look into the prediction and generation of backchannels/fillers. For example, 
\citet{skantze-2017-towards} reports that an LSTM-based model can predict the occurrence of backchannels in dialogue.
\citet{ruede2017enhancing} show that using word embeddings in a speech model can improve the accuracy of backchannel detection. \citet{10191640} proposes a transformer-based pipeline to predict backchannels and further use the predicted backchannels as an index of the agreement among interlocutors. \citet{wang-etal-2022-evaluating} build three language models to generate fillers in clean speech and evaluate how this helps to improve the naturalness of the generated speech.

\paragraph{Fine-tuning and Representation:}

Fine-tuning is an important step to adapt a pre-trained model to novel downstream tasks and learn representations that are important for downstream tasks.
It has been consistently reported that fine-tuning can improve LMs' representation capabilities at different levels of linguistic representation. \citet{mosbach-etal-2020-interplay-fine}, for example, use three sentence level classification tasks selected from the GLUE benchmark \citep{wang-etal-2018-glue} as the fine-tuning tasks on BERT \cite{devlin-etal-2019-bert}, and find that fine-tuning can indeed affect the representation of linguistic knowledge in language models, especially the last hidden layers.
It has been approved in many previous studies that fine-tuning is indispensable for an LM to perform well in different downstream tasks (e.g., \citealp[]{noble-maraev-2021-large, merchant-etal-2020-happens}).
A classical way to evaluate a fine-tuned LM is to use probing techniques to investigate the meaning representation of the hidden layer weights in the fine-tuned model, which can either be a supervised method (\emph{building a classifier to report accuracy}), or an unsupervised method (\emph{using clustering to report clustering quality before and after fine-tuning}) \citep[see, e.g.,][]{zhou-srikumar-2021-directprobe, mosbach-etal-2020-interplay-fine}.
However, what is unknown is how language inputs that are less represented in the original language model would be represented in the fine-tuned language model\footnote{%
    We observed that backchannels and fillers have high token IDs, which indicates that they are not included in the original vocabulary of the tokenizer, or the tokenizer encountered the backchannels and fillers only in a late phase of pre-training.
}.
We will try to answer this in our case study of backchannels/fillers, mainly through the use of clustering after the fine-tuning process.

\section{Methodology}
\label{sec:methodology}

\subsection{Data Selection}

In order to find datasets suitable for this study, we focussed our search specifically on datasets of transcribed spoken dialogue where fillers and backchannels are properly annotated. In order to lower the complexity of the task, the number of interlocutors per dialogue were be limited to two.
We identified the following three datasets to meet our needs: \textbf{Switchboard} \citep{LDC-swbd1r2} and \textbf{MapTask} \citep{anderson1991hcrc}, both English; and the \textbf{BTSJ} 1000 Person Japanese Natural Conversation Corpus \citep{usami2023}, which is in Japanese. 
Combined, the two English datasets are the same size as the Japanese dataset (about \num{150000} utterances).

We selected backchannels/fillers based on \citet{todd2019has} and \citet{pilan-etal-2024-conversational}, which include data of different fillers and backchannels for dialogical interaction in English and Japanese. We report details of the selected backchannels/fillers in Appendix~\ref{sec:data}.

\subsection{Task definition}

In order to learn the representation of backchannels/fillers in language models, we chose to fine-tune existing language models that do not have (or have limited) knowledge of backchannels/fillers. For fine-tuning we use downstream tasks where the models can learn the contextual information of these linguistic items. We use and compare three different fine-tuning tasks: masking (\textbf{MASK}), next token prediction (\textbf{NTP}) and turn taking prediction (\textbf{TTP}). These are described in the following.

\begin{figure*}
    \centering
    \includegraphics{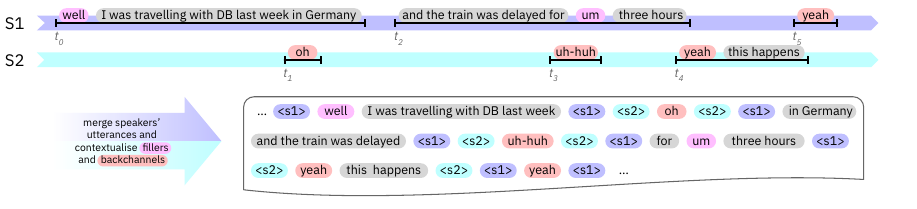}
    \caption{A dialogue example which shows turn-taking and contains various backchannels/fillers. The \textit{um} from the speaker S1 is a \textbf{filler} while the \textit{oh}, \textit{uh-huh}, and \textit{yeah} from the speaker S2 are \textbf{backchannels}. For the \textbf{NTP} and \textbf{Masking} fine-tuning tasks we merge the utterances from both speakers. For this, we take the utterances from the both and combined them into one larger sequence, considering that the utterances from one speaker is dependent on the utterances from the other speaker. We retain speaker information by adding speaker IDs (e.g., <s1>) to let the LM know that the utterances are from two different sources. Merging is not required for the \textbf{TTP} task. \textbf{TurnGPT} takes the utterances from both speakers in a linear-time order as input and predicts turn-taking probabilities.}
    \label{fig:example-dialogue}
\end{figure*}

\begin{figure}[ht]
    \centering
    \includegraphics[width=1\linewidth]{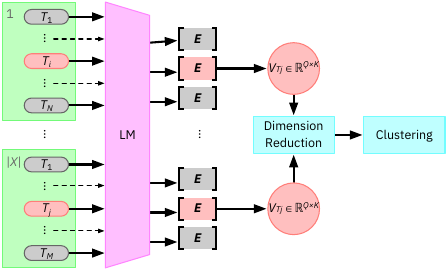}
    \caption{%
        We select the backchannel \textit{uh-huh} as an example to show how embeddings are obtained from language models. The pipeline consists of the following three steps: (1) taking a corpus of utterances of various lengths, which contain the backchannel \textit{uh-huh} in, e.g., position $T_i, T_j$, (in total $|X|$ samples); (2) encoding the text through fine-tuning to get the contextual vector representation of the backchannel \textit{uh-huh} ($Q$ is the length of the backchannel tokens, $K$ is the dimension of the selected hidden state); (3) reducing the dimension of the contextual vector and build clusters for the contextual vector of the backchannel \textit{uh-huh}.}
    \label{fig:pip_uh}
\end{figure}

\paragraph{Masking (MASK)}

Masked language modelling encourages models to utilize bidirectional context to build contextualized token representations. Inspired by the pre-training objective of BERT \citep{devlin-etal-2019-bert}\footnote{%
    Although BERT is not an autoregressive model like GPT-2 or LLaMA, it is still considered a language model.
}, we adapt this method to the conversational setting by selectively masking backchannels/fillers.

Let $\mathbf{X} = \left(x_1, x_2, \ldots, x_T\right)$ be a tokenized sequence of utterances drawn from a dialogue, where $x_t$ is the $t^\mathrm{th}$ token and $T$ denotes the sequence length. We first identify all matching spans $\mathbf{S} = \{s_1, s_2, \ldots, s_N\}$, where each $s_n$ is a continuous token subsequence of backchannels or fillers, and $N$ is the total number of such spans. For each span $s_n \in \mathbf{S}$, we define $L = \text{len}(s_n)$ and randomly apply a masking operation $\mathbf{M}$ as follows:
\begin{equation*}
    M(s_n) =
        \begin{cases}
            \text{[MASK]}^{L} & (P=0.8)\\
            \text{RandomTokens}^{L} & (P=0.1)\\
            s_n & (P=0.1)
        \end{cases}
\end{equation*}
where $P$ denotes the probability of each operation (we use the default values of \citealt{devlin-etal-2019-bert}). The resulting corrupted sequence $X'$ is then encoded into contextualized representations $H=(h_1, h_2, \ldots, h_T)$. For each masked span $s_n \in {S}$, the model predicts the original tokens through ($W$ being the classification layer weights):
\begin{equation*}
\begin{aligned}
    P(y_{s_{n,k}} \mid X') 
    &= \text{softmax}(W h_{s_{n,k}} + b), \\
    &\forall k \in \{1,\dots,L\}
\end{aligned}
\end{equation*}
We use BERT models for \href{https://huggingface.co/google-bert/bert-base-cased}{English} and \href{https://huggingface.co/tohoku-nlp/bert-base-japanese}{Japanese} from HuggingFace library.

By learning to predict masked backchannels or fillers from contextual discourse, the model can build a better representation.
As shown in Figure~\ref{fig:example-dialogue}, in order to reflect the notion that backchannels and fillers are no less different from words with substantial meaning that are conditioned by their previous contexts, in our experiment practice we merge the utterances from the two sources of speakers.
One of the drawbacks of doing so is that we will miss the information of the speaker of an utterance. As a solution for the fine-tuning task input, we add two special tokens to indicate to the LM the source speakers (<s1> and <s2>, as illustrated in Figure~\ref{fig:example-dialogue}). This setting is also applied to the fine-tuning strategy: next token prediction.

\paragraph{Next-token Prediction (NTP)}

We consider the general task as a language modelling task, i.e., estimating the probability of the next token (backchannel/filler or other word) given previous input.

Let $\mathbf{X} = \left( x_1, x_2, \ldots, x_T \right)$ represent a sequence of tokens, where $x_t$ is the token at time step $t$ (either a filler/backchannel or a regular word with substantial meaning) and $T$ is the length of the sequence. 
A pre-trained language model $f(\cdot)$ is fine-tuned to estimate the probability of the next token $x_{t+1}$ given all previous tokens $x_1, x_2, \dots, x_t$. 
Thus the probability of the next token $x_{t+1}$ given the previous tokens $x_1, x_2, \dots, x_t$ is expressed as:
\begin{equation*}
    P \left( x_{t + 1} \mid x_1, x_2, \dots, x_t \right) = f  \left( x_1, x_2, \dots, x_t; \theta \right),
\end{equation*}
where $\theta$ are the model parameters that will be adjusted during the fine-tuning process.

\begin{figure}
    \centering
    \includegraphics[width=0.5\textwidth]{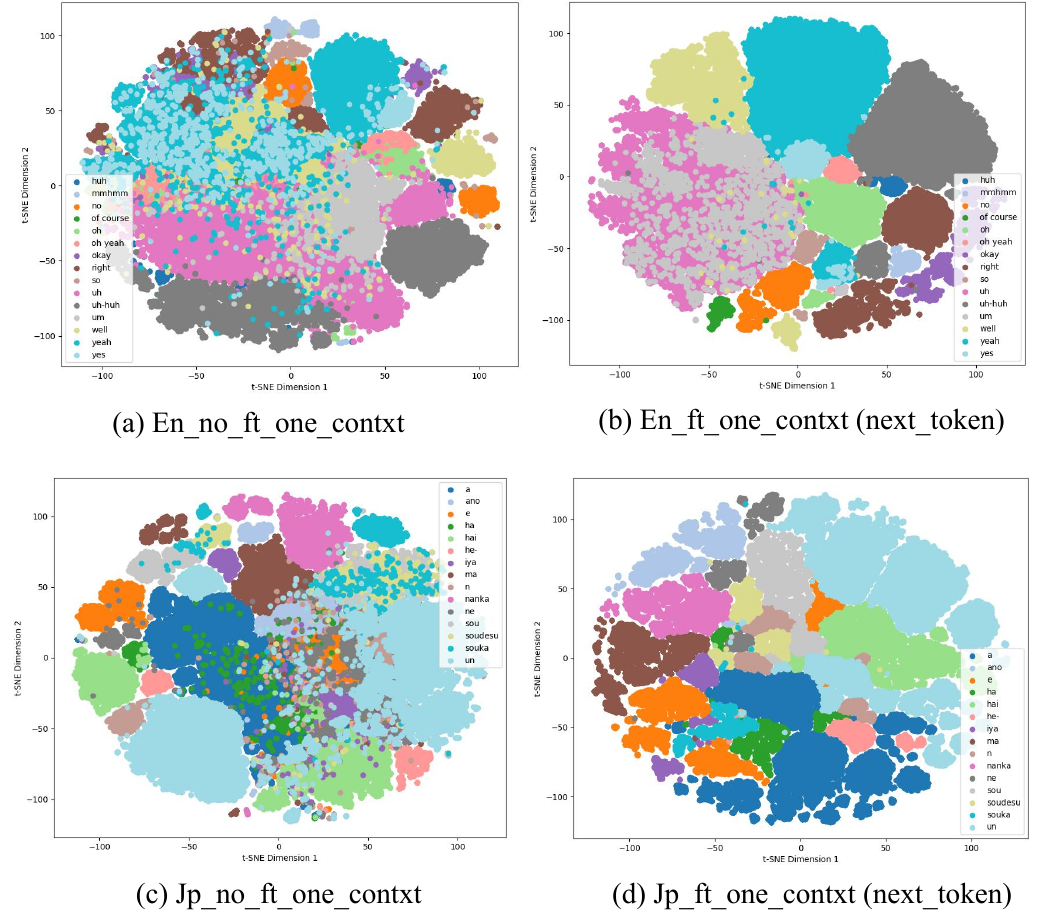}
    \caption{t-SNE plots of the English and Japanese backchannel/filler embeddings from the LLaMA-3 model (a, c) before fine-tuning, and (b, d) after \textbf{NTP} fine-tuning. Setting: one-context.}
    \label{fig:tsne_result}
\end{figure}

For this fine-tuning task, we use the Japanese \cite{rinna-japanese-gpt2-medium, sawada2024release} and English \cite{radford2019language} GPT-2 models, as well as the multilingual LLaMA-3 8B \cite{llama3modelcard} and Qwen-3 8B \cite{yang2025qwen3} models.

\paragraph{Turn-taking Prediction (TTP)}

Our third fine-tuning task is turn-taking prediction. The task is chosen due to the close relation between the use of backchannels/fillers and turn-taking/turn-holding.
Specifically, we use the framework TurnGPT \citep{ekstedt-skantze-2020-turngpt}, a language model based on GPT-2 and designed for the prediction of turn-shifts in spoken dialogue. A formal definition of the turn-taking prediction task is as follows.

Let $\mathbf{X} = \left( x_1, x_2, \dots, x_T \right)$ denote a sequence of tokens consisting of the linear ordering of utterances from both interlocutors (e.g., most tokens are from \textbf{interlocutor A} while \textbf{interlocutor B} produces backchannels/fillers, where $x_t$ is the $t^\text{th}$ token, and $T$ is the sequence length.
The fine-tuning task is then to estimate the probability distribution $P \left( y^* \mid X \right)$, where $y \in \{1, 2, \dots, T\}$ indicates the likelihood of a turn-taking event occurring after token $x_y$. The final predicted turn-taking location is then based on:

\begin{equation*}
    y^* = \arg\max_{y} P\left( y \mid \mathbf{X} \right)
\end{equation*}
Similar to the \textbf{NTP} task, training TurnGPT also requires speaker ID (e.g., <s1>) for each utterance in order to reflect turn shifts.
The details on our data pre-processing are summarised in Appendix~\ref{sec:dataprop}.

\begin{figure*}
    \centering
    \includegraphics[width=1\linewidth]{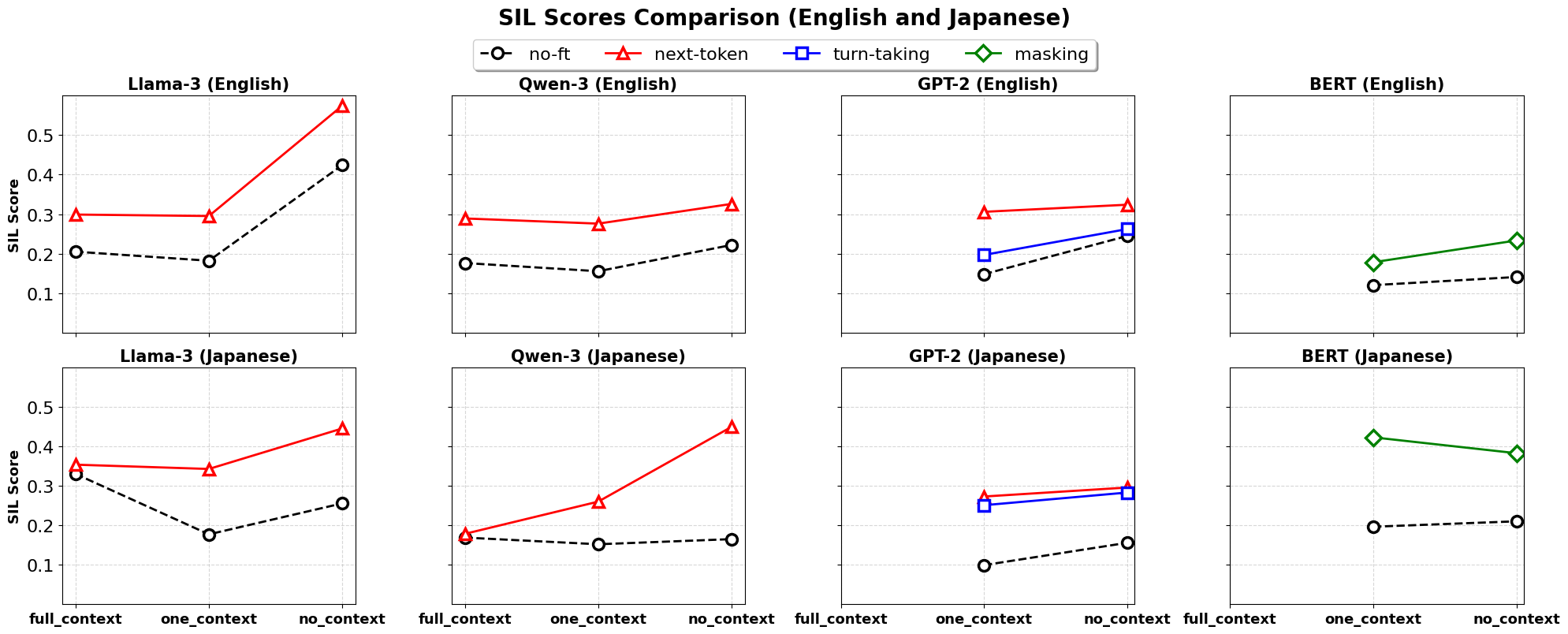}
    \caption{Average silhouette (SIL) scores of backchannels/fillers before and after masking, and two other fine-tuning methods. \textbf{NTP} (red line) is applied to GPT-2 based models, Qwen-3 8B and LLaMA-3 8B, while \textbf{TTP} (blue line) is only used for GPT-2 based models. For BERT, fine-tuning is performed using the \textbf{MASK} (green line) approach. The full-context setting is not available for the GPT-2 and BERT based models given the model's input size limitations. Dashed lines are results without fine-tuning (“no-ft”), solid lines are those obtained after fine-tuning. The robustness test on the results are summarised in Table~\ref{tab:bootstrap_results} in the Appendix.}
    \label{fig:silhouette-scores}
    \vspace{4mm}
\end{figure*}

\subsection{Experimental Set-Up}
\label{sec:methods}

Our experiment workflow is shown in Figure~\ref{fig:pip_uh}.
The chosen LMs are fine-tuned in advance based on the three different fine-tuning tasks\footnote{%
    Note that for our analysis, when we extract the embeddings of backchannels/fillers for the pre-trained LMs without fine-tuning, the pipeline works in the same way.}.
For the NTP and MASK tasks, we selected 80\% of the conversation data for fine-tuning and the remaining 20\% for subsequent generative evaluation. Conversely, for the TTP task, all datasets were partitioned into training, validation, and testing sets exclusively for the fine-tuning process. For the evaluation of the representations of backchannels/fillers, we used all the conversation data. 

To illustrate the workflow, we use the backchannel \textit{uh-huh} as an example. In the initial stage, we pass through all input samples $|\chi|$ and filter out all utterances that contain \textit{uh-huh}. We add special token \textit{<ds>} to mark and represent the backchannel (the backchannel \textit{uh-huh} is transcribed as \textit{<ds> uh-huh <ds>}).
We then encode the corresponding utterances through the different hidden layers of a fine-tuned LM, obtaining the representation vector of \textit{uh-huh} from the final hidden layer.
Next, we check the dimensions of the representation matrix. Given that some of the backchannels/fillers can consist of more than one token (as \textit{uh-huh} in our example), we then apply dimension reduction, simply taking the weighted average value of one dimension of the matrix so that the matrix can be levelled down to a vector representation (embedding). 
With the embeddings of the backchannels/fillers \textit{`uh-huh'}, we examine its representation via clustering. 
For BERT and GPT-2 models used in the fine-tuning task, the dimension of the hidden layer is 768 for English GPT-2 model and 1024 for the Japanese GPT-2 model. TurnGPT is based on the GPT-2, thus shares the same dimension.
A special case is LLaMA-3 8B and Qwen-3 8B as their hidden layer dimension is 4096. For computational reasons, we reduce the dimensionality of the obtained embeddings from all of the models to 100 using Principal Component Analysis \cite[PCA;][]{abdi2010principal}.
See Appendix~\ref{sec:TD} for technical details of the fine-tuning (e.g., GPU run time, use of LoRA for parameter optimisation when fine-tuning the LLaMA model \citealp[]{hu2022lora}).

Moreover, for extracting the embeddings of backchannels/fillers after fine-tuning, we have three different context settings: 
    (i) no context information; 
    (ii) one context information; 
    (iii) full context information. 
In the first setting, when we encounter a backchannel/filler in an utterance, only the utterance containing that backchannel/filler is fed to obtain its embedding. 
In the second setting, we use the previous and subsequent utterances of a backchannel/filler to build the context before obtaining its embedding. 
In the third setting, we combine all previous utterances when we encounter an utterance containing a backchannel/filler, using this combination as the input to obtain the embedding of that backchannel/filler. This setting is only applicable to the LLaMA-3 8B and Qwen-3 8B models since both have no input length limitation.

\section{Analysis and Results}
\label{sec:res}

\paragraph{Overall observation}

We chose the top 15 most frequent backchannels/fillers in our Japanese and English data to check how the representation of backchannels/fillers changes after fine-tuning.
Among the 15 selected backchannels/fillers in each language, typical examples are: the ones indicating positive or negative feedback as the reference objects, and signals for turn-holding (in English, \textit{yes}, \textit{uh}, \textit{yeah}, etc.;
in Japanese, examples are \begin{CJK}{UTF8}{min}`はい' (hai, `yes'), `うん' (un, `yeah'), `ああ' (aa, `ah')\end{CJK}, etc.).

We report the preliminary results using t-SNE visualizations \citep{van2008visualizing}. 
The embeddings of the 15 selected backchannels/fillers in each language are the input data. These embeddings are extracted from the last hidden layer of the LMs, before and after fine-tuning, as shown in Figure~\ref{fig:pip_uh}. 
We have selected t-SNE visualisations from the LLaMA-3 model with a single context setting to demonstrate how fine-tuning (in this instance, NTP) modifies the representation of backchannels/fillers within the language models (LMs).
Figure~\ref{fig:tsne_result}a shows that for English data, when we obtain the embeddings of a selected backchannel/filler from the pre-trained LLaMA-3 model, the distinction between different backchannels/fillers are not clear enough. We believe that the large overlap of different data points is due to the fact that the pre-trained LMs have limited knowledge of the backchannel/filler and thus will assign random values to the encountered backchannel/filler.
After fine-tuning, a clearer distinction among the embeddings of different backchannels/fillers starts to appear in the English data (see Figure~\ref{fig:tsne_result}b). The effect of fine-tuning for Japanese data is similar. After fine-tuning distinctions between different backchannels/fillers emerge as clearer boundaries among different colours appear in the t-SNE visualization (Figure~\ref{fig:tsne_result}c and ~\ref{fig:tsne_result}d).
The t-SNE visualisations from other LMs and different settings can be found in Appendix~\ref{app:additional-tSNE} (Figures~\ref{fig:tsne_result1} to~\ref{fig:tsne_result10}). 

\begin{figure*}[ht]
    \centering
    \begin{subfigure}[t]{0.40\linewidth}
        \centering
        \includegraphics[width=\linewidth]{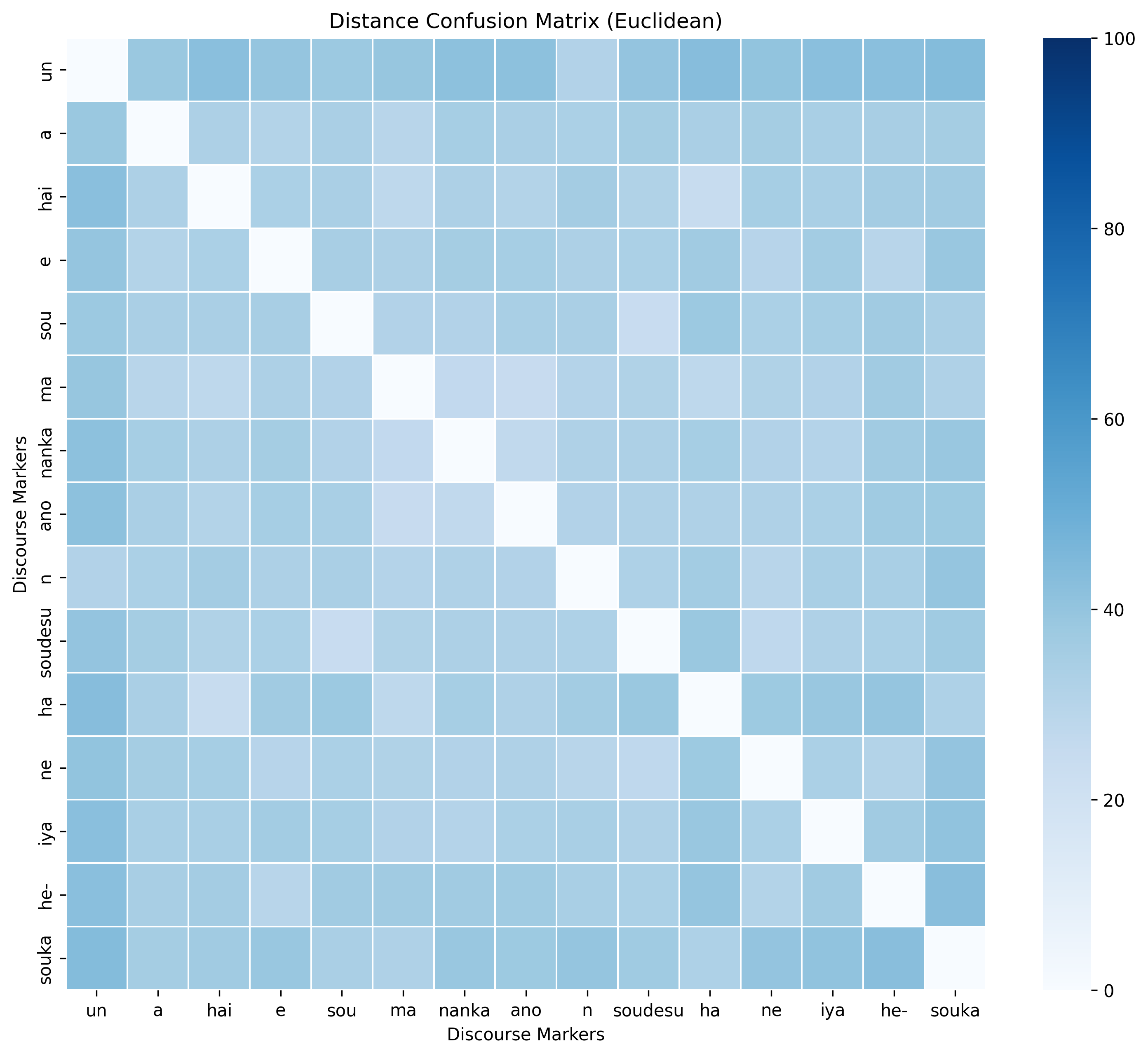}
        \caption{no\_ft, one\_context}
        \label{fig:mtja2a}
    \end{subfigure}
    \hspace{2cm}
    \begin{subfigure}[t]{0.40\linewidth}
        \centering
        \includegraphics[width=\linewidth]{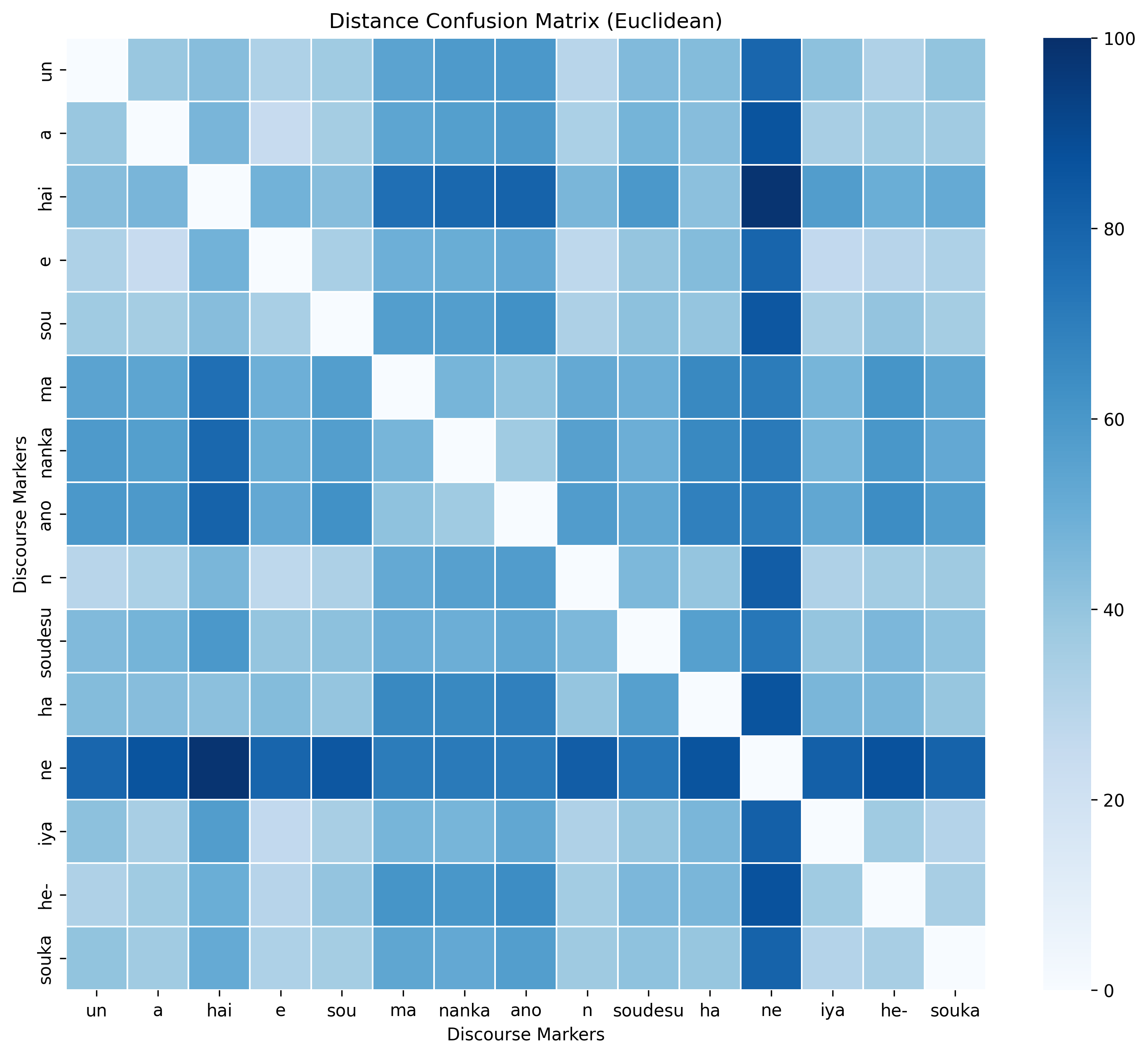}
        \caption{ft, one\_context (NTP)}
        \label{fig:mtja2b}
    \end{subfigure}
    \caption{Distance matrices for the top 15 Japanese backchannels/fillers in the Qwen-3 model (a) before and (b) after fine-tuning.}
    \label{fig:mtja2}
\end{figure*}

\paragraph{Analysis Using K-Means Clustering}
As a further analysis, we look at all the corresponding embeddings and apply k-means clustering (from \href{https://scikit-learn.org/stable/modules/generated/sklearn.cluster.KMeans.html}{scikit-learn}) individually for backchannels/fillers in the English and Japanese data.
This analysis is motivated by the fact that backchannels/fillers, as functional words, often have different pragmatic functions (e.g., indicating agreement, hesitation, etc.; \citealt{petukhova2009towards}). A similar idea was discussed in annotation work, which revealed that the same backchannel/filler can serve several different communicative functions \citep[fig.~4]{figueroa-etal-2022-annotation}.
In terms of our study targets, for example, the Japanese backchannel/filler \begin{CJK}{UTF8}{min}`うん' (un, `yeah')\end{CJK} can indicate both `confirmation' and `hesitation'.
When LMs have limited knowledge of backchannels/fillers and are asked to generate their representation, models will usually give random vector values to them, which will lead to two possible clustering results: either a large or a small $k$ value.
Therefore, if fine-tuning can improve the representation learning of backchannels/fillers, we should expect that the clustering effect of their embeddings will be more salient after fine-tuning. That is to say, in general we should see an increase in $k$ value, which can indicate that fine-tuned LM's meaning representation can reflect different pragmatic functions of backchannels/fillers if $k$ was small; in contrast, if the $k$ value is initially large, it should be smaller after fine-tuning.

We first look at how our proposed approaches can distinguish the meaning representations of different backchannels/fillers. We perform $k$-means clustering and select the associated centroids (embedding values) of the backchannels/fillers before and after fine-tuning (\textbf{NTP} and \textbf{MASK}). The centroids are later averaged to produce a single representative embedding for each backchannel/filler. We select the top-15 most frequent backchannels/fillers and use confusion matrices (based on Euclidean distance) to capture the difference before and after fine-tuning.

Figure~\ref{fig:mtja2} shows a representative confusion matrix, where Qwen-8B is fine-tuned on Japanese dialogue data under the one-context setting (see Appendix~\ref{app:conmatric} for the other confusion matrices). In the confusion matrix, darker colours indicate greater distances and lighter colours indicate
greater similarity. As can be seen, the inter-difference among the different backchannels/filler is more salient after fine-tuning. For example, the embeddings of the backchannel \begin{CJK}{UTF8}{min}`ね' (ne, `right?', `hmm', etc.)\end{CJK} are separated from the embeddings of \begin{CJK}{UTF8}{min}`はい' (hai, `yes', `mm-hmm', etc.)\end{CJK}

To quantitatively analyse the quality of the clustering of the embeddings of backchannels/fillers before and after fine-tuning, we use silhouette scores as a measure \cite{rousseeuw1987silhouettes}. Given a range of $k$-values, used for $k$-means clustering, we calculate the corresponding silhouette coefficient $\textbf{s}(i)$ for a data point $i$ (a backchannel/filler):
\begin{equation*}
    \mathbf{s}(i) = \frac{b(i)-a(i)}{\max(a(i), b(i))},
\end{equation*}
where $a(i)$ is the average Euclidean distance between the embedding of the backchannel/filler $i$ and all other embeddings in the same cluster as $i$. $b(i)$ is the minimum average Euclidean distance from embedding $i$ to all embeddings in any other cluster, $-1 \leq \mathbf{s}(i) \leq 1$. 
The \textbf{silhouette score} (\textbf{SC}) for a clustering is then measured as the average silhouette coefficient over all $n$ embeddings:
\begin{equation*}
    \textbf{SC} = \frac{1}{n} \sum_{i=1}^{n} \mathbf{s}(i),
\end{equation*}
the higher the value, the better the clustering quality.

The final result is summarised in Figure~\ref{fig:silhouette-scores}. In all of the four selected LMs, a general tendency we can observe is that with different fine-tuning strategies, the average silhouette score increases.
Detailed statistics can be found in in Tables~\ref{tab:en_sil1} to \ref{tab:jap_sil4} in Appendix~\ref{app:silhouette}.

\begin{figure*}
    \centering
    \includegraphics[width=1\linewidth]{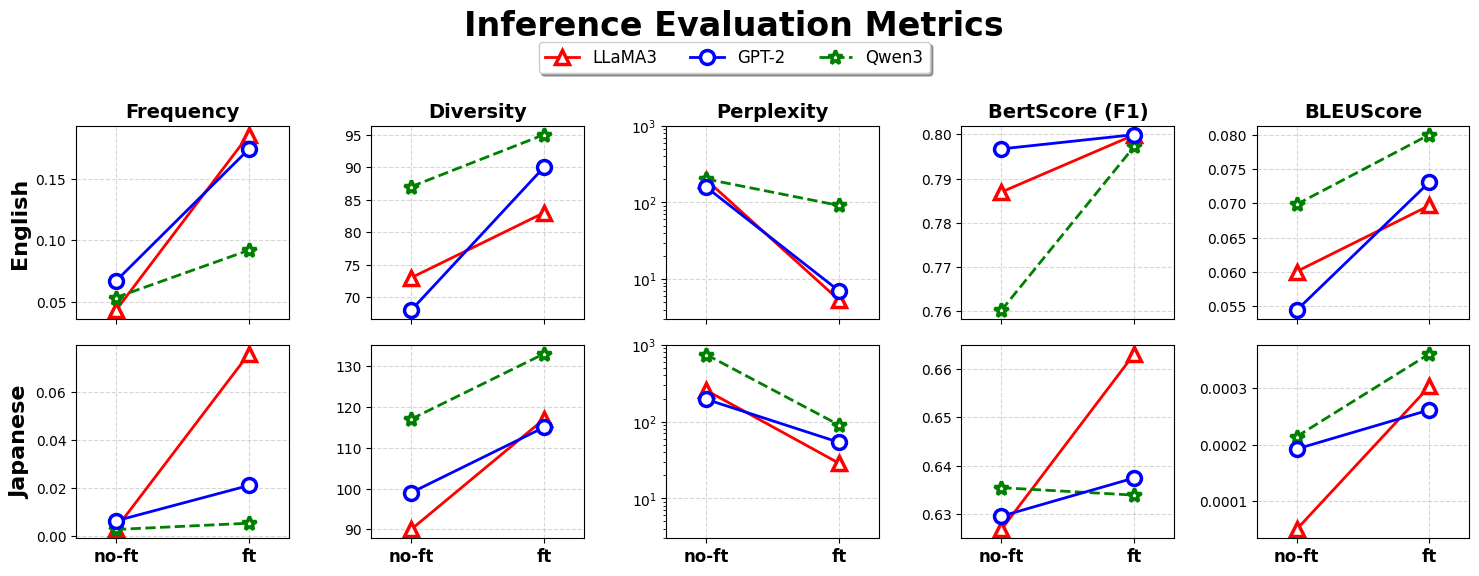}
    \caption{Evaluation metrics of generated backchannels/fillers in the English and Japanese NLG tasks. The models are evaluated before and after fine-tuning with respect to frequency, diversity, frequency-weighted perplexity, BERTScore (F1), and BLEUScore (details such as metric definition and performance results are in Section~\ref{app:supp-nlg}, Tables~\ref{tab:nlg1} and \ref{tab:nlg2} in the Appendix.).}
    \label{fig:nlg_res}
\end{figure*}

\section{Discussion}
\label{sec:discuss}

In Section~\ref{sec:res}, the t-SNE visualization (Figure~\ref{fig:tsne_result}) reveals that fine-tuning LMs will change the representation of backchannels/fillers in a positive way.
The analysis using $k$-means clustering and silhouette scores further shows that fine-tuning has beneficial effects on the representation of backchannels/fillers. Here, we further discuss these results in light of our research questions. 

\paragraph{RQ1: Can the representations for backchannels/fillers in modern language models be learned/improved through fine-tuning?}

From Figure~\ref{fig:silhouette-scores} (as well as Tables~\ref{tab:en_sil1} to \ref{tab:jap_sil4} in the Appendix), we can see that fine-tuning the pre-trained LMs with dialogue data that contains contextual information on backchannels/fillers leads to a general increase of silhouette scores, which indicates a more salient clustering effect after fine-tuning and serves as crucial evidence that LMs can learn
the representation of backchannels/fillers with the selected fine-tuning strategies. At the same time Figure~\ref{fig:k_value_change} (Appendix) shows that the numbers of clusters $k$ become more salient: slightly increasing when $k$ was small and decreasing when $k$ was large.

\paragraph{RQ2: Role of contextual information}

For both fine-tuned and not fine-tuned LMs, The average silhouette score generally exhibits an inverse relationship with the context size used for embeddings (as shown in Figure~\ref{fig:silhouette-scores}, exceptions are LLaMA-3 and BERT for Japanese under the fine-tuning setting, and Qwen-3 without fine-tuning). We interpret this result as follows: When context information is added, LMs tend to treat backchannels/fillers as functional words, whose representations would be compressed into more limited regions of the semantic space. The increased context size encodes more information from the surrounding content words to the backchannels/fillers, which ``dilutes'' their representations to the average level.

\paragraph{RQ3: Differences among the selected language models}

In this paper, we targeted four types of language models: BERT and GPT-2 are smaller in size (parameters) and monolingual; LLaMA-3 8B and Qwen-3 8B are much larger and multilingual. It turns out (Figure~\ref{fig:silhouette-scores}) that the average silhouette scores of clustering from the LLaMA-3 embeddings are best among the four selected LMs. Model size, however, does not necessarily correlate with better representation in terms of silhouette scores, e.g., the results from the fine-tuned BERT model (smallest in size) are comparable to those of the Qwen-3 model.

\paragraph{RQ4: Effect of different fine-tuning strategies}

For the GPT-2-based models, we employed NTP and TTP fine-tuning strategies. The results show that both strategies help the LM learn backchannels/fillers, with no significant differences between the strategies. However, the embeddings learnt by NTP produce slightly larger silhouette scores (Figure~\ref{fig:silhouette-scores} than those learnt by TTP. Although turn-taking is widely regarded as being highly relevant to the use of fillers and backchannels, our experiment shows that TTP does not significantly improve the learning of backchannel and filler representations compared to NTP. Another surprising result is that the MASK fine-tuning task for the Japanese BERT model makes significantly improves the representation of Japanese backchannels/fillers, comparable to the the results obtained with LLaMA-3 8B and Qwen-3-8B.

\section{Further Analysis via NLG Evaluation}
\label{sec:nlg-evaluation}

Besides using silhouette score to evaluate the representation learning of backchannels/fillers in LMs, we also examine generation results to provide evidence of improved representation after fine-tuning. For this evaluation, we randomly selected 20\% of the utterances from both the English and Japanese corpora. The generation task is defined as follows: given two turns of a dialogue in English or Japanese, both pre-trained and fine-tuned (NTP) LMs are guided to continue writing the dialogue. Based on the generated content, we evaluate backchannels/fil\-lers along five dimensions against the ground-truth responses: (i) frequency of backchannels/fillers, (ii) diversity of backchannel/filler types, (iii) frequency-weighted perplexity of generated backchannels/fil\-lers, (iv) BERTScore (F1), and (v) BLEUScore. The results are summarized in Figure~\ref{fig:nlg_res} (additional details are in Appendix~\ref{app:supp-nlg}).

After fine-tuning, we observe a general increase in both the frequency and diversity of generated backchannels/fillers, as well as improvements in BERTScore (except for Qwen-3 Japanese) and BLEUScore. At the same time, the frequency-weighted perplexity of backchannels/fillers decreases, offering further evidence that LMs achieve better representations of backchannels/fillers after fine-tuning. We use six illustrative examples in Figure~\ref{fig:nlg_examples} (Appendix~\ref{app:qa}) to provide qualitative human evaluation on the generated dialogues compared to the ground truth. 

As we were concerned that fine-tuning LMs in this study might undermine a LM's general capability of language understanding, we investigate this issue with an additional task (reported in Table~\ref{tab:dialogue-act}, Appendix~\ref{app:da-prediction}), which indicates minor side effects.

\section{Conclusion}
\label{sec:conclusions}

We investigate the representations of backchannels/fillers in dialogue corpora learned by transformer-based language models, through three different fine-tuning strategies, masking (MASK), next token prediction (NTP) and turn-taking prediction (TTP). 
The main findings are:
    Firstly, fine-tuning results in more salient representations of backchannels/fillers as evidenced by the increased clustering performance in semantic space. 
    Secondly, fine-tuned LMs generate utterances that are closer to actual human dialogue, as evidenced by higher backchannel/filler frequency and diversity, lower perplexity on these tokens, and improved similarity to ground-truth conversations (higher BLEU and BERTScore).

Our findings suggest that although backchannels/fillers are typically considered semantically bleached and having only pragmatic functions, their semantic representations are affected by dedicated fine-tuning tasks that incorporate more context information -- in a similar way to content words that have concrete meanings. From a broader perspective, this is a case study to investigate LMs' capability of learning under-represented tokens in training data. 
In a narrow sense, we focus on LMs' capacity of representing backchannels/fillers, which shows the potentials and challenges in developing LMs that can mimic human-like speech styles.

\section*{Limitations}

Within the scope of this study, we consider the following limitations, which we believe can be further addressed in future work. First of all, at the beginning of the data selection for our experiment, we did consider including language resources such as the German corpora Verbmobil (VM2) \citep{kay1992verbmobil} and MUNDEX \citep{turk2023mundex} in order to give our results a broader linguistic basis. However, in the end we exclude the German corpora due to their comparatively small size and less formatted annotation, which, for now, leaves us with only English and Japanese data.
 
Secondly, the study would be more thorough with  additional tests of large language models, such as Gemini and GPT-4. This was difficult given limited computing resources, where fine-tuning language models like LLaMA-3-8B was our limit. Moreover, a further step will be analysing the representation of backchannels/fillers in different hidden layers, instead of focusing on the last hidden layer of the models. There are a number of papers that examine different layers of models to answer the interpretability question (e.g., \citealp[]{jawahar-etal-2019-bert,zhao-etal-2024-layer}). Although we have examined the representation of backchannels/fillers in different hidden layers (results in Appendix~\ref{app:hlayer}), a more systematic analysis is considered as our future work. 

Thirdly, in this paper, we only consider how fine-tuning tasks affect representation learning of backchannels/fillers. We did not study what kind of effects different fine-tuning techniques can bring to the representation learning of backchannels/fillers. We notice that some studies propose different fine-tuning techniques, e.g., surgical fine-tuning \citep{lee2023surgical}, which selects subsets of layers to perform fine-tuning while preserving weights in other layers.

Fourth, there is a big gap between language models and speech models. In speech models, even for the same backchannels/fillers, e.g., `uh' in English, different representations can be expected based on the differences in voice quality, pitch, and emotional state during speech. 
How vocal signals of backchannels/fillers are represented in speech models is a future study we will consider. 

Finally, we chose to conduct qualitative analysis on our generated dialogues from linguistic perspective to check the use of backchannels and fillers (details reported in Appendix~\ref{app:qa}), which we believe adheres to the generally suggested practice when it comes to NLG evaluation: qualitative text analysis is recommended when the goal is to improve the system \cite{VANDERLEE2021101151}. However, we currently do not have a large-scale user study, involving comprehensive human evaluation to further support our claim. We plan recruit native speakers of English and Japanese to evaluate the use of backchannels/fillers in the generated dialogues.

\section*{Supplementary Material}

Code and data are available at \url{https://github.com/colalao/discourse_markers} (for your convenience), and as a future-proof data publication at \url{https://doi.org/10.5281/zenodo.19473821}.

\section*{Ethics statement}

Given the scope of this study, there do not appear to be any ethical issues. All of the data and models used in this study are openly available or open weights. We checked the content of the selected dialogue data and made sure that there is no leakage of participants' personal information such as name and ID. Refinement of text and experiment code were supported by ChaptGPT and GitHub Copilot.

\section*{Acknowledgements}

The early structure of this paper benefited significantly from discussions with Lívia Qian, Erik Ekstedt and Siyang Wang from TMH Department, KTH as well as group members from CLCS lab, SUSTech. We sincerely thank all the reviewers for their feedback on the paper. 
This work is funded by the \href{https://www.dfg.de}{Deutsche For\-schungs\-gemeinschaft (DFG)}: \href{https://gepris.dfg.de/gepris/projekt/438445824}{TRR 318/3 2026 – 438445824}, A02.
This work also acknowledges the funding from Shenzhen Science and Technology Program (No. JCYJ20240813094612017) and Guangdong Province ZJRC Program (No. 2024QN11X145). 
This work also acknowledges the funding from the Swedish Research Council (VR) project 2020-03812.

\bibliography{bibliography-acl2026-representations}

\clearpage
\appendix

\section{Data Statistics}
\label{sec:data}

In this section, we provide details on the backchannels/fillers used in this paper, see Table~\ref{tab:en_dm} (English) and Table~\ref{tab:jap_dm} (Japanese). We selected the 15 most frequent backchannels/fillers for our data analysis.

Table~\ref{tab:en_dm} shows the backchannels/fillers in the Switchboard \citep{LDC-swbd1r2} and MapTask \citep{anderson1991hcrc} corpora that we selected as our study objects.

\begin{table}
\small
\caption{%
    The top 15 selected English backchannels/fillers and their combined frequency of occurrence in \textbf{Switchboard} and \textbf{MapTask} (from a total number of \num{127672} backchannels/fillers).}
\label{tab:en_dm}
\begin{tabularx}{\columnwidth}{XX}
\toprule
    \textbf{Backchannel/Filler} & \textbf{Occurrence} \\
\midrule
    \textit{uh}        & 24.56\% \\
    \textit{yeah}      & 17.36\% \\ 
    \textit{uh-huh}    & 13.46\% \\
    \textit{well}      &  7.21\% \\ 
    \textit{right}     &  6.70\% \\ 
    \textit{oh}        &  5.81\% \\
    \textit{um}        &  5.49\% \\
    \textit{okay}      &  3.76\% \\
    \textit{no}        &  2.63\% \\
    \textit{yes}       &  1.88\% \\ 
    \textit{so}        &  1.08\% \\ 
    \textit{oh yeah}   &  0.93\% \\
    \textit{huh}       &  0.74\% \\
    \textit{mmhmm}     &  0.71\% \\ 
    \textit{of course} &  0.59\% \\
\bottomrule
\end{tabularx}
\end{table}

Table~\ref{tab:jap_dm} shows the backchannels/fillers in the BTSJ 1000 Person Japanese Natural Conversation Corpus (BTSJ) \citep{usami2023} that we selected as our study objects.
The Japanese data was selected based on prior guidance and analyses \citep{kawamori1996phonological, 10.1145/3570945.3607298}. In the table, the backchannel/filler is used to represent its different variants, for example \begin{CJK}{UTF8}{min}`うん' (un, `yeah') includes `うん、' (un,) and `うん。' (un), etc\end{CJK}.

\begin{table*}[ht]
\small
\caption{The top 15 selected Japanese backchannels/fillers (with variants), their transcription, and frequency of occurrence in the \textbf{BTSJ} corpus (from a total number of \num{170898} backchannels/fillers).}
\begin{tabularx}{\textwidth}{lXll}
\toprule
    \textbf{Backchannel/Filler} & \textbf{Variants} & \textbf{Transcription} & \textbf{Occurrence} \\
\midrule
    \begin{CJK}{UTF8}{min}うん\end{CJK} & \begin{CJK}{UTF8}{min}`うんうん', `うんうんうん', `ううん', `うーん', `うんー'\end{CJK} & `un' & 23.18\%  \\
    \begin{CJK}{UTF8}{min}はい\end{CJK} & \begin{CJK}{UTF8}{min}`はいー', `はーい'\end{CJK} & `hai' & 17.41\%  \\
    \begin{CJK}{UTF8}{min}あ\end{CJK}  & \begin{CJK}{UTF8}{min}`ああ', `あああ', `あー', `あっ'\end{CJK} & `a' & 8.67\% \\
    \begin{CJK}{UTF8}{min}え\end{CJK} & \begin{CJK}{UTF8}{min}`ええ', `えええ', `えー', `えっ'\end{CJK} & `e' & 6.08\% \\
    \begin{CJK}{UTF8}{min}そう\end{CJK} & \begin{CJK}{UTF8}{min}`そうそう', `そうそうそう', `そうー', `そーう'\end{CJK} & `sou' & 5.25\%  \\
    \begin{CJK}{UTF8}{min}ま\end{CJK} & \begin{CJK}{UTF8}{min}`まー', `まあ', `まあー'\end{CJK} & `ma' & 4.91\%  \\
    \begin{CJK}{UTF8}{min}なんか\end{CJK} & \begin{CJK}{UTF8}{min}`なんかー', `なんかね', `なんかねー'\end{CJK} & `nanka' & 4.73\%  \\
    \begin{CJK}{UTF8}{min}あの\end{CJK} & \begin{CJK}{UTF8}{min}`あのー', `あのね'\end{CJK} & `ano' & 4.25\% \\
    \begin{CJK}{UTF8}{min}ん\end{CJK} & \begin{CJK}{UTF8}{min}`んー'\end{CJK} & `n' & 2.67\%  \\
    \begin{CJK}{UTF8}{min}そうです\end{CJK} & \begin{CJK}{UTF8}{min}`そうですね', `そうですねー', `そうですよ', `そうですよね', `そうですよねー', `そーうですね', `そうっすね'\end{CJK} & `soudesu' & 2.40\%\\
    \begin{CJK}{UTF8}{min}は\end{CJK} & \begin{CJK}{UTF8}{min}`はは', `ははは', `はー', `はあ', `はあー', `はっ'\end{CJK} & `ha' & 2.24\% \\
    \begin{CJK}{UTF8}{min}ね\end{CJK} & \begin{CJK}{UTF8}{min}`ねー'\end{CJK} & `ne' & 2.19\%  \\
    \begin{CJK}{UTF8}{min}いや\end{CJK} & \begin{CJK}{UTF8}{min}`いやいや', `いやいやいや', `いやー'\end{CJK} & `iya' & 1.77\% \\
    \begin{CJK}{UTF8}{min}へー\end{CJK} & --- & `he-' & 1.65\% \\
    \begin{CJK}{UTF8}{min}そうか\end{CJK} & --- & `souka' & 1.63\% \\
\bottomrule
\end{tabularx}
\label{tab:jap_dm}
\end{table*}

\section{Data Pre-Processing}
\label{sec:dataprop}

As we observed in our data, fillers are usually surrounded by others words with substantial meaning, backchannels on the other hand, usually stand out as an independent utterance. Therefore, our data preprocessing aims to contextualize the backchannels/fillers that are not surrounded by other words so that their meaning can be reflected by the other words, i.e., distributional semantics (\citealp[pp.~1–32]{firth1957synopsis}, \citealp[]{mikolov2013}). Figure~\ref{fig:example-dialogue} illustrates our concept.

We noticed that some of the backchannels/fillers can be subsequences of other words (e.g., \textit{um} is a subsequence in `maxim\textit{um}'), which affects the tokenization of the data when we add backchannels/fillers to the vocabulary as special tokens. In addition, some backchannels/fillers can be ambiguous, for example, \textit{`well'} and \textit{`right'} in English can have substantial meaning, but also serve as feedback, expressing either a positive or negative attitude towards the previous utterance.
The same applies to the Japanese word \mbox{\begin{CJK}{UTF8}{min}`ちょっと'\end{CJK}} (‘chotto’), which can mean ‘a little’ or be used as a backchannel/filler to indicate hesitation, for example.
Our solution to this problem is to add a special token to backchannels/fillers so that the tokenizer will not mistake them.
This practice is based on our observation in the data that for the backchannels/fillers which are ambiguous, the use of these words as backchannels/fillers usually occur as the first word in the utterance, followed by a comma as a filler (e.g., `\textit{well\textbf{,}}' or co-occur with other backchannels/fillers (e.g., \textit{okay, right}).

\section{Implementation Details}
\label{sec:TD}

We used a total of eight L40 48G GPUs for our experiments, with one GPU assigned to each task. For the BERT model, the runtime for each task was approximately 3 hours for both English and Japanese, which includes fine-tuning and extracting embeddings for the clustering tasks. For fine-tuning GPT-2 Japanese and English, each task of the Japanese experiment took about 2 hours, which includes fine-tuning and extracting embeddings for the clustering tasks. For English, the tasks took about 1 hour.

For the fine-tuning of TurnGPT tasks for Japanese and English, we used the following parameters: batch size $4$; weight decay $0.01$; dropout rate $0.3$; learning rate $0.0005$.
A total of $15$ epochs were used to complete the fine-tuning tasks. 
After each epoch, a checkpoint (model) was generated, saving the weight parameters gained during training. The model with the minimum loss value was chosen as the final model for estimating the probability of the turn transition potential.

For LLaMA-3 8B and Qwen-3 8B, we used LoRa to accelerate fine-tuning. The parameters for LoRA were set as follows. The rank of the low-rank matrices is $16$, and a dropout rate of $0.1$ is used to the LoRA layers to improve regularization. LoRA is applied to the ‘q\_proj’ and ‘v\_proj’ layers within the model's attention mechanism.
For LLaMA-3 8B, each task took about 8 hours for the Japanese data and about 7 hours for the English data. In contrast, the Qwen-3 8B model required around 15 and 10 hours for the same tasks, respectively.

For $k$-means clustering, we first standardized the obtained embeddings and applied PCA (Principal Component Analysis) to reduce the dimensionality to 100, facilitating subsequent clustering operations. The number of clusters $k$ ranged from 2 to 15, and the optimal $k$ was selected based on the highest silhouette score achieved.

\subsection{Other Method for Obtaining Meaning Representation}
\label{no_contra}

Parallel to fine-tuning, there are also some other methods to acquire the meaning representations. In terms of learning the representation of backchannels/fillers in LMs, one important method is contrastive learning \cite{gao-etal-2021-simcse}.
To the best of our knowledge, the only highly relevant work for our study is by \citet{qian24b_interspeech}, who use contrastive learning methods to test how the speech models HuBert \cite{hsu2021hubert} and Whisper \cite{radford23a}, as well as the language model BERT can represent the different functions of feedback signals (namely backchannels). Their results show that the learnt embeddings can carry information about different functions a backchannel possesses (see Figure~3 in \citealp{qian24b_interspeech} for details). 
In this paper, we focus on fine-tuning only as our method to acquire the representation of backchannels/fillers. 
We leave contrastive learning aside, since, unlike in \citet{qian24b_interspeech}, the types of backchannels/fillers are much larger and thus induce much higher computational costs.
The detailed reason for not using contrastive learning to get the representation of backchannels/fillers is mainly due to its computational cost and uncertainty (the setting and objects of study are different and simpler in \citealp[]{qian24b_interspeech}). 
As an example, in the Japanese data, although we listed the most frequent backchannels/fillers in our paper, we also have the tail examples (those examples with very few occurrence, less than 50 times in the whole dataset) and together we have more than 80 types of backchannels/fillers. 

The most important part of contrastive learning is its negative sampling mechanism. In negative sampling, given a positive example (a natural utterance with a backchannel/filler), several negative samples are generated (synthesised utterances in which the original backchannels/fillers are replaced with random ones). These negative samples are then used jointly in training to move positive examples closer to each other. In \citet{qian24b_interspeech} this was doable and worked well because the candidate negative examples are just selected from the backchannels which are classified as feedback, thus limited negative sample candidates and therefore much lower computational cost. Moreover, both positive and negative samples are feedback but with different function types so the experimental results are quite controllable.

\section{Further Supporting Results}
\label{sec:fsr}

\subsection{Representation of Backchannels/Fillers From LMs' Different Hidden Layers}
\label{app:hlayer}

\begin{figure*}[ht]
    \centering
    \includegraphics[width=0.95\linewidth]{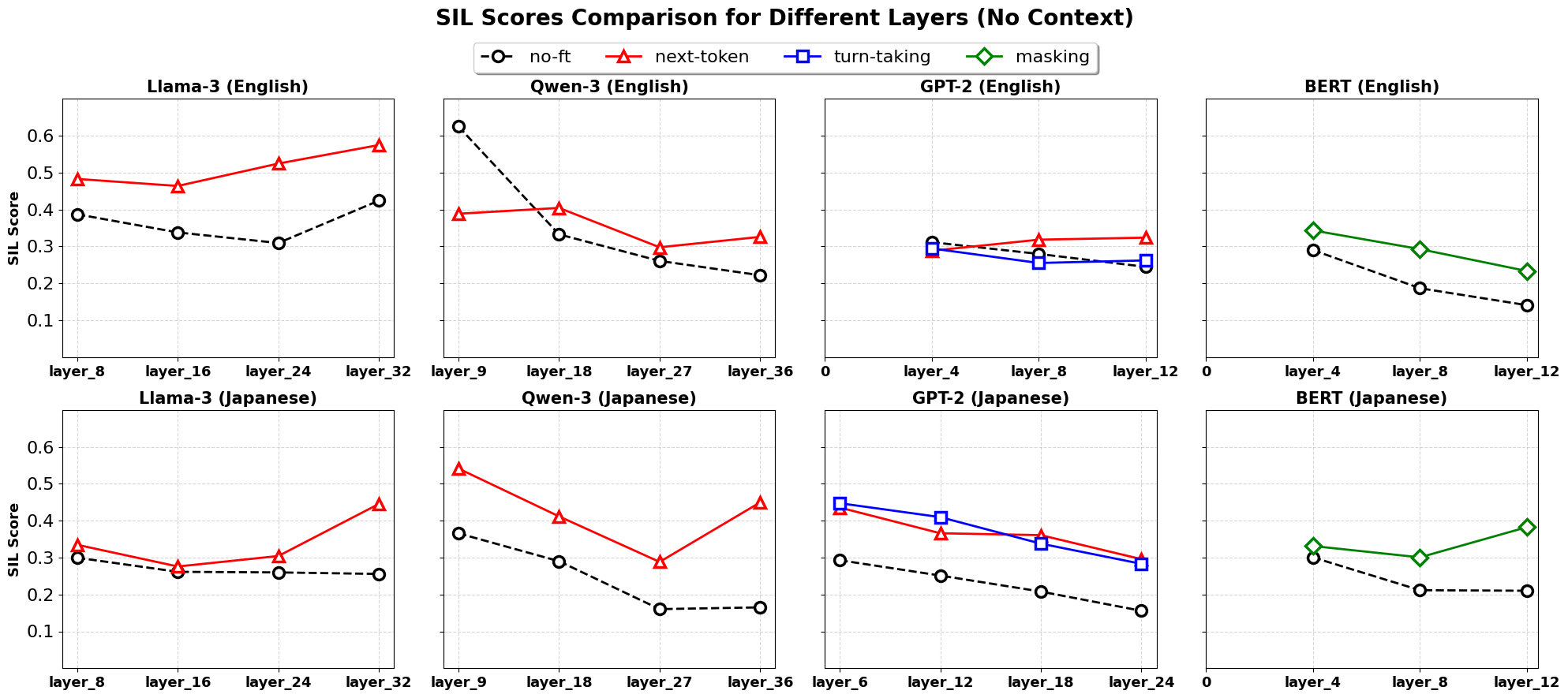}
    \caption{Improvement of the representation of backchannels/fillers on selected layers. Setting: \textbf{no-context}.}
    \label{fig:layerDiff1}
\end{figure*}

\begin{figure*}[ht]
    \centering
    \includegraphics[width=0.95\linewidth]{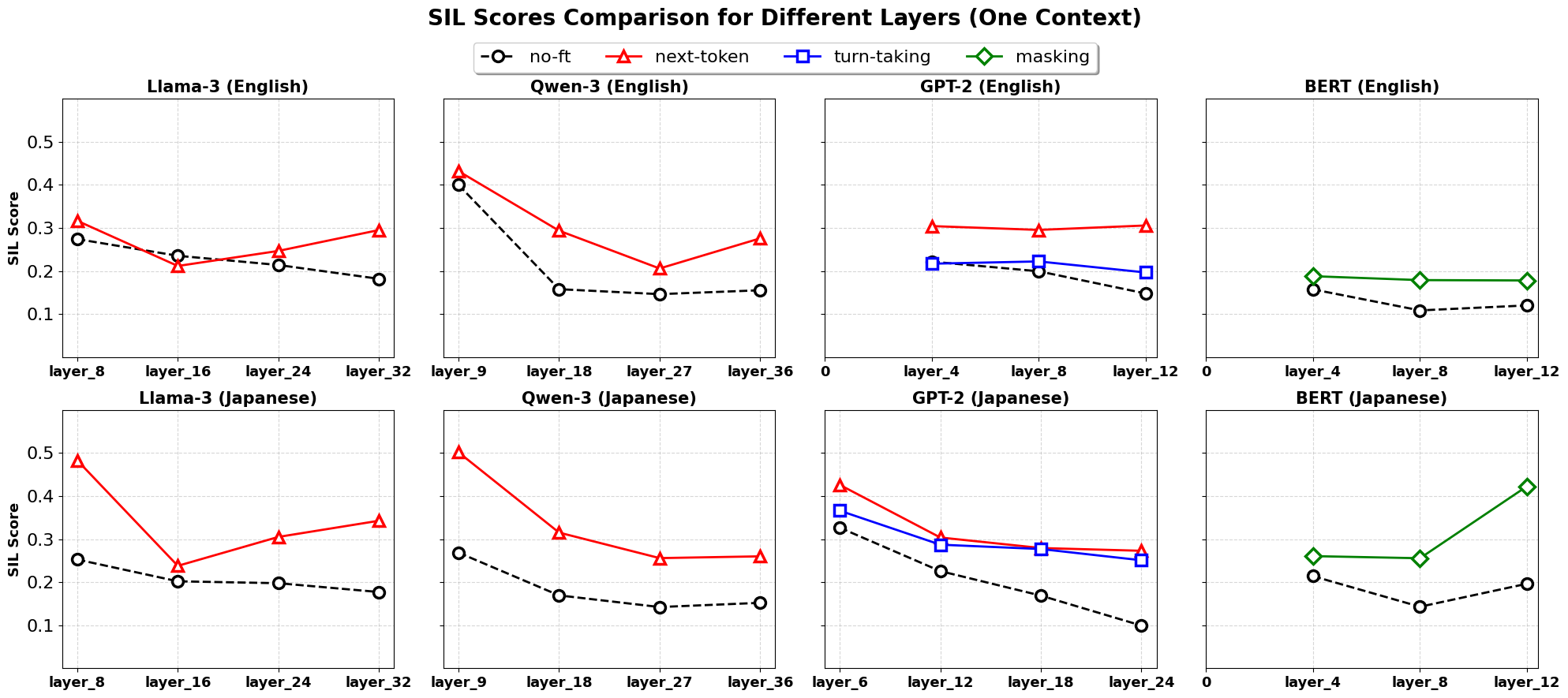}
    \caption{Improvement of the representation of backchannels/fillers on selected layers. Setting: \textbf{one-context}.}
    \label{fig:layerDiff2}
\end{figure*}

In the main text, we evaluate the improvement of representations by applying clustering analysis on the weights extracted from the last hidden layers of the LMs. Here we include additional experiments to further investigate the improvement from different layers before and after fine-tuning. To reflect the difference from different hidden layers, for LLaMA-3 8B, we selected layers 8, 16, 24, 32, for Qwen-3 8B, we selected layers 9, 18, 27, 36, for GPT-2 English and BERT models, we selected layers 4, 8, 12, for GPT-2 Japanese model, we selected layers 6, 12, 18, 24. The results are shown in Figure~\ref{fig:layerDiff1} and Figure~\ref{fig:layerDiff2}. 

First of all, it turns out that improvement of representation in different selected layers, as quantified by silhouette score, is in general observable in the fine-tuned models in both settings. Second, before fine-tuning silhouette scores are generally higher in the shallow layer than the deeper layers, which indicates that shallow layers have better representations of backchannels/fillers compared to deeper layers. Fine-tuning seems to break this tendency. For example, fine-tuned Llama-3 8b models, under two settings, have improved representation in their deeper hidden layers. Similar patterns can also been seen in Japanese BERT, GPT models and Qwen-3b for English.

\subsection{General Performance of Fine-tuned LMs}
\label{app:da-prediction}

Our biggest concern toward the experimental results reported in this study is whether the fine-tuning tasks, used to leverage the usage of backchannels and fillers in LMs, has the potential to undermine an LM's general performance. We therefore investigate this issue by checking the LMs' capability of doing a dialogue act prediction task, which is, in principle, feasible on the \textbf{Switchboard} and \textbf{MapTask} datasets. A dialogue act is an annotation label on utterance(s) of a dialogue that indicates the communicative functions of the utterance. The communicative functions include \textit{`statement', `agreement/ac\-cept', `wh-question', `backchannel/acknowledge'}, etc. \citep{StolckeRies2000}. We specifically compare the accuracy on dialogue act prediction task before and after fine-tuning and examine whether performance changes. The results are summarised in the Table~\ref{tab:dialogue-act}.
As can be seen, BERT, GPT-2, and Qwen-3 exhibit a slight decrease in accuracy after fine-tuning, while \mbox{LLaMA-3} shows improvements. These results suggest that our fine-tuning strategy does not compromise the language understanding ability of models and may, in some cases, even yield modest gains.

\begin{table}
\centering
\caption{%
    Accuracy on the dialogue act classification task (MapTask dataset) before and after fine-tuning, with linear probing. We used five training epochs for BERT and GPT-2, and three training epochs for LLaMA-3 and Qwen-3.}
\begin{tabularx}{\columnwidth}{lXX}
\toprule
    \textbf{Model}   & \textbf{no fine-tuning}  & \textbf{fine-tuning} \\
\midrule
    BERT    & 59.0\% & 58.5\% \\
    GPT-2   & 56.6\% & 55.7\% \\
    LLaMA-3 & 45.4\% & 51.2\% \\
    Qwen-3  & 45.3\% & 42.9\% \\
\bottomrule
\end{tabularx}
\label{tab:dialogue-act}
\end{table}

\begin{figure*}[ht!]

\begin{subfigure}{0.48\textwidth}
\small
\begin{tcolorbox}[title = {\textbf{Dialogue 1}}]
\textbf{Input:}
/A \textbf{<ds>yeah</ds>} /A ///

/B so we went in and the financed the car for five and we didn't know that we got kind of taken just a little bit so we kind of upset that we having to you know pay this fifth year on the car because it was just not a wise thing but we learned a good lesson so you know /B ///
\vspace{1mm}

\textbf{Ground truth:}
/A \textbf{<ds>well</ds>} five years is not that \textbf{<ds>uh</ds>} five years is not that bad of a problem only because you paying more interest but your payments are lower /A
\vspace{1mm}

\textbf{Output (no\_ft, one\_ctx):}
/A this all the experience that we got in that situation yeah I just want to say that it was not just the car that we had to deal with but it was /A

\vspace{1mm}
\textbf{Output (ft, one\_ctx)}:
/A \textbf{<ds>yeah</ds>} we were just \textbf{<ds>um</ds>} my my parents bought the car that we would drive in high school and they put like two hundred down and and I … /A
\end{tcolorbox}
\vspace{0.3cm}
\end{subfigure}
\hfill
\begin{subfigure}{0.48\textwidth}
\small
\begin{tcolorbox}[title = {\textbf{Dialogue 2}}]
\textbf{Input:}
/A but then \textbf{<ds>uh</ds>} we also didn't do spend \textbf{<ds>uh</ds>} time with them trying to trying to \textbf{<ds>uh</ds>} you know just family time /A ///

/B \textbf{<ds>uh-huh</ds>}/B ///
\vspace{1mm}

\textbf{Ground truth:}
/A \textbf{<ds>uh</ds>} we had a at a fairly young age we started teaching them to play games and card games and so forth and and \textbf{<ds>uh</ds> <ds>uh</ds>} but we didn't \textbf{<ds>uh</ds>} do it as much or as often as \textbf{<ds>uh</ds>} I would have liked to I would like to do that as a as a family and /A
\vspace{1mm}

\textbf{Output (no\_ft, one\_ctx)}:
/A with the family that we have now with the family that we have now. What are you going to do with that kind of logic? I mean, I'd like to solve this problem.  But I can't. Can you help me? I have a little bit of time before the family comes over.  So, can you give … /A
\vspace{1mm}

\textbf{Output (ft, one\_ctx)}:
/A I think \textbf{<ds>uh</ds>} my father was really into them and we would you know go camping and stuff and \textbf{<ds>uh</ds>} you know I think if you do that then it important to spend some time with them you know /A ///

/B \textbf{<ds>uh-huh</ds>} /B 
\end{tcolorbox}
\vspace{0.3cm}
\end{subfigure}

\begin{subfigure}{0.48\textwidth}
\small
\begin{tcolorbox}[title = {\textbf{Dialogue 3}}]
\textbf{Input:}
/A so it becomes \textbf{<ds>uh</ds>} a general battle of the w- sort of the election of the lesser of two evils I think so /A ///

/B \textbf{<ds>yeah</ds>} /B ///
\vspace{1mm}

\textbf{Ground truth:}
/A I don't know I just \textbf{<ds>uh</ds>} I unhappy with it but and well I I sort of semi-fortunate right now I a graduate student so I don't make that much /A
\vspace{1mm}

\textbf{Output (no\_ft, one\_ctx):}
/A The concept of "lesser of two evils" suggests that when faced with a difficult decision, it's better to choose the option that is less undesirable. In the context of the 202 /A
\vspace{1mm}

\textbf{Output (ft, one\_ctx):}
/A it always seems to be that way \textbf{<ds>so</ds> <ds>um</ds>} what do you think is the other major issue now /A ///

/B \textbf{<ds>um</ds>} I I don't know I think …/B 
\end{tcolorbox}
\vspace{0.3cm}
\end{subfigure}
\hfill
\begin{subfigure}{0.48\textwidth}
\small
\begin{tcolorbox}[title = {\textbf{Dialogue 4}}]
\textbf{Input:}
/A \textbf{<ds>yeah</ds> <ds>well</ds>} my \textbf{<ds>uh</ds>} my \textbf{<ds>uh</ds>} probably one of the biggest decisions I think that was very strengthening for our family was rather than have one child make that decision /A ///

/B \textbf{<ds>uh-huh</ds>} /B ///
\vspace{1mm}

\textbf{Ground truth:}
/A than just delegate it I think that they they had a great deal of \textbf{<ds>um</ds>} all the brothers and sisters got together and they actually had a conference /A
\vspace{1mm}

\textbf{Output (no\_ft, one\_ctx):}
/A one year. For one of one or a rather long time I am one I do one I did one I have one I will one one I had one /A

\textbf{Output (ft, one\_ctx)}:
/A I think we went to church and I think we sought \textbf{<ds>um</ds>} the advice of those that could see it from a different perspective /A
\end{tcolorbox}
\vspace{0.3cm}
\end{subfigure}
\hfill

\begin{subfigure}{0.48\textwidth}
\small
\begin{tcolorbox}[title = {\textbf{Dialogue 5}}]
\textbf{Input:}
/A you don't have a stone circle? /A ///
/B no /B ///
/A \textbf{<ds>okay</ds>} do you have a stone circle on the map at all /A ///
/B \textbf{<ds>um</ds>} i have a ruined monastery /B ///
\vspace{1mm}

\textbf{Ground truth:}
/A fine \textbf{<ds>um</ds>} do you have any obstacle below the telephone kiosk /A
\vspace{1mm}

\textbf{Output (no\_ft, one\_ctx):}
/A hat's good too we can explore it 1a /A
\vspace{1mm}

\textbf{Output (ft, one\_ctx):}
/A you have a ruined monastic \textbf{<ds>uh-huh</ds>} \textbf{<ds>well</ds>}  you're going to go down and then you're going to go right and then you're going to go up over the top of the stone monument and then down again /A
\end{tcolorbox}
\end{subfigure}
\hfill
\begin{subfigure}{0.48\textwidth}
\small
\begin{tcolorbox}[title = {\textbf{Dialogue 6}}]
\textbf{Input:}
/A \textbf{<ds>eh</ds>} can you dr-- actually draw round the /A ///

/B \textbf{<ds>well</ds>} \textbf{<ds>yeah</ds>} i suppose /B ///
/A \textbf{<ds>well</ds>} if you could draw round them then/A ///
\vspace{1mm}

\textbf{Ground truth:}
/B straight down to the giraffes /B
\vspace{1mm}

\textbf{Output (no\_ft, one\_ctx):}
/B then maybe we can make a plan  1. but then i can't actually get from the other end of the rope bridge to the giraffe /B
\vspace{1mm}

\textbf{Output (ft, one\_ctx)}:
/B i don't know whether there's a river or not i've got a river which is on the left-hand side /B
\end{tcolorbox}
\end{subfigure}

\caption{Six illustrative examples generated with the non fine-tuned and fine-tuned LLaMA-3 model in the one-context setting.}
\label{fig:nlg_examples}
\end{figure*}

\subsection{Further Qualitative and Quantitative Analyses of Generation}
\label{app:qa}

As further evidence of improved representation capability of backchannels/fillers in fine-tuned LMs, in this subsections, we performs a qualitative analysis on the generation results under the \textbf{fine\_tuning, no\_context} and \textbf{fine\_tuning, one\_context} settings respectively. Both fine-tuned and non fine-tuned LMs are guided to perform an NLG task which requires them to complete the dialogue based on the given context. We selected around 4000 utterances from both English and Japanese corpora. 
As summarised in Tables~\ref{tab:nlg1} and \ref{tab:nlg2}, the results show that the fine-tuned LLaMA-3 model increases the usage of backchannels/fillers in the generation task (i.e., the frequency increases). Moreover, different types of backchannels/fillers are used (diversity increases). This is further and crucial evidence showing that the models do learn the representation of backchannels/fillers. 

As our small scale human evaluation, here we select six representative dialogue examples (see Figure~\ref{fig:nlg_examples}) generated by the LLaMA-3 8b model in English as our qualitative analysis on the LLM's capability of generating backchannels and fillers after fine-tuning. Below is the instruction for reading the dialogue examples: 
\begin{itemize}
    \item \textbf{/A…/A and /B…/B}: the beginning and the end of the utterance(s) from speaker A, speaker B.
    \item \textbf{///}: marking of turn shifts.
    \item \textbf{<ds>…</ds>}: annotation of backchannels/fillers.
    \item \textbf{Input}: the incomplete dialogues used for the generation task.
    \item  \textbf{Ground truth}: the utterances which are the continuation of the incomplete dialogue. (\textbf{Input})
    \item \textbf{Output (no\_ft, one\_ctx)}: The generated utterances from the off-the-shelf LLaMa-3 model under the one-context setting.
    \item \textbf{Output (ft, one\_ctx)}: The generated utterances from the fine-tuned LLaMa-3 model under the one-context setting.
\end{itemize}

We notice that only under the ft, one\_cx setting (fine-tuned, one-context ) does the LLaMA-3 model begin to use backchannels/fillers in its generated output, which further indicates the effectiveness of fine-tuning and context information. 

The ft, one\_ctx outputs of examples dialogues 1 and 4 show that the fine-tuned LLaMA-3 model can generate feedback signals (e.g., \textit{yeah}) to acknowledge the previous utterance. It can also use fillers to mimic disfluency or hesitation in an utterance (e.g., \textit{um} in both examples).

In output (ft, one\_ctx) of example dialogue 2, we can see that the fine-tuned LLaMA-3 model can generate a backchannel to show attentiveness to the previous speaker's utterance (the \textit{uh-huh} in response to the utterance by speaker A). 

In output (ft, one\_ctx) of dialogue example 3, we can see a more complex use of fillers: the first filler \textit{so} indicates the transition of topic, and again a filler \textit{um} indicates disfluency or cognitive load in the production of speech. The examples shown here all support the idea that we can indeed fine-tune a LM to become a conversational agent which mimics the way humans produce speech.

A notable finding is the emergence of advanced pragmatic competencies of backchannels and fillers in the fine-tuned LMs. As illustrated in output (ft, one\_ctx) of dialogue examples 3 and 5, the fine-tuned LLaMA-3 model successfully distinguishes between the structural role of \textit{so}, a filler which marks a sequence transition, and the cognitive signalling of the filler \textit{um}.

Similarly, in output (ft, one\_ctx) in example dialogue 5, the \textit{uh-huh} is a backchannel which indicates confirmation and signals grounding while \textit{well} is a filler which shows transition and framing. Furthermore, as evidenced by example dialogue 6, the fine-tuned LM does not overgenerate backchannels and fillers to complete every dialogue. Instead, it demonstrates a nuanced capacity to omit backchannels and fillers when they are not pragmatically required.

\subsection{Supporting Details for NLG Evaluation Result Shown in Figure~\ref{fig:nlg_res}}
\label{app:supp-nlg}

Tables~\ref{tab:nlg1} and~\ref{tab:nlg2} summarize the evaluation results of models generating the next utterance based on a two-turn dialogue context, using the entire 20\% evaluation split of the English and Japanese corpora (one utterance per speaker). The generated utterance is constrained to match the length of the ground-truth response. Metrics are defined as follows: 
(1) \textbf{Diversity} counts the number of distinct backchannel/filler types; 
(2) \textbf{Frequency} is the proportion of backchannel/filler tokens in the generated text, normalized by the total number of words for English and by the total number of characters for Japanese; 
(3) \textbf{Perplexity} is the frequency-weighted perplexity computed only on generated backchannel/filler tokens, 
$\mathrm{PPL}=\exp\!\big(-\frac{1}{\sum_i f_i}\sum_i f_i \log p(w_i \mid c_i)\big)$, 
where $w_i$ is a generated backchannel/filler, $c_i$ its context, and $f_i$ its count; 
(4) \textbf{BERTScore (F1)} and (5) \textbf{BLEUScore} are computed against the ground-truth continuation. 
Backchannels/fillers are detected using a curated lexicon with boundary-aware matching. We report results for both the pre-trained (\texttt{no\_ft\_one}) and fine-tuned (\texttt{ft\_one}) models.

\begin{table*}
\scriptsize
\centering
\caption{%
    Evaluation metrics for generated backchannels/fillers in the \textbf{English} NLG task.}
\begin{tabularx}{\textwidth}{lXXXXXXX}
\toprule
    \textbf{Metric}
    & \multicolumn{2}{c}{\textbf{LLaMA-3}} 
    & \multicolumn{2}{c}{\textbf{Qwen-3}} 
    & \multicolumn{2}{c}{\textbf{GPT-2}} \\
\cmidrule(lr){2-3} \cmidrule(lr){4-5} \cmidrule(lr){6-7}
    & no\_ft, one\_ctx & ft, one\_ctx & no\_ft, one\_ctx & ft, one\_ctx & no\_ft, one\_ctx & ft, one\_ctx \\
\midrule
Diversity $\uparrow$           & 73 & 83 & 87 & 95 & 68 & 90 \\
Frequency (\%) $\uparrow$     & 4.29\% & 18.61\% & 5.33\% & 9.19\% & 6.68\% & 17.43\% \\
Perplexity  $\Downarrow$       & 197.67 $\pm$ 84.75 & 5.30 $\pm$ 1.10 & 202.06 $\pm$ 92.69 & 91.32 $\pm$ 11.70 & 158.64 $\pm$ 72.19 & 6.98 $\pm$ 1.51 \\
BERTScore (F1 \%) $\uparrow$  & 78.69\% $\pm$ 0.13\% & 79.99\% $\pm$ 0.04\% & 76.02\% $\pm$ 0.06\% & 79.73\% $\pm$ 0.05\% & 79.67\% $\pm$ 0.05\% & 79.99\% $\pm$ 0.04\% \\
BLEUScore  $\uparrow$         & 0.0600 $\pm$ 0.0012 & 0.0697 $\pm$ 0.0012 & 0.0698 $\pm$ 0.0012 & 0.0800 $\pm$ 0.0013 & 0.0544 $\pm$ 0.0011 & 0.0731 $\pm$ 0.0012 \\
\bottomrule
\end{tabularx}
\label{tab:nlg1}
\end{table*}

\begin{table*}
\scriptsize
\centering
\caption{%
    Evaluation metrics for generated backchannels/fillers in the \textbf{Japanese} NLG task.}
\begin{tabularx}{\textwidth}{lXXXXXXX}
\toprule
    \textbf{Metric}
    & \multicolumn{2}{c}{\textbf{LLaMA-3}} 
    & \multicolumn{2}{c}{\textbf{Qwen-3}} 
    & \multicolumn{2}{c}{\textbf{GPT-2}} \\
\cmidrule(lr){2-3} \cmidrule(lr){4-5} \cmidrule(lr){6-7}
    & no\_ft, one\_ctx & ft, one\_ctx & no\_ft, one\_ctx & ft, one\_ctx & no\_ft, one\_ctx & ft, one\_ctx \\
\midrule
    Diversity  $\uparrow$         & 90 & 117 & 117 & 133 & 99 & 115 \\
    Frequency (\%) $\uparrow$     & 0.31\% & 7.57\% & 0.27\% & 0.54\% & 0.63\% & 2.10\% \\
    Perplexity   $\Downarrow$      & 255.94 $\pm$ 72.48 & 28.51 $\pm$ 4.58 & 748.63 $\pm$ 31.24 & 90.17 $\pm$ 22.50 & 195.36 $\pm$ 93.86 & 53.51 $\pm$ 28.74 \\
    BERTScore (F1 \%) $\uparrow$  & 62.68\% $\pm$ 0.06\% & 66.31\% $\pm$ 0.04\% & 63.55\% $\pm$ 0.07\% & 63.39\% $\pm$ 0.04\% & 62.96\% $\pm$ 0.05\% & 63.75\% $\pm$ 0.05\% \\
    BLEUScore   $\uparrow$       & 0.00005 $\pm$ 0.00005 & 0.00030 $\pm$ 0.00010 & 0.00021 $\pm$ 0.00010 & 0.00036 $\pm$ 0.00010 & 0.00019 $\pm$ 0.00010 & 0.00026 $\pm$ 0.00005 \\
\bottomrule
\end{tabularx}
\label{tab:nlg2}
\end{table*}

\subsection{Additional t-SNE Visualisations}
\label{app:additional-tSNE}

\begin{figure*}
    \centering
    \includegraphics[width=0.95\linewidth]{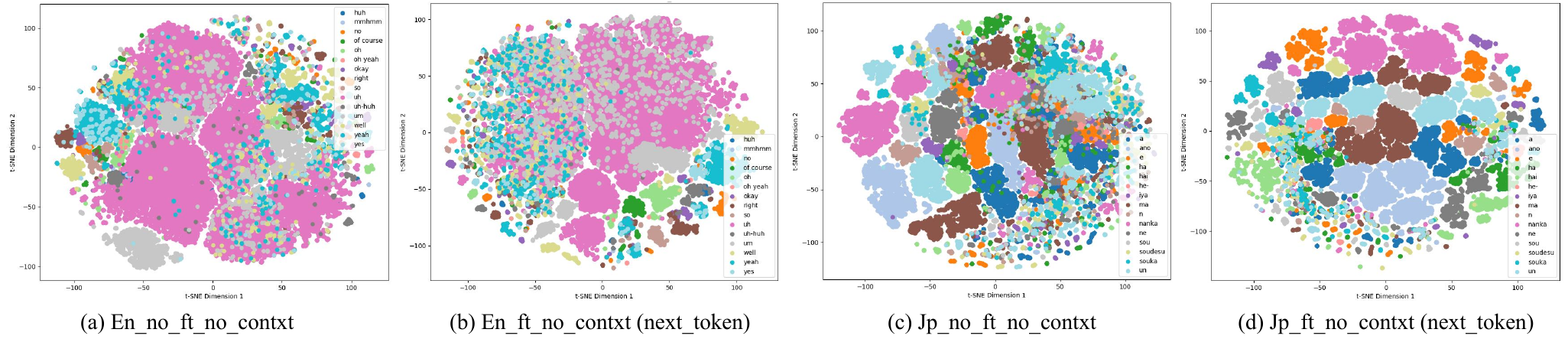}
    \caption{t-SNE plots of the backchannels/fillers embeddings from the \textbf{LLaMA-3} model (\textbf{NTP}). Setting: \textbf{no-context}.}
    \label{fig:tsne_result1}
\end{figure*}

\begin{figure*}
    \centering
    \includegraphics[width=0.95\linewidth]{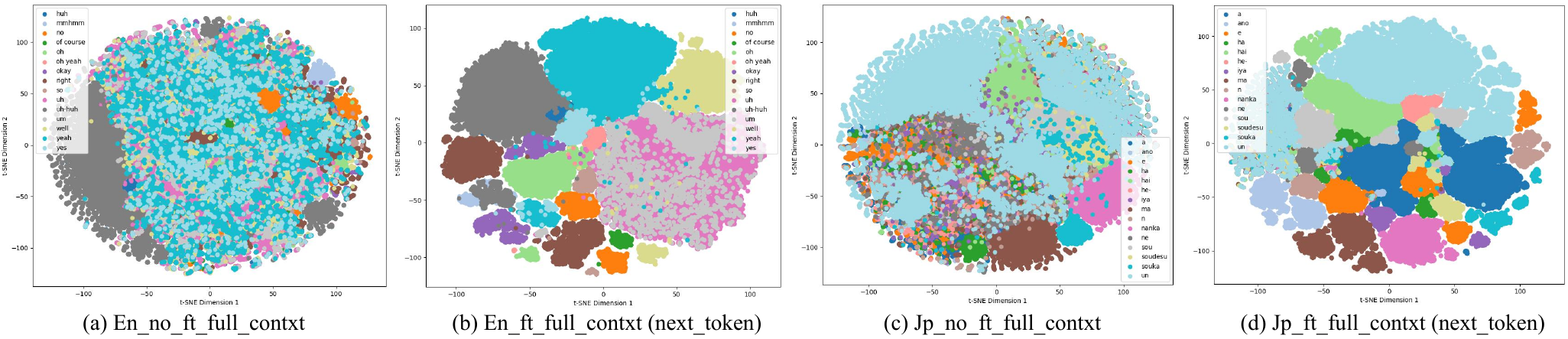}
    \caption{t-SNE plots of the  backchannels/fillers embeddings from the \textbf{LLaMA-3} model (\textbf{NTP}). Setting: \textbf{full-context}.}
    \label{fig:tsne_result2}
\end{figure*}

\begin{figure*}
    \centering
    \includegraphics[width=0.95\linewidth]{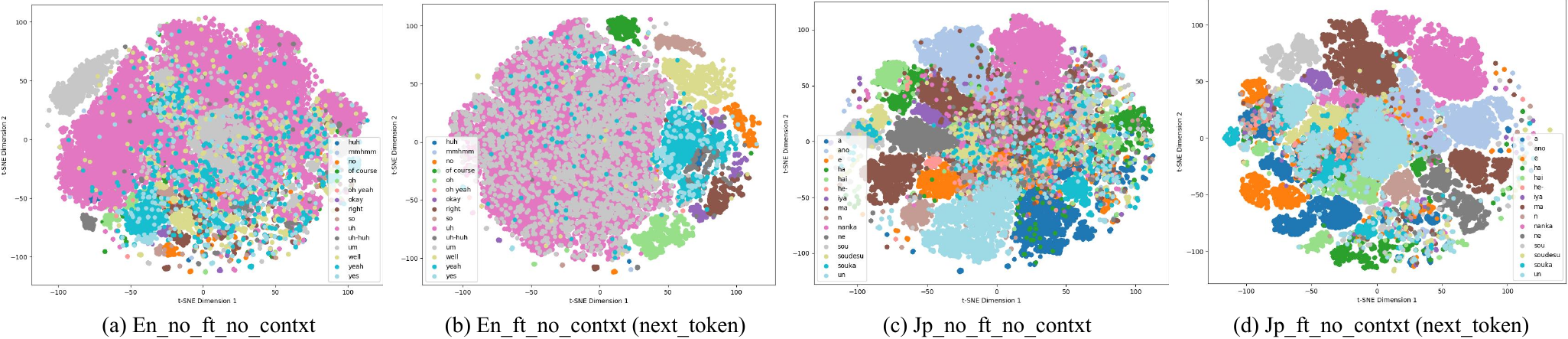}
    \caption{t-SNE plots of the backchannels/fillers embeddings from the \textbf{Qwen-3} model (\textbf{NTP}). Setting: \textbf{no-context}.}
    \label{fig:tsne_result3}
\end{figure*}

\begin{figure*}
    \centering
    \includegraphics[width=0.95\linewidth]{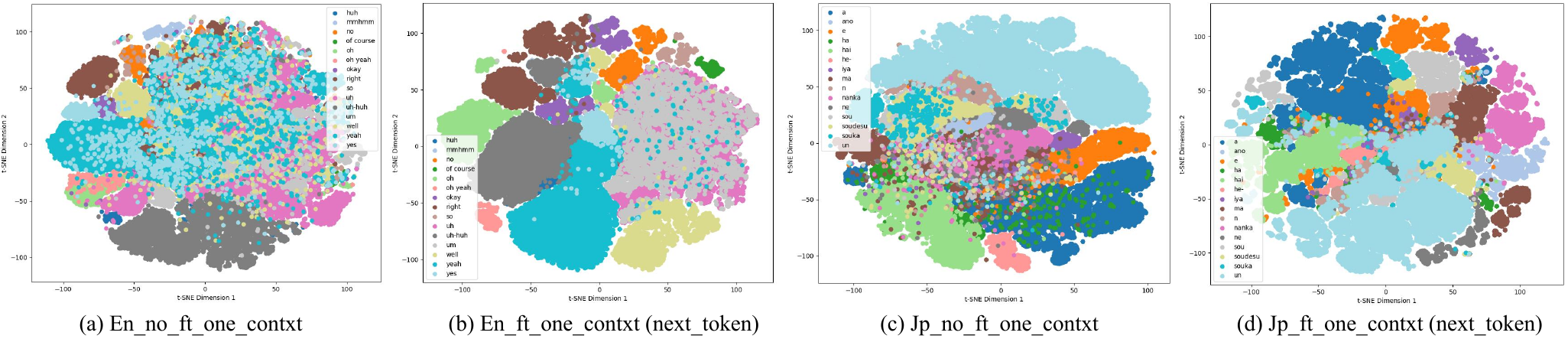}
    \caption{t-SNE plots of the backchannels/fillers embeddings from the \textbf{Qwen-3} model (\textbf{NTP}). Setting: \textbf{one-context}.}
    \label{fig:tsne_result4}
\end{figure*}

\begin{figure*}
    \centering
    \includegraphics[width=0.95\linewidth]{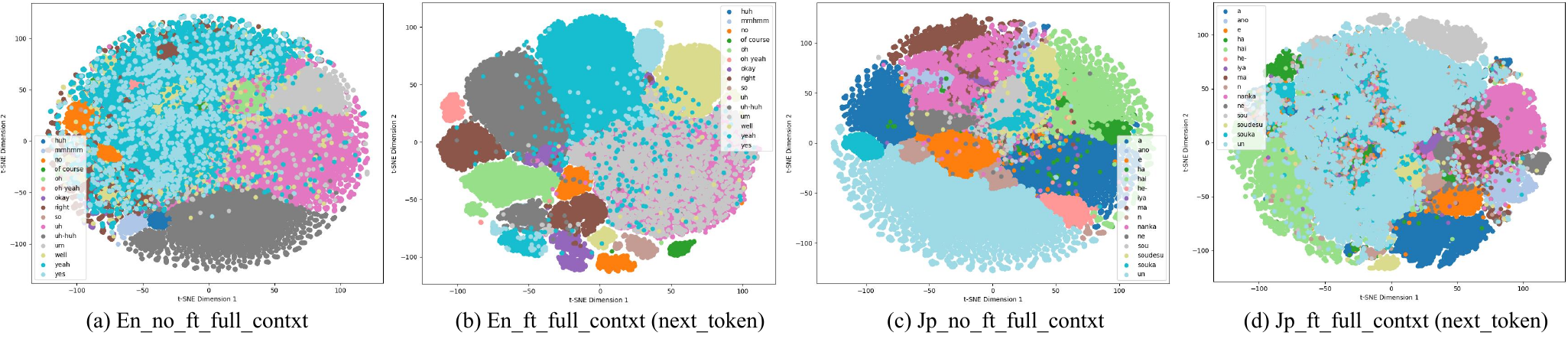}
    \caption{t-SNE plots of the backchannels/fillers embeddings from the \textbf{Qwen-3} model (\textbf{NTP}). Setting: \textbf{full-context}.}
    \label{fig:tsne_result5}
\end{figure*}

\begin{figure*}
    \centering
    \includegraphics[width=0.95\linewidth]{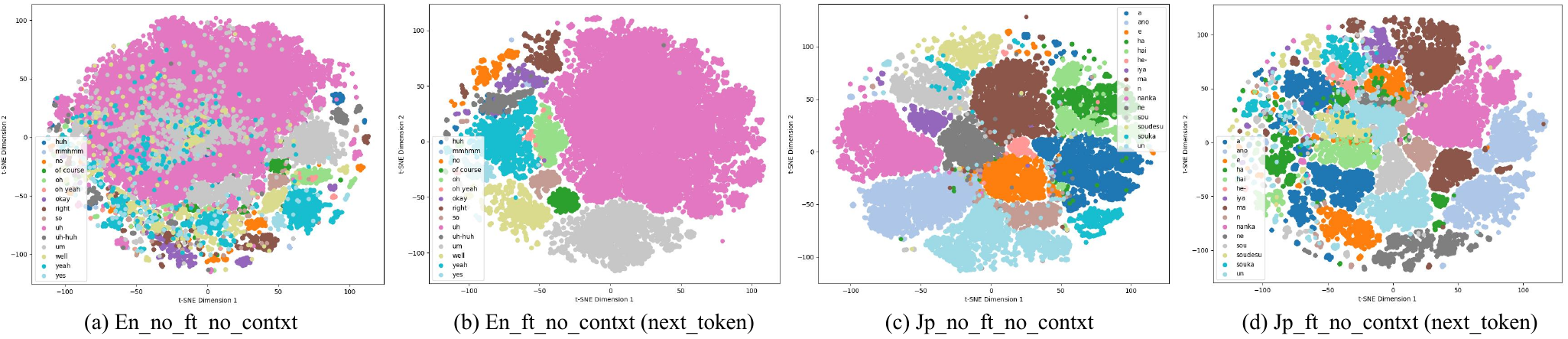}
    \caption{t-SNE plots of the backchannels/fillers embeddings from the \textbf{GPT-2} model (\textbf{NTP}). Setting: \textbf{no-context}.}
    \label{fig:tsne_result6}
\end{figure*}

\begin{figure*}
    \centering
    \includegraphics[width=0.95\linewidth]{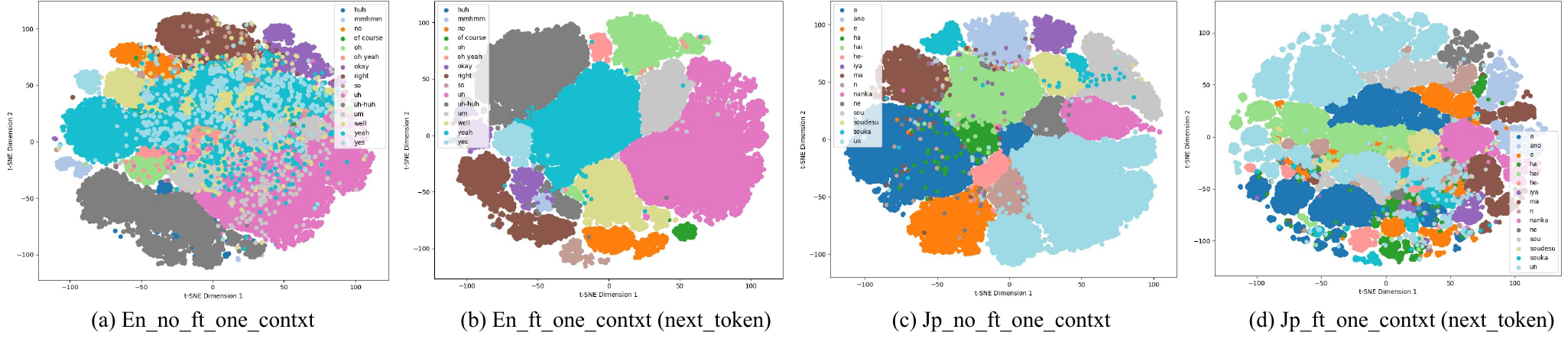}
    \caption{t-SNE plots of the backchannels/fillers embeddings from the \textbf{GPT-2} model (\textbf{NTP}). Setting: \textbf{one-context}.}
    \label{fig:tsne_result7}
\end{figure*}

\begin{figure*}
    \centering
    \includegraphics[width=0.95\linewidth]{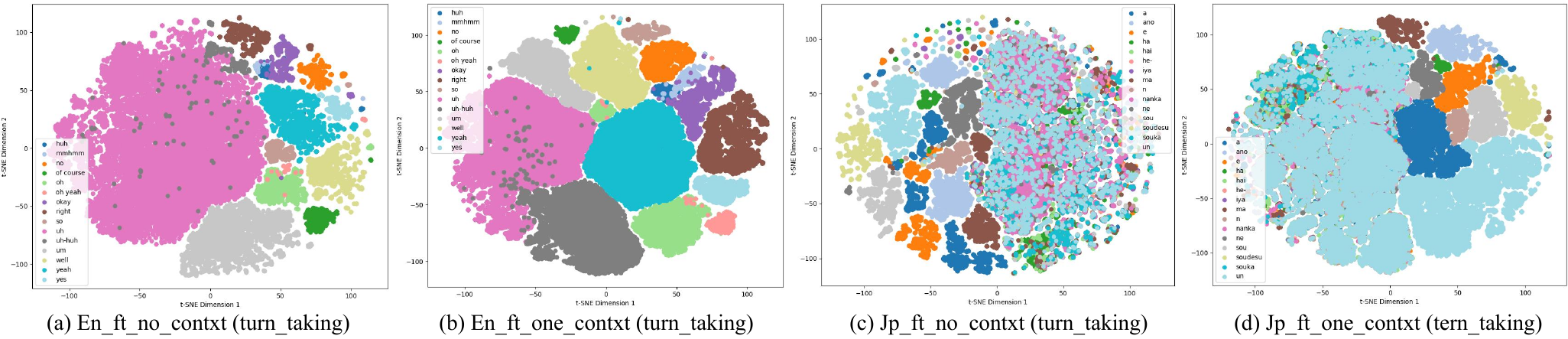}
    \caption{t-SNE plots of the backchannels/fillers embeddings from the \textbf{GPT-2} model (\textbf{TTP}). Setting: \textbf{no-} and \textbf{one-context}.}
    \label{fig:tsne_result8}
\end{figure*}

\begin{figure*}
    \centering
    \includegraphics[width=0.95\linewidth]{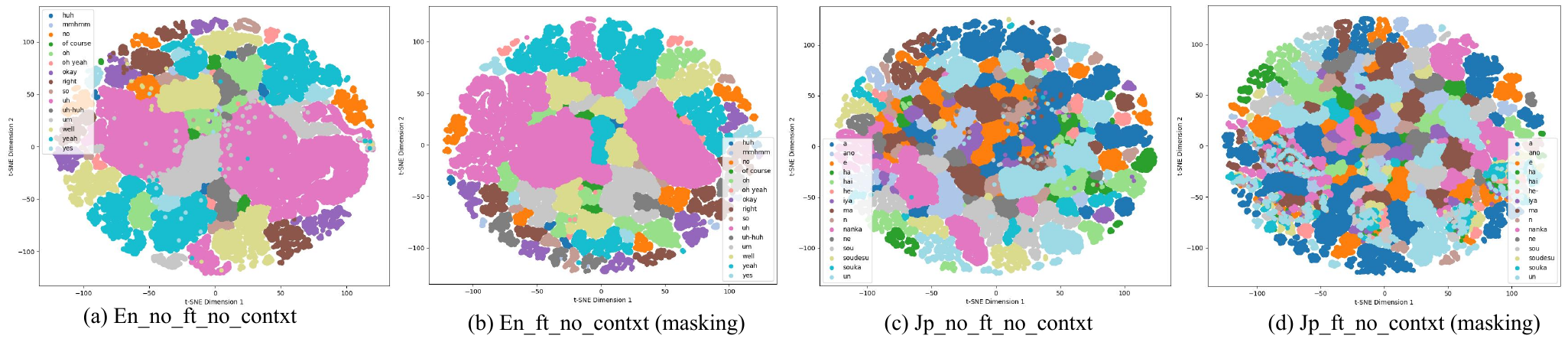}
    \caption{t-SNE plots of the backchannels/fillers embeddings from the \textbf{BERT} model (\textbf{MASK}). Setting \textbf{no-context}.}
    \label{fig:tsne_result9}
\end{figure*}

\begin{figure*}
    \centering
    \includegraphics[width=0.95\linewidth]{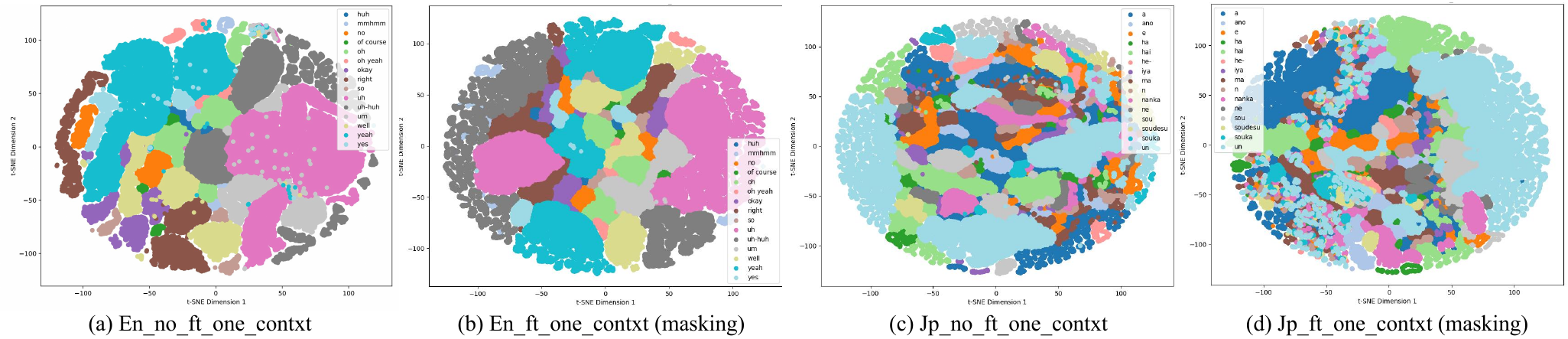}
    \caption{t-SNE plots of the  backchannels/fillers embeddings from the \textbf{BERT} model (\textbf{MASK}): Setting: \textbf{one-context}.}
    \label{fig:tsne_result10}
\end{figure*}

Figure~\ref{fig:tsne_result} in Section~\ref{sec:res} shows t-SNE plots of how the embeddings of backchannels/fillers in the LLaMA-3 model change with fine-tuning in the one-context setting. In Figures~\ref{fig:tsne_result1} to \ref{fig:tsne_result10} we provide additional t-SNE visualisations for the other models and settings.
One of the general trends which can be observed is that including more contextual information can lead to better representation of backchannels/fil\-lers (in terms of clearer borders among different ones.). For example, based on English data from Figures~\ref{fig:tsne_result1} and~\ref{fig:tsne_result2} (no-context vs. full-context), Figures~\ref{fig:tsne_result3} and~\ref{fig:tsne_result4} (no-context vs. one-context), it can be concluded that including contextual information in fine-tuning leads to better t-SNE visualisation. However, there also exist exceptions. In the Japanese data in Figures~\ref{fig:tsne_result4} and~\ref{fig:tsne_result5} (one-context vs. full-context), for example, increasing context information for fine-tuning results in poorer t-SNE visualisations.

Another issue that we noticed is that for the BERT models the improvement of the backchannels/fillers representation is not reflected well in the t-SNE visualisations: In comparison to modern LLMs (such as LLaMA-3 and Qwen-3), BERT is much more lightweight and supposedly saw less backchannels/fillers during its pre-training. However, according to Figures~\ref{fig:tsne_result9} and~\ref{fig:tsne_result10}, the border among different backchannels/fillers is already quite clear before fine-tuning. This concern is not further discussed here, and could be studied in future work.

\subsection{Statistics of Silhouette Scores and K-Means Values Under PCA}
\label{app:silhouette}

\begin{figure*}
    \centering
    \includegraphics[width=\linewidth]{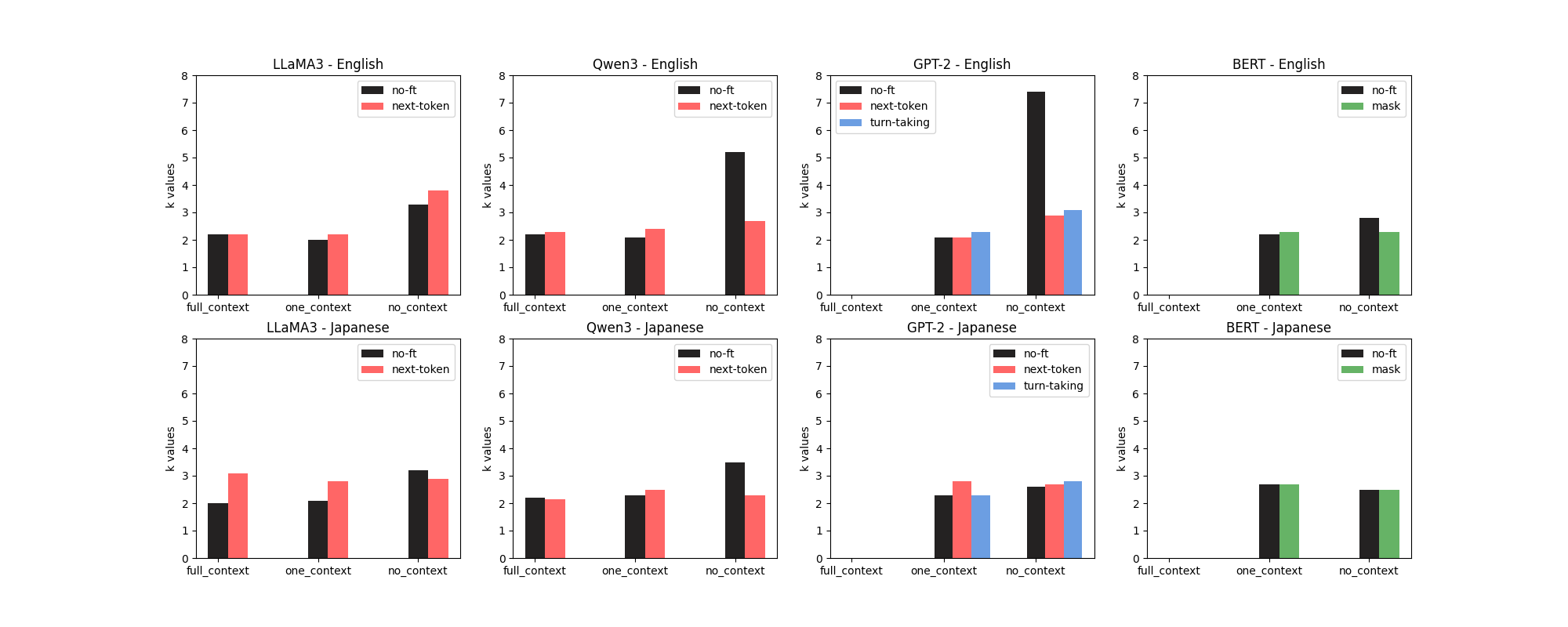}
    \caption{Change in the $k$-value before and after fine-tuning, as extracted from the clustering analysis of different LMs' results. A slight increasing of $k$ after fine-tuning can generally be observed among the different fine-tuning strategies. Exceptions are the no-context setting for the LLaMA-3 model for Japanese and the Qwen-3 model for both languages. It is worth noting that a decrease of the $k$-value does not necessarily indicate failure, but can also mean that the clustering effect is more salient after fine-tuning.}
    \label{fig:k_value_change}
\end{figure*}

To ensure the statistical robustness of our evaluation, we further estimate the average silhouette scores using a bootstrap resampling approach ($n=1000$) \citep{efron1992bootstrap}. As summarized in Table ~\ref{tab:bootstrap_results}, we sample the embeddings with replacement and report the mean silhouette score alongside its 95\% bootstrap confidence interval half-width. Tables~\ref{tab:en_sil1} to~\ref{tab:jap_sil4} respectively show the silhouette scores of each backchannel/filler in both the original dimensional space and after dimensionality reduction to 100 dimensions using Principal Component Analysis (PCA), along with the average silhouette scores for English and Japanese data. A general trend is that, with fine-tuning the silhouette score increases for most of the cases. When there is no fine-tuning, adding context size (no\_ctx to one\_ctx to full\_ctx) will also have the same effect. However, combining context size with fine-tuning as well as increasing context size will lower the silhouette score. 
Additionally, Figure~\ref{fig:k_value_change} and Tables~\ref{tab:en_k1} to~\ref{tab:jap_k4} summarize how the $k$-value changes with fine-tuning.

\subsection{Additional Confusion Matrices}
\label{app:conmatric}

\begin{figure*}[ht]
\centering
\begin{subfigure}[t]{0.27\linewidth}
    \centering
    \includegraphics[width=\linewidth]{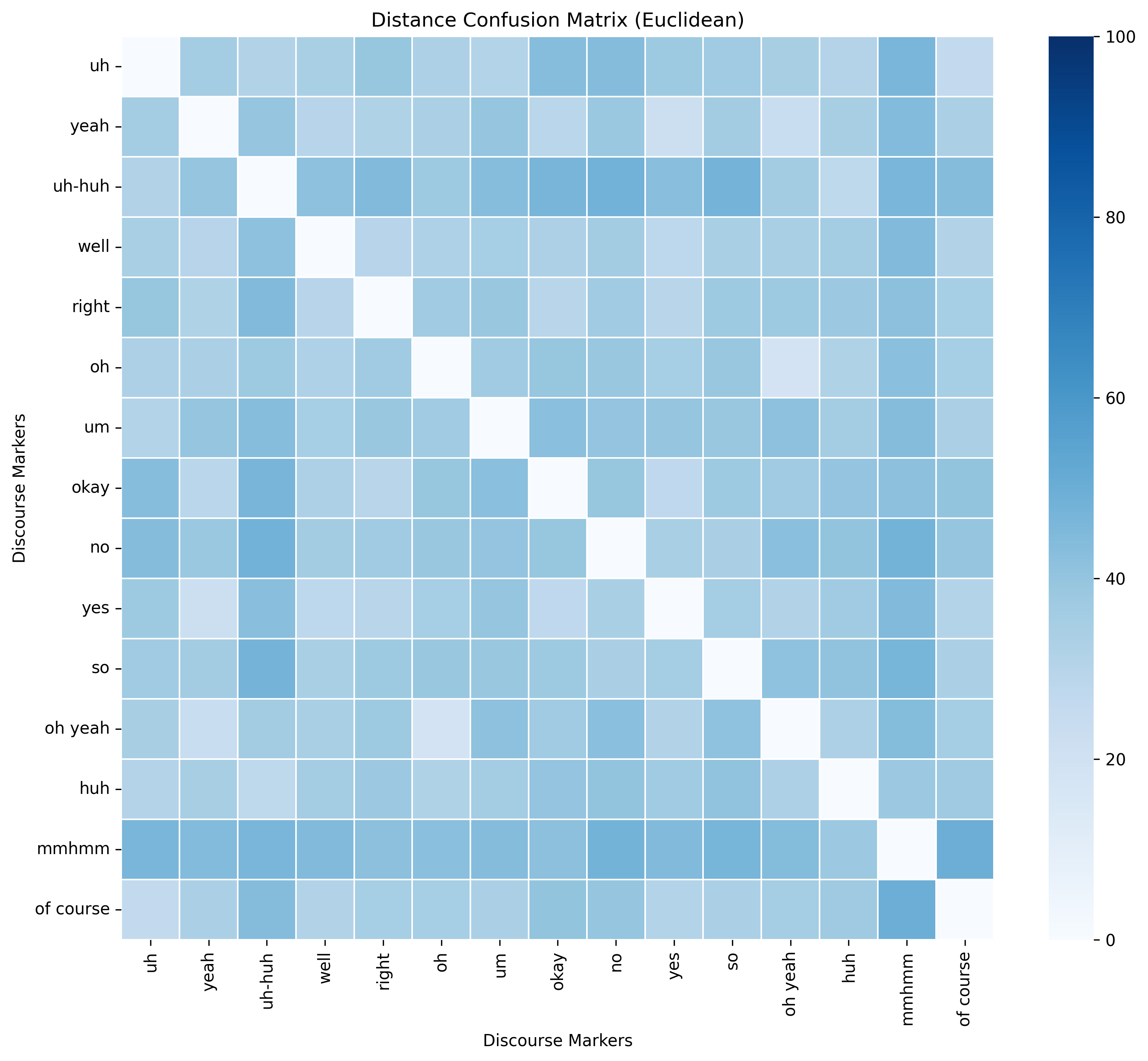}
    \caption{no\_ft, one-context}
    \label{fig:mten1a}
\end{subfigure}
\hspace{2cm}
\begin{subfigure}[t]{0.27\linewidth}
    \centering
    \includegraphics[width=\linewidth]{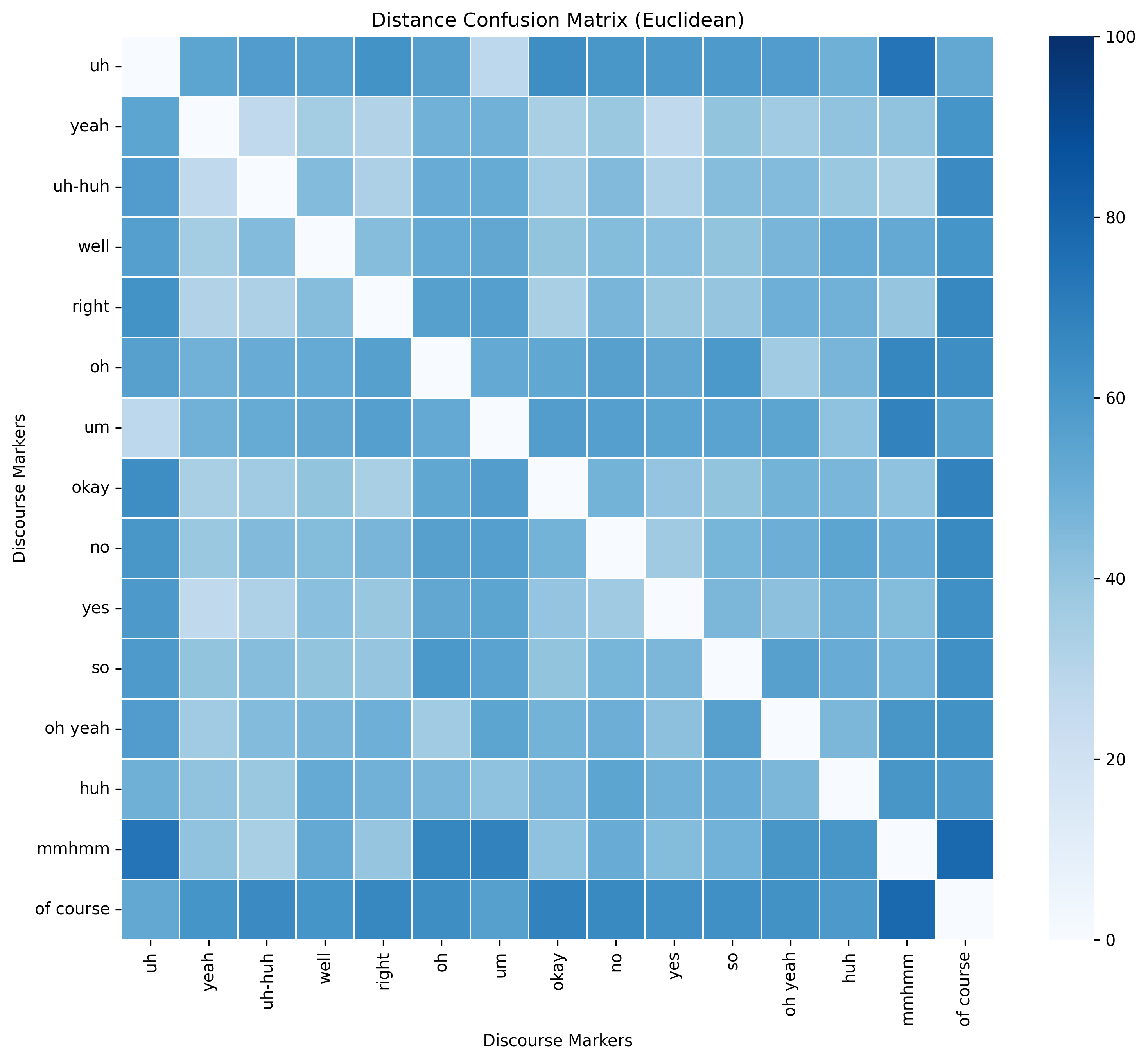}
    \caption{ft, one-context (NTP)}
    \label{fig:mten1b}
\end{subfigure}  
\caption{Distance matrices for the top 15 \textbf{English} backchannels/fillers in the \textbf{LLaMA-3} model (a) before and (b) after fine-tuning.}
\label{fig:mten1}
\end{figure*}

\begin{figure*}[ht]
\centering
\begin{subfigure}[t]{0.27\linewidth}
    \centering
    \includegraphics[width=\linewidth]{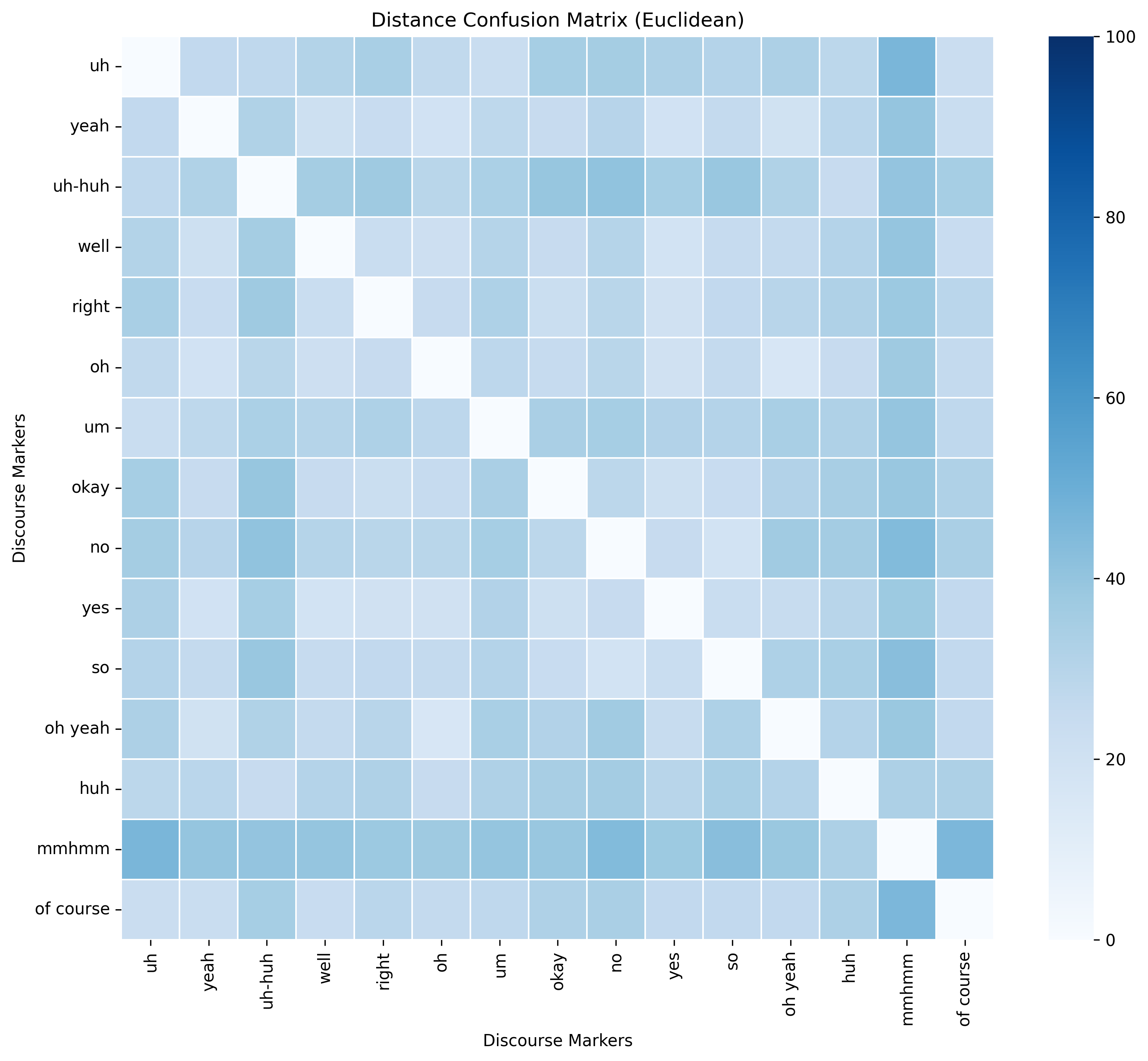}
    \caption{no\_ft, one-context}
    \label{fig:mten2a}
\end{subfigure}
\hspace{2cm}
\begin{subfigure}[t]{0.27\linewidth}
    \centering
    \includegraphics[width=\linewidth]{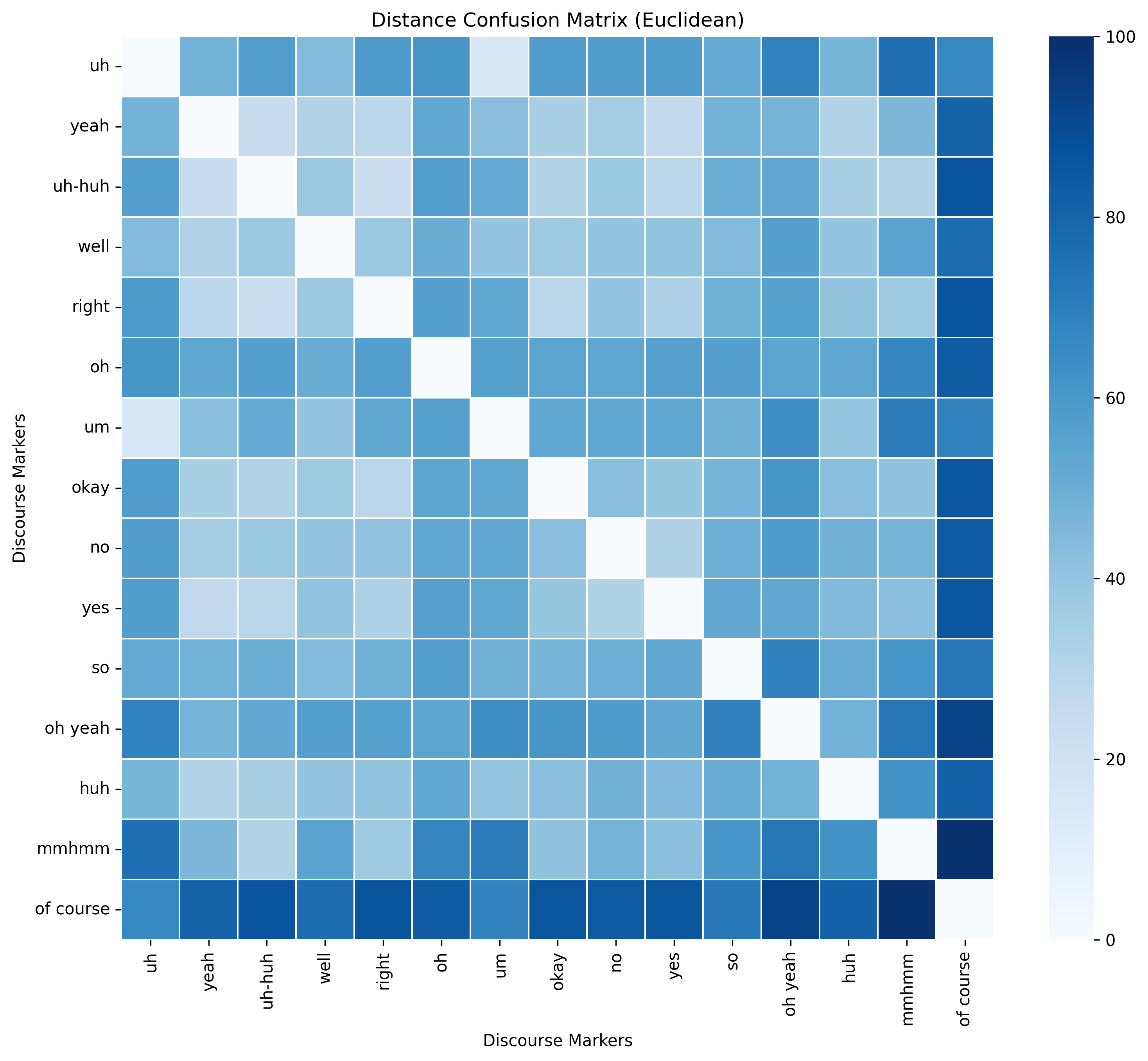}
    \caption{ft, one-context (NTP)}
    \label{fig:mten2b}
\end{subfigure}
\caption{Distance matrices for the top 15 \textbf{English} backchannels/fillers in the \textbf{Qwen-3} model (a) before and (b) after fine-tuning.}
\label{fig:mten2}
\end{figure*}

\begin{figure*}[ht]
\centering
\begin{subfigure}[t]{0.27\linewidth}
    \centering
    \includegraphics[width=\linewidth]{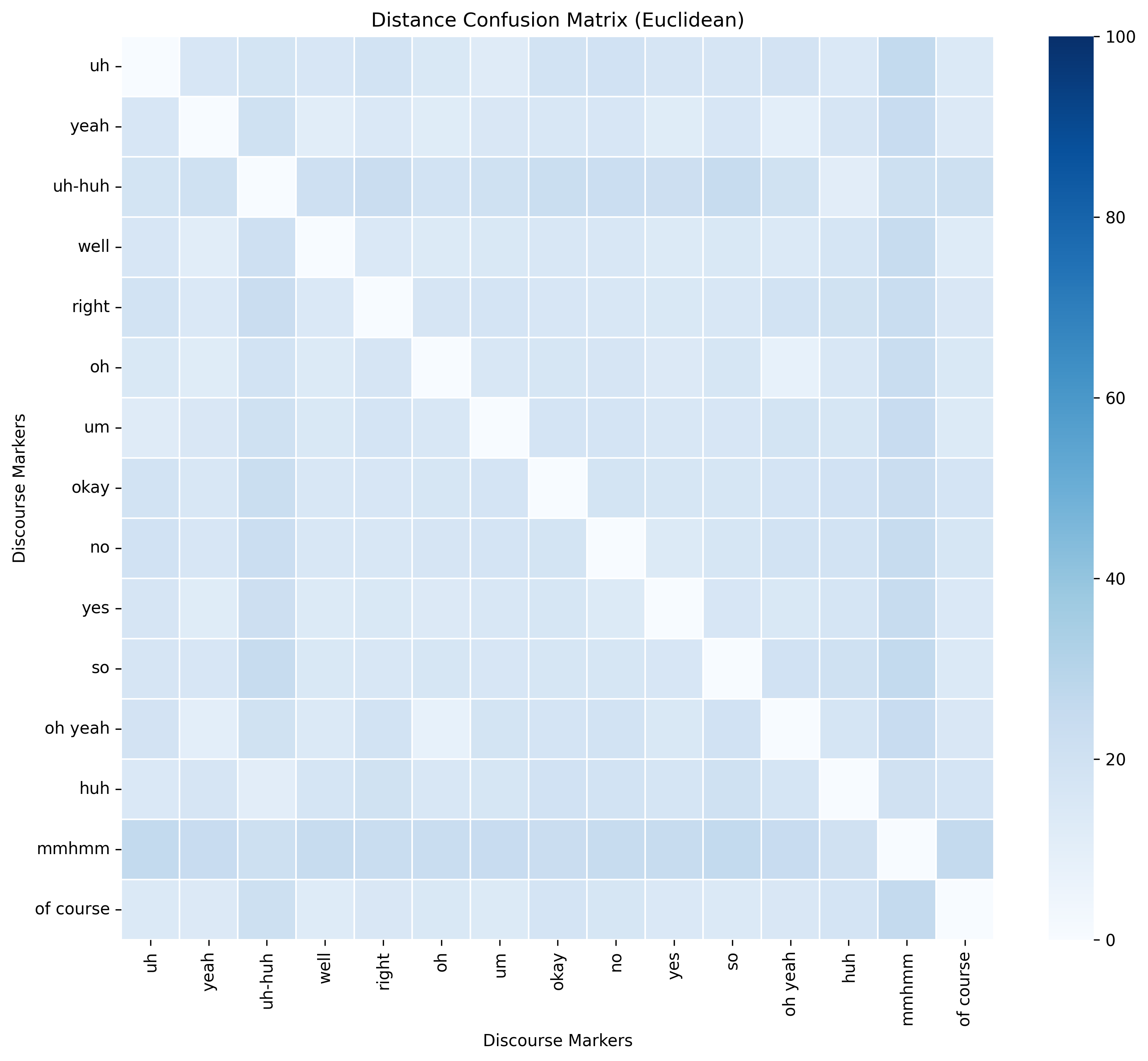}
    \caption{no\_ft, one-context}
    \label{fig:mten3a}
\end{subfigure}
\hspace{2cm}
\begin{subfigure}[t]{0.27\linewidth}
    \centering
    \includegraphics[width=\linewidth]{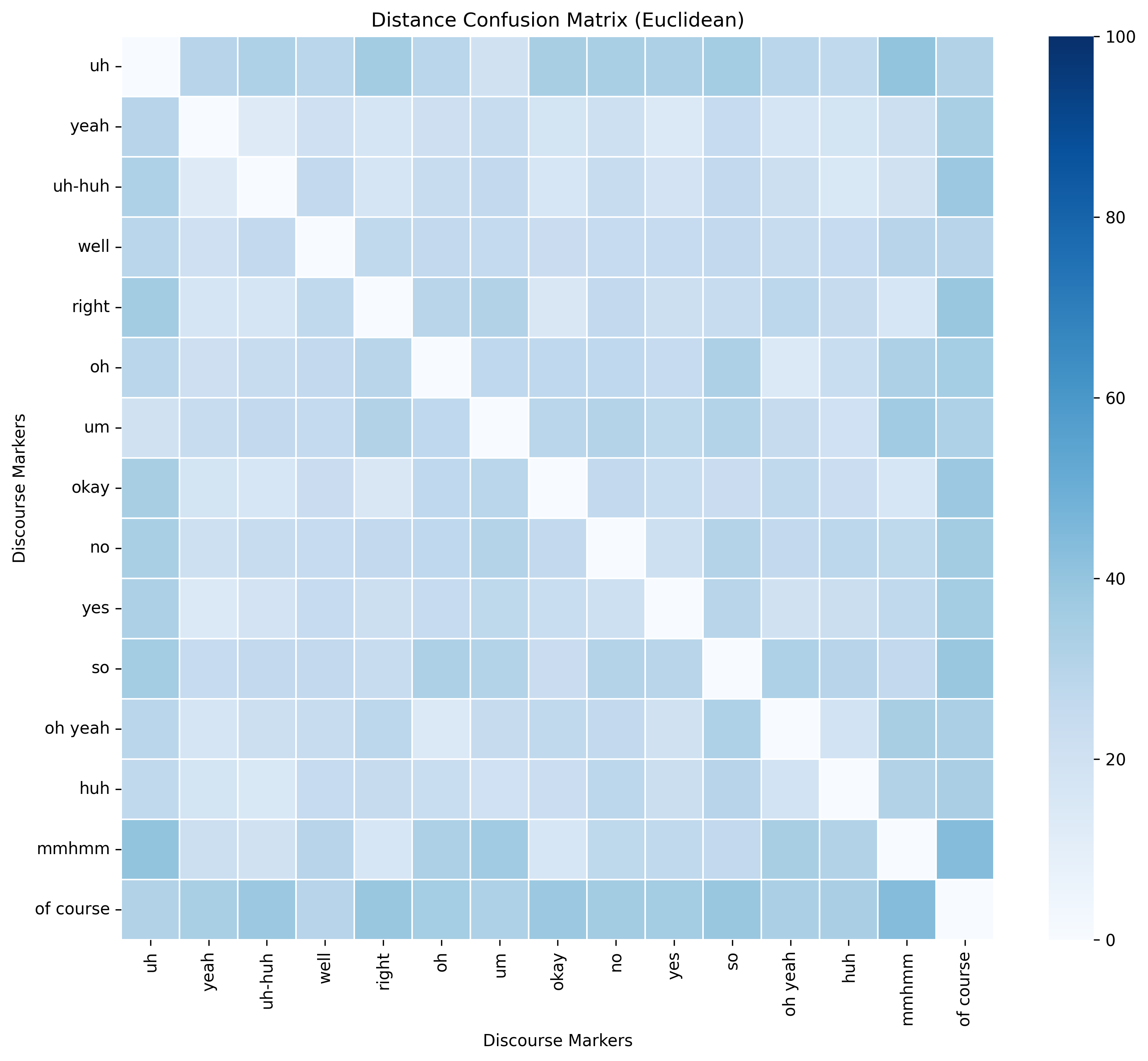}
    \caption{ft, one-context (NTP)}
    \label{fig:mten3b}
\end{subfigure}
\caption{Distance matrices for the top 15 \textbf{English} backchannels/fillers in the \textbf{GPT-2} model (a) before and (b) after fine-tuning.}
\label{fig:mten3}
\end{figure*}

\begin{figure*}[ht]
\centering
\begin{subfigure}[t]{0.27\linewidth}
    \centering
    \includegraphics[width=\linewidth]{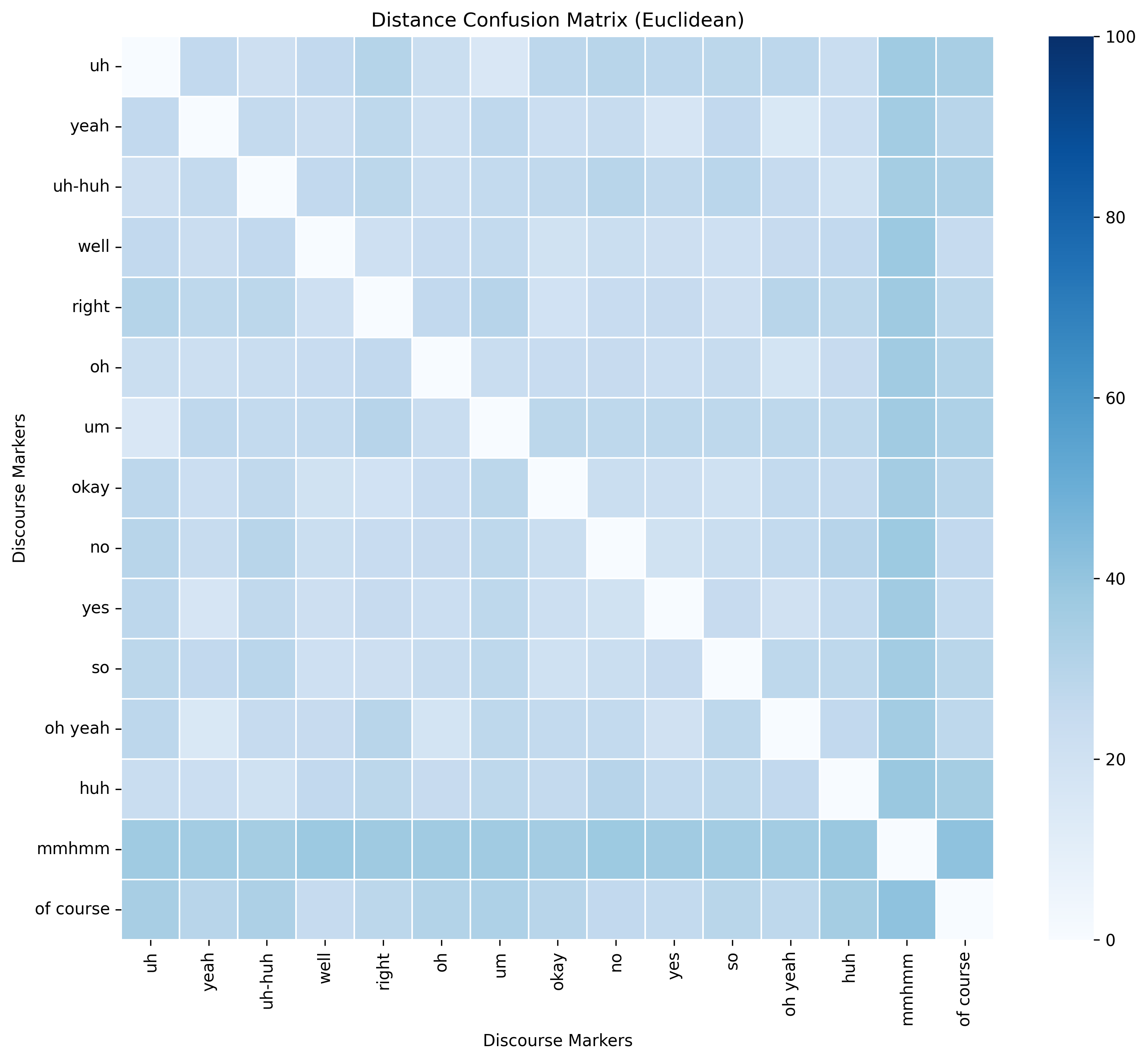}
    \caption{no\_ft, one-context}
    \label{fig:mten4a}
\end{subfigure}
\hspace{2cm}
\begin{subfigure}[t]{0.27\linewidth}
    \centering
    \includegraphics[width=\linewidth]{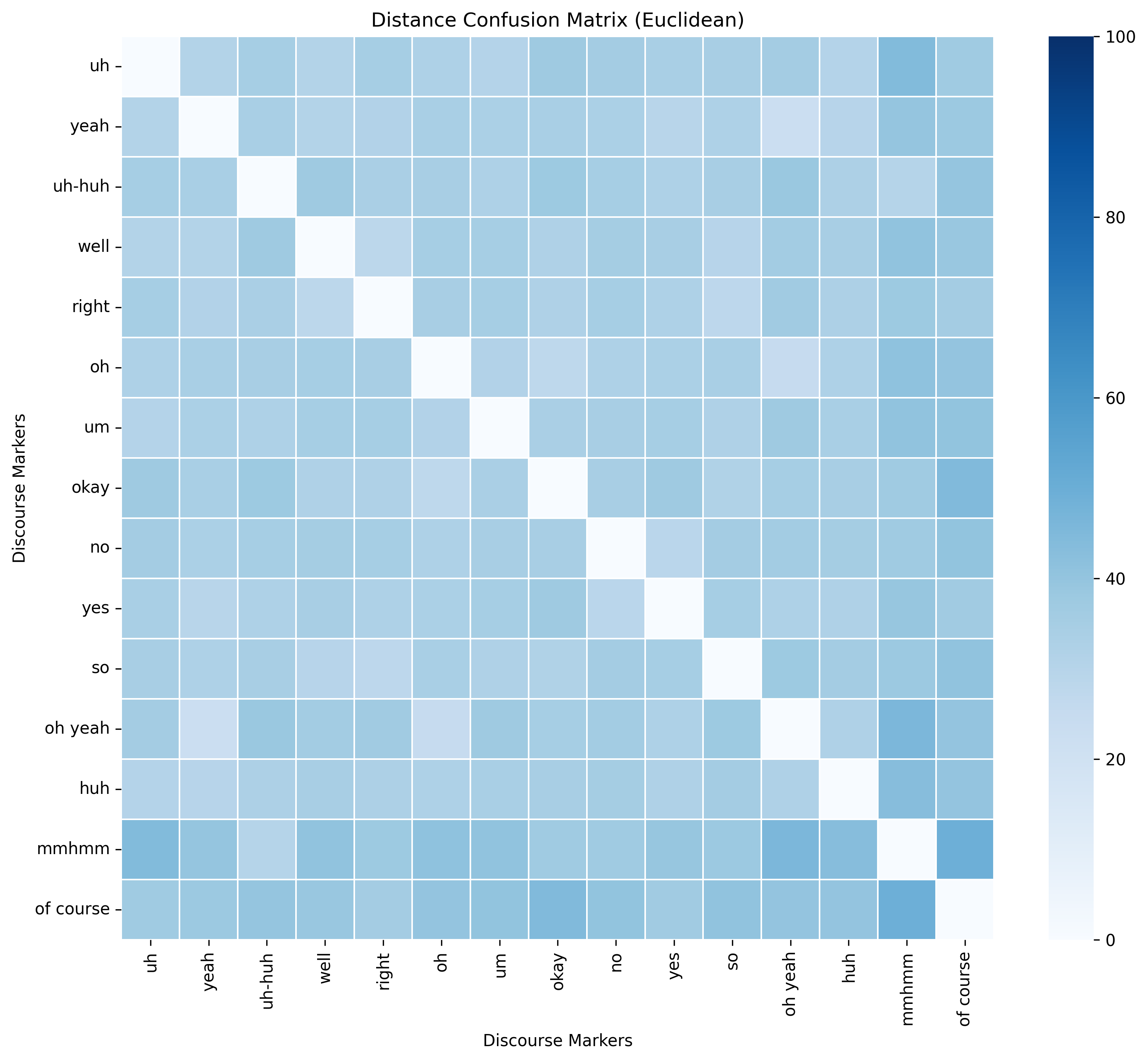}
    \caption{ft, one-context (MASK)}
    \label{fig:mten4b}
\end{subfigure}
\caption{Distance matrices for the top 15 \textbf{English} backchannels/fillers in the \textbf{BERT} model (a) before and (b) after fine-tuning.}
\label{fig:mten4}
\end{figure*}

\begin{figure*}[ht]
\centering
\begin{subfigure}[t]{0.27\linewidth}
    \centering
    \includegraphics[width=\linewidth]{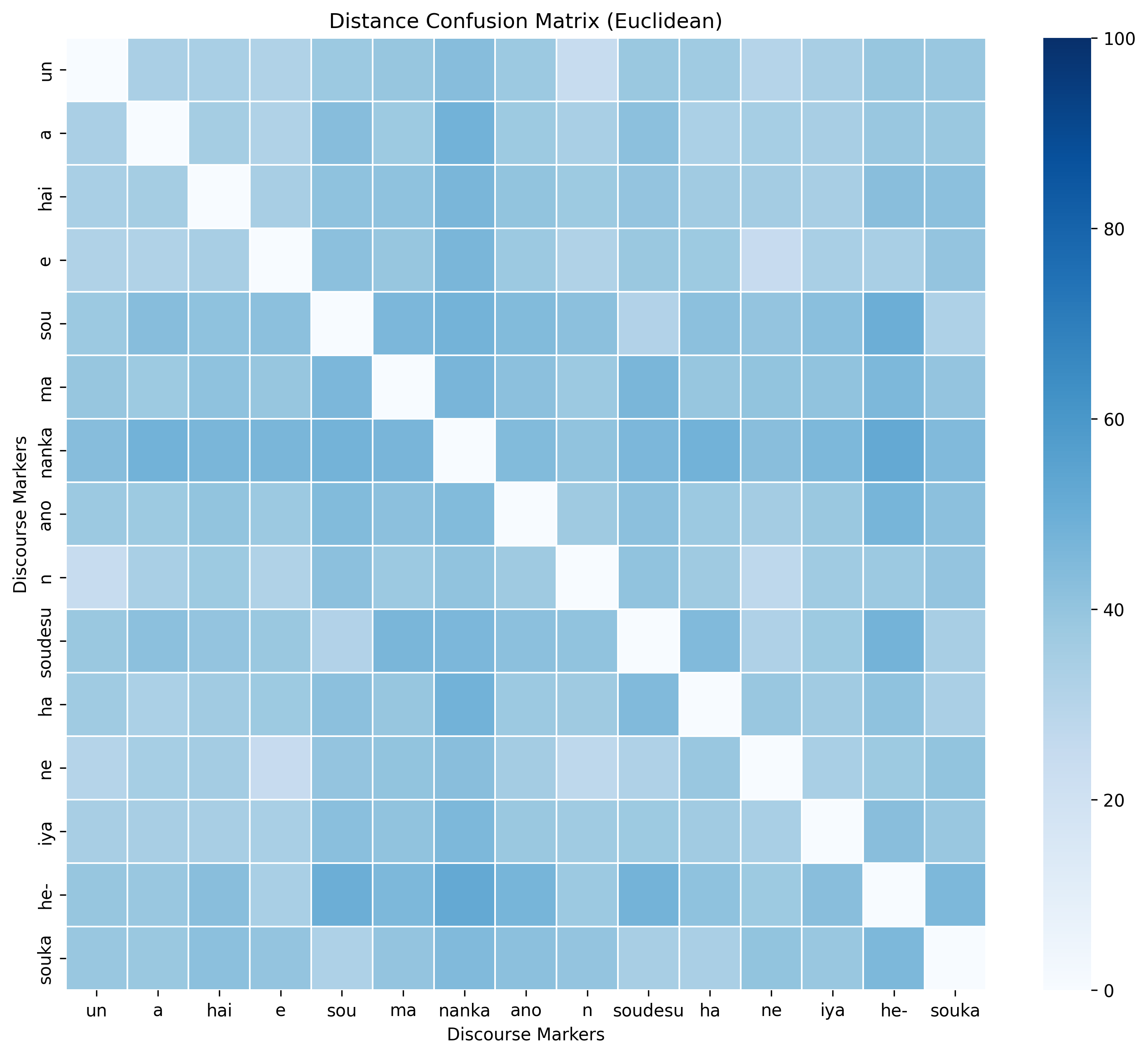}
    \caption{no\_ft, one-context}
    \label{fig:mtja1a}
\end{subfigure}
\hspace{2cm}
\begin{subfigure}[t]{0.27\linewidth}
    \centering
    \includegraphics[width=\linewidth]{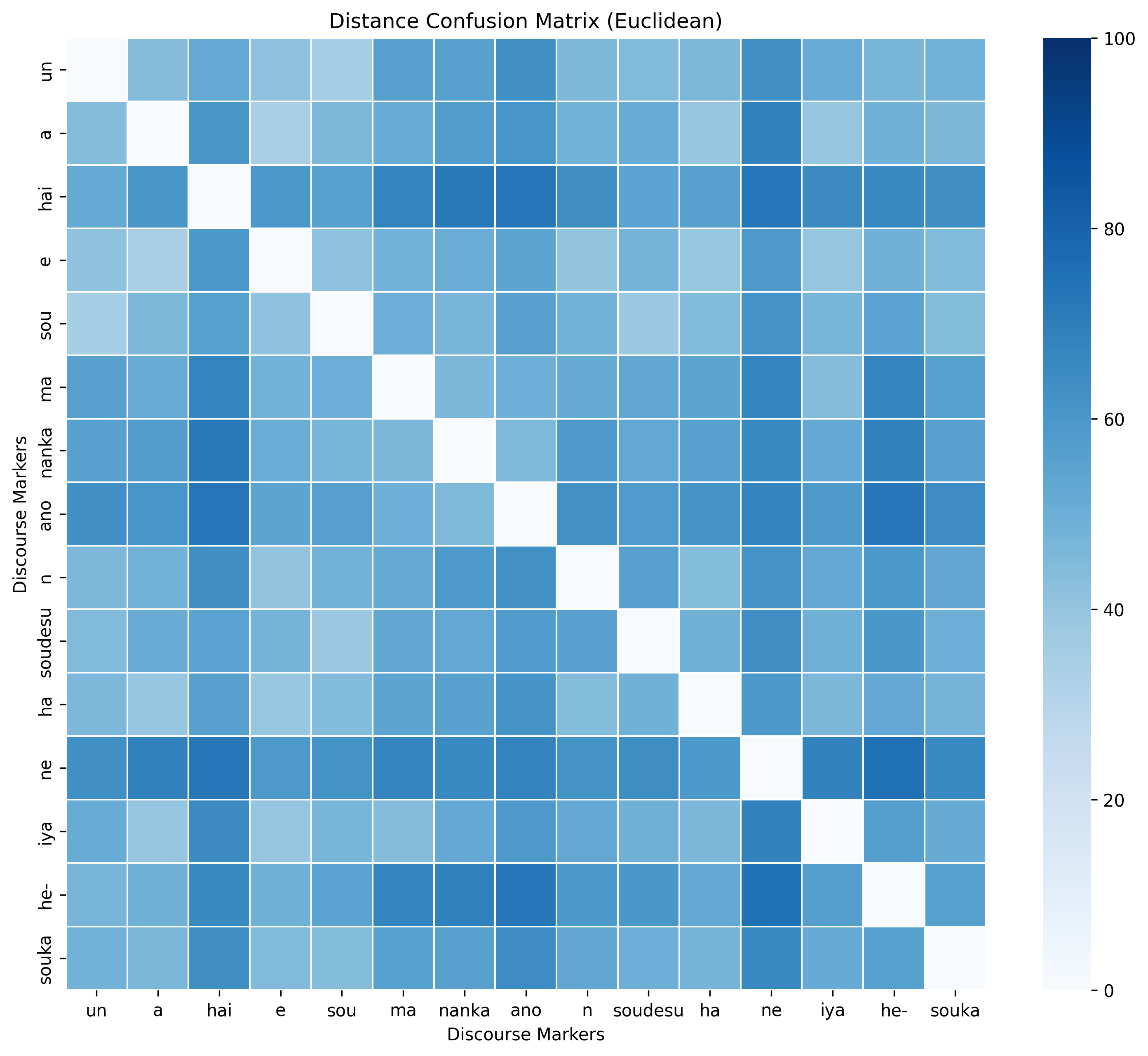}
    \caption{ft, one-context (NTP)}
    \label{fig:mtja1b}
\end{subfigure}
\caption{Distance matrices for the top 15 \textbf{Japanese} backchannels/fillers in the \textbf{LLaMA-3} model (a)~before and (b)~after fine-tuning.}
\label{fig:mtja1}
\end{figure*}

\begin{figure*}[ht]
\centering
\begin{subfigure}[t]{0.27\linewidth}
    \centering
    \includegraphics[width=\linewidth]{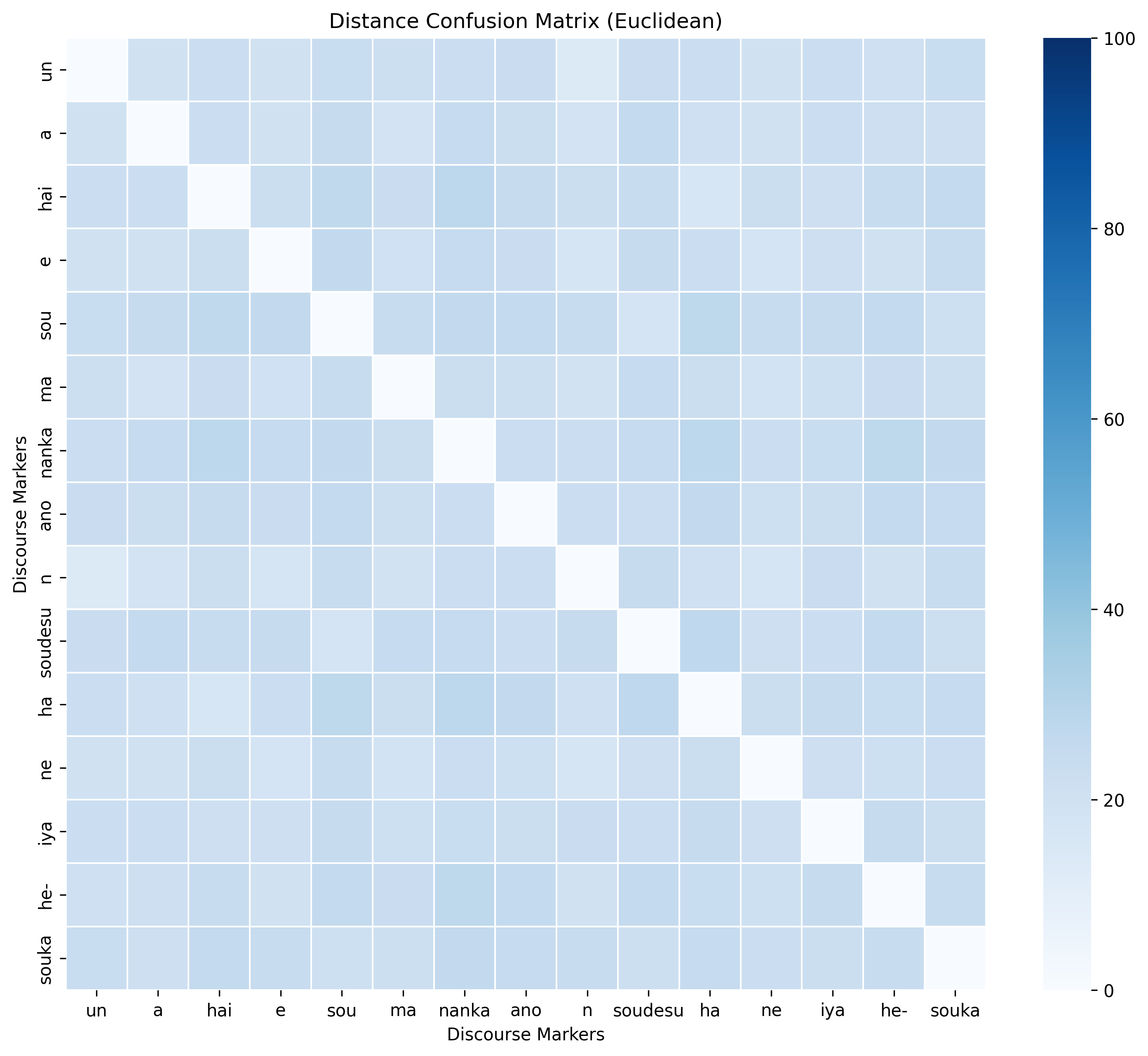}
    \caption{no\_ft, one-context}
    \label{fig:mtja3a}
\end{subfigure}
\hspace{2cm}
\begin{subfigure}[t]{0.27\linewidth}
    \centering
    \includegraphics[width=\linewidth]{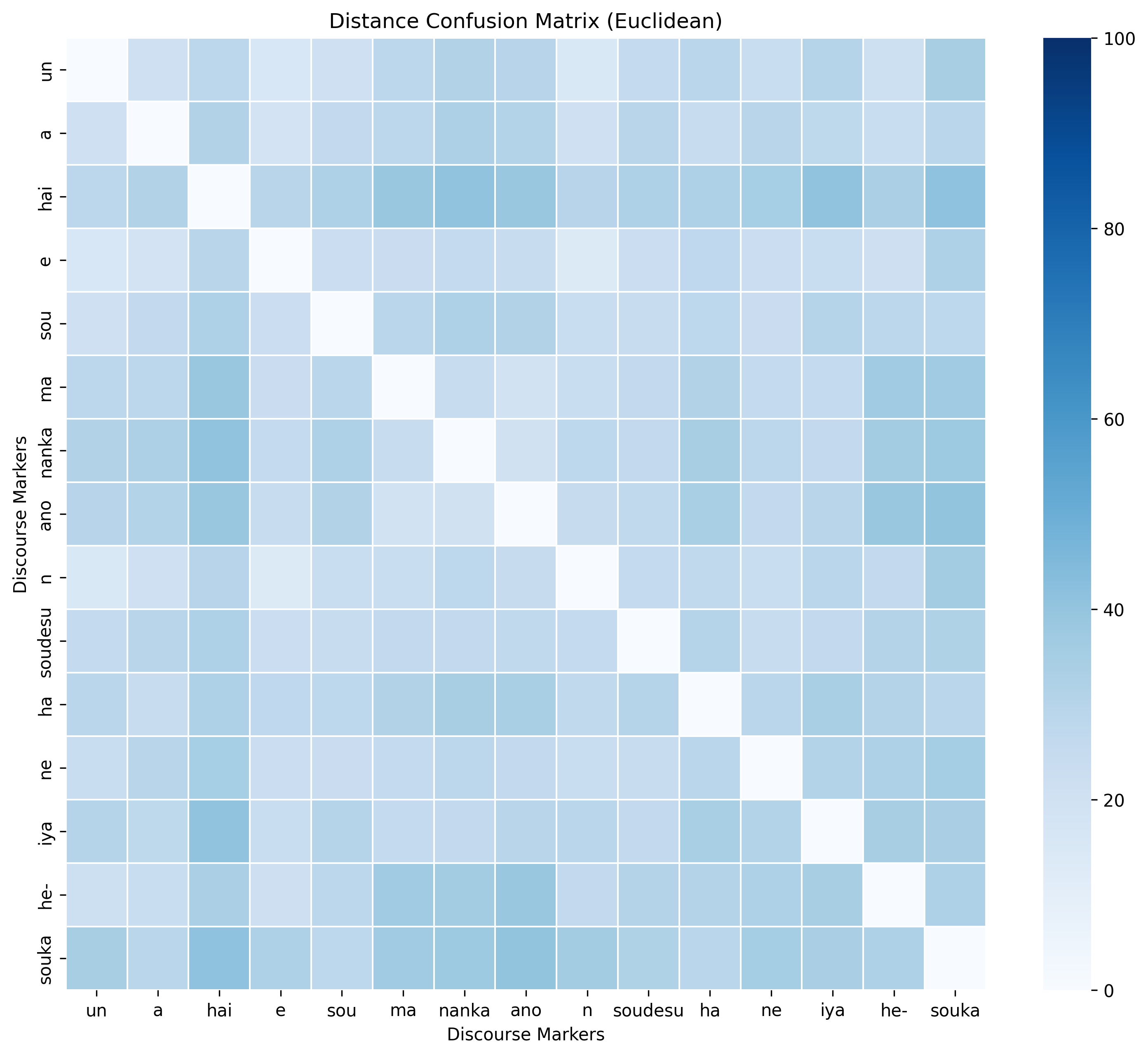}
    \caption{ft, one-context (NTP)}
    \label{fig:mtja3b}
\end{subfigure}
\caption{Distance matrices for the top 15 \textbf{Japanese} backchannels/fillers in the \textbf{GPT-2} model (a)~before and (b)~after fine-tuning.}
\label{fig:mtja3}
\end{figure*}

\begin{figure*}[t]
\centering
\begin{subfigure}[t]{0.27\linewidth}
    \centering
    \includegraphics[width=\linewidth]{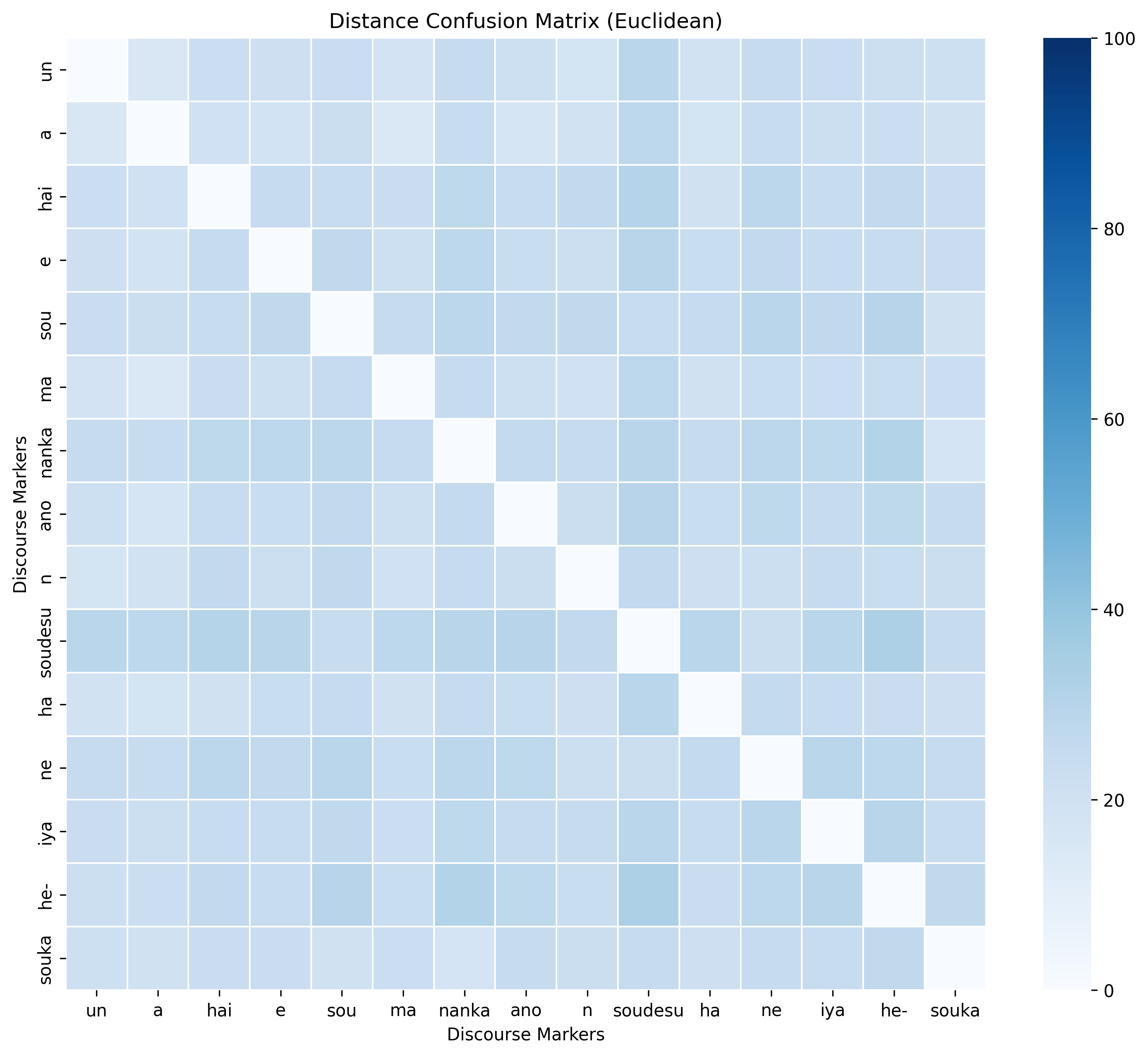}
    \caption{no\_ft, one-context}
    \label{fig:mtja4a}
\end{subfigure}
\hspace{2cm}
\begin{subfigure}[t]{0.27\linewidth}
    \centering
    \includegraphics[width=\linewidth]{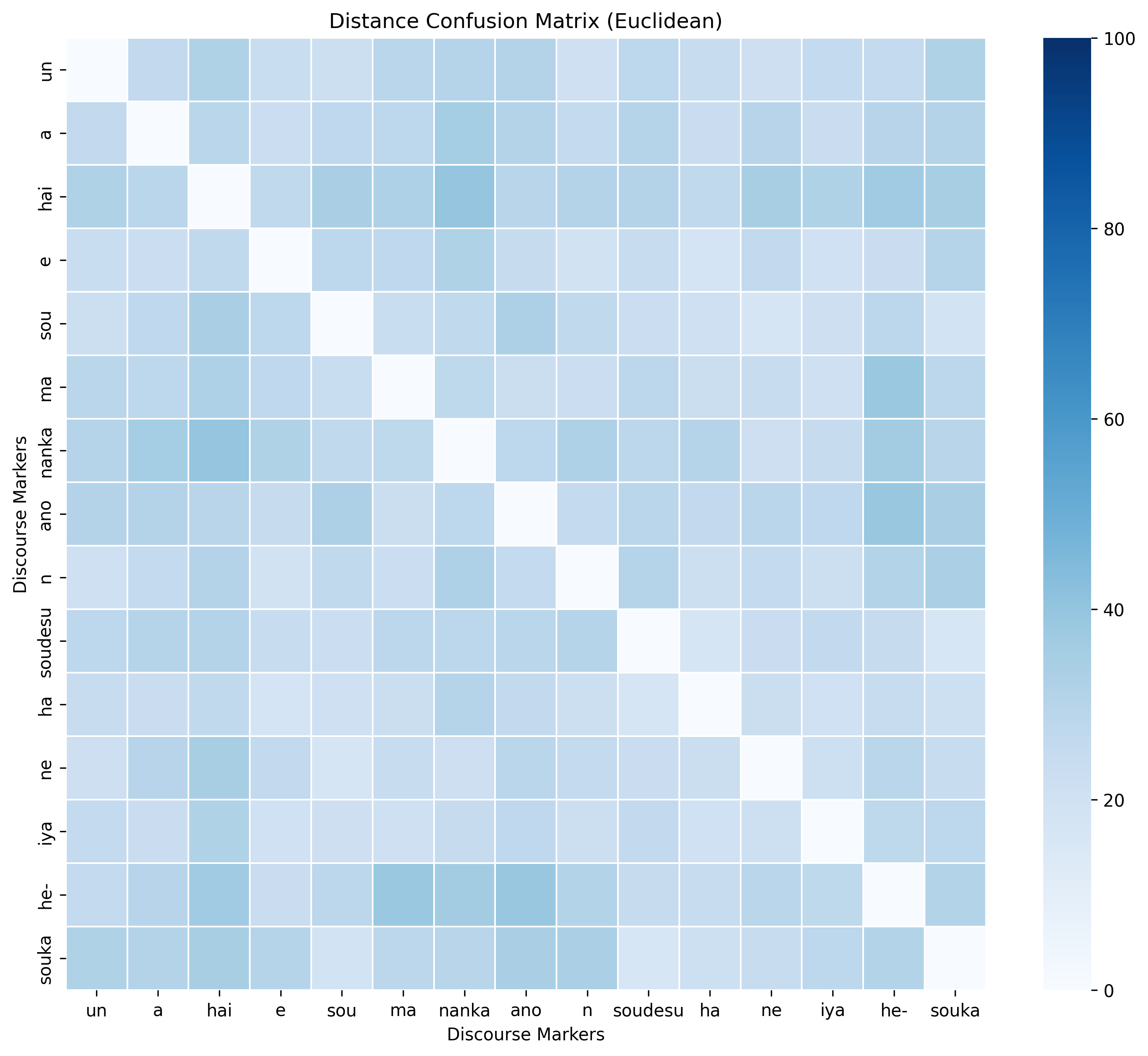}
    \caption{ft, one-context (MASK)}
    \label{fig:mtja4b}
\end{subfigure}
\caption{Distance matrices for the top 15 \textbf{Japanese} backchannels/fillers in the \textbf{BERT} model (a)~before and (b)~after fine-tuning.}
\label{fig:mtja4}
\end{figure*}

Figure~\ref{fig:mtja2} in Section~\ref{sec:res} shows the confusion matrices of distances among the top 15 Japanese backchannels/fillers in the Qwen-3 model before and after fine-tuning. Here, we provide additional confusion matrices for English (LLaMA-3, Qwen-3, GPT-2, and BERT) in Figures~\ref{fig:mten1} to \ref{fig:mten4} and Japanese (LLaMA-3, GPT-2, and BERT) in Figures~\ref{fig:mtja1} to \ref{fig:mtja4}.

In the confusion matrices, darker colours indicate greater distances and lighter colours indicate greater similarity. The confusion matrices for English and Japanese show that for the more recent LLM models (LLaMA-3 and Qwen-3), fine-tuning makes the differences among selected backchannels/fillers more pronounced. Change is less evident for GPT-2 and BERT.


\FloatBarrier 
\clearpage

\begin{table*}[t]
\footnotesize
\caption{Evaluation results across different models and tasks. Results are reported as mean $\pm$ 95\% bootstrap confidence interval half-width. `Base' denotes the original model without fine-tuning, and `FT' denotes the fine-tuned model. Full-context evaluation is exclusive to LLaMA-3 and Qwen-3.}
\resizebox{\linewidth}{!}{
\begin{tabular}{lllcccccc}
\toprule
\textbf{Model} & \textbf{Task} & \textbf{Lang} & \multicolumn{2}{c}{\textbf{No-Context}} & \multicolumn{2}{c}{\textbf{One-Context}} & \multicolumn{2}{c}{\textbf{Full-Context}} \\
\cmidrule(lr){4-5} \cmidrule(lr){6-7} \cmidrule(lr){8-9}
& & & Base & FT & Base & FT & Base & FT \\
\midrule
\multirow{2}{*}{BERT} & \multirow{2}{*}{MASK} & EN & 0.144 $\pm$ 0.011 & 0.241 $\pm$ 0.008 & 0.213 $\pm$ 0.008 & 0.391 $\pm$ 0.017 & --- & --- \\
& & JP & 0.213 $\pm$ 0.008 & 0.391 $\pm$ 0.017 & 0.197 $\pm$ 0.004 & 0.429 $\pm$ 0.012 & --- & --- \\
\midrule
\multirow{4}{*}{GPT-2} & \multirow{2}{*}{NTP} & EN & 0.274 $\pm$ 0.012 & 0.328 $\pm$ 0.021 & 0.149 $\pm$ 0.002 & 0.311 $\pm$ 0.020 & --- & --- \\
& & JP & 0.157 $\pm$ 0.004 & 0.288 $\pm$ 0.010 & 0.101 $\pm$ 0.003 & 0.273 $\pm$ 0.008 & --- & --- \\
\cmidrule{2-9}
& \multirow{2}{*}{TTP} & EN & --- & 0.289 $\pm$ 0.019 & --- & 0.211 $\pm$ 0.018 & --- & --- \\
& & JP & --- & 0.284 $\pm$ 0.014 & --- & 0.261 $\pm$ 0.014 & --- & --- \\
\midrule
\multirow{2}{*}{LLaMA-3} & \multirow{2}{*}{NTP} & EN & 0.450 $\pm$ 0.009 & 0.588 $\pm$ 0.011 & 0.183 $\pm$ 0.002 & 0.291 $\pm$ 0.027 & 0.210 $\pm$ 0.020 & 0.301 $\pm$ 0.018 \\
& & JP & 0.257 $\pm$ 0.009 & 0.450 $\pm$ 0.014 & 0.179 $\pm$ 0.003 & 0.335 $\pm$ 0.010 & 0.318 $\pm$ 0.021 & 0.408 $\pm$ 0.037 \\
\midrule
\multirow{2}{*}{Qwen-3} & \multirow{2}{*}{NTP} & EN & 0.253 $\pm$ 0.013 & 0.379 $\pm$ 0.065 & 0.157 $\pm$ 0.003 & 0.292 $\pm$ 0.021 & 0.189 $\pm$ 0.020 & 0.322 $\pm$ 0.028 \\
& & JP & 0.172 $\pm$ 0.011 & 0.452 $\pm$ 0.068 & 0.154 $\pm$ 0.002 & 0.263 $\pm$ 0.014 & 0.173 $\pm$ 0.004 & 0.181 $\pm$ 0.012 \\
\bottomrule
\end{tabular}
} 
\label{tab:bootstrap_results}
\end{table*}

\begin{table*}[ht]
\scriptsize
\caption{%
    Maximum silhouette scores on the $k$-means clustering of the top 15 selected \textbf{English} backchannels/fillers when using the \textbf{LLaMA-3} model in both the original dimensional space and after dimensionality reduction to 100 dimensions using PCA
    (\textbf{no\_ft}: no fine-tuning; \textbf{ft}: fine-tuning; Settings: \textbf{no\_ctx}: no-context; \textbf{one\_ctx}: one-context; \textbf{full\_ctx}: full-context)}
\begin{tabularx}{\textwidth}{lXXXXXXXXXXXX}
\toprule
    \textbf{Backchannel/Filler} & \multicolumn{2}{c}{\textbf{no\_ft, no\_ctx}} & \multicolumn{2}{c}{\textbf{ft, no\_ctx (NTP)}}  & \multicolumn{2}{c}{\textbf{no\_ft, one\_ctx}} & \multicolumn{2}{c}{\textbf{ft, one\_ctx (NTP)}}  & \multicolumn{2}{c}{\textbf{no\_ftm full\_ctx}} & \multicolumn{2}{c}{\textbf{ft, full\_ctx (NTP)}}  \\
    \cmidrule(lr){2-3} \cmidrule(lr){4-5} \cmidrule(lr){6-7} \cmidrule(lr){8-9} \cmidrule(lr){10-11} \cmidrule(lr){12-13}
     & \textbf{orig.} & \textbf{100} & \textbf{orig.} & \textbf{100} & \textbf{orig.} & \textbf{100} & \textbf{orig.} & \textbf{100} & \textbf{orig.} & \textbf{100} & \textbf{orig.} & \textbf{100} \\
\midrule
	\textit{uh} &0.302 &0.355 &0.470 &0.500 &0.090 &0.114 &0.057 &0.097 &0.052 &0.233 &0.059 &0.095\\
    \textit{yeah} &0.418 &0.423 &0.565 &0.579 &0.195 &0.205 &0.319 &0.387 &0.084 &0.194 &0.318 &0.392\\
    \textit{uh-huh} &0.470 &0.490 &0.606 &0.623 &0.232 &0.191 &0.363 &0.431 &0.147 &0.222 &0.372 &0.437\\
    \textit{well} &0.403 &0.417 &0.519 &0.479 &0.226 &0.216 &0.186 &0.349 &0.227 &0.275 &0.292 &0.359\\
    \textit{right} &0.403 &0.433 &0.596 &0.572 &0.211 &0.206 &0.402 &0.496 &0.182 &0.202 &0.410 &0.506\\
    \textit{oh} &0.337 &0.364 &0.506 &0.513 &0.205 &0.206 &0.239 &0.279 &0.121 &0.182 &0.226 &0.278\\
    \textit{um} &0.335 &0.374 &0.482 &0.516 &0.109 &0.127 &0.076 &0.149 &0.054 &0.102 &0.074 &0.144\\
    \textit{okay} &0.420 &0.425 &0.583 &0.563 &0.170 &0.203 &0.323 &0.391 &0.170 &0.295 &0.346 &0.427\\
    \textit{no} &0.424 &0.438 &0.591 &0.580 &0.207 &0.229 &0.262 &0.363 &0.128 &0.167 &0.294 &0.415\\
    \textit{yes} &0.397 &0.422 &0.600 &0.598 &0.159 &0.191 &0.267 &0.358 &0.155 &0.280 &0.276 &0.372\\
    \textit{so} &0.332 &0.339 &0.538 &0.579 &0.163 &0.142 &0.376 &0.477 &0.101 &0.120 &0.386 &0.485\\
    \textit{oh yeah} &0.525 &0.359 &0.676 &0.620 &0.183 &0.192 &0.089 &0.141 &0.154 &0.208 &0.078 &0.135\\
    \textit{huh} &0.434 &0.461 &0.590 &0.632 &0.186 &0.184 &0.129 &0.161 &0.099 &0.153 &0.107 &0.137\\
    \textit{mmhmm} &0.559 &0.505 &0.661 &0.690 &0.232 &0.230 &0.123 &0.247 &0.306 &0.362 &0.103 &0.217\\
    \textit{of course} &0.350 &0.376 &0.522 &0.580 &0.081 &0.091 &0.060 &0.102 &0.034 &0.079 &0.048 &0.087\\ 
\midrule
    \textbf{Average} \textuparrow &\textbf{0.407} &\textbf{0.424} &\textbf{0.567} &\textbf{0.575} &\textbf{0.177} &\textbf{0.182} &\textbf{0.218} &\textbf{0.295} &\textbf{0.134} &\textbf{0.205} &\textbf{0.226} &\textbf{0.299} \\ 
\bottomrule
\end{tabularx}
\label{tab:en_sil1}
\end{table*}

\begin{table*}[ht]
\scriptsize
\centering
\caption{%
    Maximum silhouette scores value on the $k$-means clustering of the top 15 selected \textbf{English} backchannels/fillers when using the \textbf{Qwen-3} model in both the original dimensional space and after dimensionality reduction to 100 dimensions using PCA.}
\begin{tabularx}{\textwidth}{lXXXXXXXXXXXX}
\toprule
    \textbf{Backchannel/Filler} & \multicolumn{2}{c}{\textbf{no\_ft, no\_ctx}} & \multicolumn{2}{c}{\textbf{ft, no\_ctx (NTP)}} & \multicolumn{2}{c}{\textbf{no\_ft, one\_ctx}} & \multicolumn{2}{c}{\textbf{ft, one\_ctx (NTP)}} & \multicolumn{2}{c}{\textbf{no\_ft, full\_ctx}} & \multicolumn{2}{c}{\textbf{ft, full\_ctx (NTP)}} \\
\cmidrule(lr){2-3} \cmidrule(lr){4-5} \cmidrule(lr){6-7} \cmidrule(lr){8-9} \cmidrule(lr){10-11} \cmidrule(lr){12-13}
     & \textbf{orig.} & \textbf{100} & \textbf{orig.} & \textbf{100} & \textbf{orig.} & \textbf{100} & \textbf{orig.} & \textbf{100} & \textbf{orig.} & \textbf{100} & \textbf{orig.} & \textbf{100} \\
\midrule
    \textit{uh} & 0.333 & 0.172 & 0.126 & 0.106 & 0.332 & 0.144 & 0.119 & 0.155 & 0.286 & 0.107 & 0.153 & 0.174 \\ 
    \textit{yeah} & 0.396 & 0.252 & 0.222 & 0.231 & 0.312 & 0.192 & 0.339 & 0.203 & 0.220 & 0.146 & 0.391 & 0.197 \\ 
    \textit{uh-huh} & 0.376 & 0.224 & 0.265 & 0.286 & 0.320 & 0.199 & 0.393 & 0.445 & 0.170 & 0.135 & 0.444 & 0.471 \\ 
    \textit{well} & 0.287 & 0.143 & 0.847 & 0.236 & 0.243 & 0.129 & 0.319 & 0.237 & 0.442 & 0.217 & 0.382 & 0.221 \\ 
    \textit{right} & 0.322 & 0.199 & 0.837 & 0.266 & 0.353 & 0.185 & 0.427 & 0.476 & 0.224 & 0.270 & 0.483 & 0.514 \\ 
    \textit{oh} & 0.262 & 0.134 & 0.202 & 0.234 & 0.294 & 0.135 & 0.269 & 0.317 & 0.215 & 0.122 & 0.337 & 0.301 \\ 
    \textit{um} & 0.316 & 0.156 & 0.147 & 0.132 & 0.295 & 0.111 & 0.178 & 0.222 & 0.298 & 0.130 & 0.202 & 0.226 \\ 
    \textit{okay} & 0.349 & 0.176 & 0.281 & 0.345 & 0.304 & 0.142 & 0.335 & 0.382 & 0.482 & 0.260 & 0.403 & 0.445 \\ 
    \textit{no} & 0.358 & 0.204 & 0.204 & 0.272 & 0.395 & 0.200 & 0.279 & 0.340 & 0.208 & 0.221 & 0.356 & 0.418 \\ 
    \textit{yes} & 0.291 & 0.207 & 0.825 & 0.278 & 0.313 & 0.132 & 0.236 & 0.354 & 0.521 & 0.209 & 0.351 & 0.219 \\ 
    \textit{so} & 0.322 & 0.163 & 0.815 & 0.266 & 0.238 & 0.136 & 0.257 & 0.291 & 0.253 & 0.253 & 0.461 & 0.490 \\ 
    \textit{oh} yeah & 0.417 & 0.478 & 0.909 & 0.742 & 0.295 & 0.154 & 0.083 & 0.128 & 0.198 & 0.110 & 0.075 & 0.128 \\ 
    \textit{huh} & 0.356 & 0.255 & 0.725 & 0.676 & 0.262 & 0.151 & 0.162 & 0.189 & 0.329 & 0.145 & 0.131 & 0.152 \\ 
    \textit{mmhmm} & 0.389 & 0.412 & 0.614 & 0.635 & 0.336 & 0.189 & 0.106 & 0.186 & 0.640 & 0.216 & 0.184 & 0.201 \\ 
    \textit{of course} & 0.272 & 0.155 & 0.312 & 0.185 & 0.274 & 0.130 & 0.202 & 0.213 & 0.185 & 0.099 & 0.199 & 0.180 \\ 
\midrule
    \textbf{Average \textuparrow} & \textbf{0.336} & \textbf{0.222} & \textbf{0.489} & \textbf{0.326} & \textbf{0.304} & \textbf{0.155} & \textbf{0.247} & \textbf{0.276} & \textbf{0.311} & \textbf{0.176} & \textbf{0.304} & \textbf{0.289} \\ 
\bottomrule
\end{tabularx}
\label{tab:en_sil2}
\end{table*}

\begin{table*}[ht]
\scriptsize
\caption{%
    Maximum silhouette scores value on the $k$-means clustering of the top 15 selected \textbf{English} backchannels/fillers when using the \textbf{GPT-2} model in both the original dimensional space and after dimensionality reduction to 100 dimensions using PCA.}
\begin{tabularx}{\textwidth}{lXXXXXXXXXXXX}
\toprule
    \textbf{Backchannel/Filler} & \multicolumn{2}{c}{\textbf{no\_ft, no\_ctx}} & \multicolumn{2}{c}{\textbf{ft, no\_ctx (NTP)}} & \multicolumn{2}{c}{\textbf{ft, no\_ctx (TTP)}}   & \multicolumn{2}{c}{\textbf{no\_ft, one\_ctx}} & \multicolumn{2}{c}{\textbf{ft, one\_ctx (NTP)}} & \multicolumn{2}{c}{\textbf{ft, one\_ctx (TTP)}} \\
\cmidrule(lr){2-3} \cmidrule(lr){4-5} \cmidrule(lr){6-7} \cmidrule(lr){8-9} \cmidrule(lr){10-11} \cmidrule(lr){12-13}
     & \textbf{orig.} & \textbf{100} & \textbf{orig.} & \textbf{100} & \textbf{orig.} & \textbf{100} & \textbf{orig.} & \textbf{100} & \textbf{orig.} & \textbf{100} & \textbf{orig.} & \textbf{100} \\
\midrule
    \textit{uh} &0.355 &0.168 &0.443 &0.068 &0.307 &0.110 &0.404 &0.110 &0.475 &0.072 &0.277 &0.109\\
    \textit{yeah} &0.423 &0.206 &0.351 &0.378 &0.356 &0.147 &0.365 &0.160 &0.408 &0.413 &0.277 &0.132\\
    \textit{uh-huh} &0.490 &0.282 &0.727 &0.391 &0.271 &0.310 &0.680 &0.172 &0.489 &0.460 &0.299 &0.194\\ 
    \textit{well} &0.417 &0.170 &0.414 &0.282 &0.440 &0.311 &0.400 &0.136 &0.332 &0.276 &0.286 &0.123\\ 
    \textit{right} &0.433 &0.219 &0.291 &0.388 &0.315 &0.289 &0.406 &0.149 &0.498 &0.487 &0.288 &0.352\\ 
    \textit{oh} &0.364 &0.176 &0.434 &0.205 &0.388 &0.146 &0.363 &0.131 &0.514 &0.330 &0.271 &0.270\\ 
    \textit{um} &0.374 &0.150 &0.479 &0.069 &0.349 &0.105 &0.413 &0.105 &0.489 &0.166 &0.314 &0.186\\ 
    \textit{okay} &0.425 &0.213 &0.744 &0.397 &0.282 &0.368 &0.438 &0.161 &0.371 &0.411 &0.237 &0.318\\ 
    \textit{no} &0.438 &0.257 &0.746 &0.417 &0.337 &0.309 &0.382 &0.149 &0.334 &0.353 &0.269 &0.219\\ 
    \textit{yes} &0.422 &0.211 &0.753 &0.329 &0.384 &0.229 &0.383 &0.139 &0.360 &0.350 &0.273 &0.188\\ 
    \textit{so} &0.339 &0.166 &0.404 &0.395 &0.244 &0.320 &0.486 &0.149 &0.526 &0.457 &0.246 &0.231\\ 
    \textit{oh yeah} &0.359 &0.575 &0.731 &0.559 &0.509 &0.556 &0.374 &0.167 &0.367 &0.262 &0.298 &0.113\\
    \textit{huh} &0.461 &0.246 &0.761 &0.337 &0.293 &0.170 &0.366 &0.150 &0.501 &0.194 &0.296 &0.217\\
    \textit{mmhmm} &0.505 &0.481 &0.779 &0.549 &0.512 &0.529 &0.442 &0.214 &0.289 &0.268 &0.260 &0.193\\
    \textit{of course} &0.376 &0.149 &0.318 &0.094 &0.348 &0.121 &0.422 &0.124 &0.314 &0.096 &0.281 &0.108\\
\midrule
    \begin{CJK}{UTF8}{min} \bf Average \textuparrow  \end{CJK} &\textbf{0.409} &\textbf{0.245} &\textbf{0.558} &\textbf{0.324} &\textbf{0.356} &\textbf{0.263} &\textbf{0.422} &\textbf{0.148} &\textbf{0.418} &\textbf{0.306} &\textbf{0.278} &\textbf{0.196} \\
\bottomrule
\end{tabularx}
\label{tab:en_sil3}
\end{table*}

\begin{table*}[ht]
\scriptsize
\centering
\caption{%
    Maximum silhouette scores value on the $k$-means clustering of the top 15 selected \textbf{English} backchannels/fillers when using the \textbf{BERT} model in both the original dimensional space and after dimensionality reduction to 100 dimensions using PCA.}
\begin{tabularx}{\textwidth}{lXXXXXXXX}
\toprule
    \textbf{Backchannel/Filler} & \multicolumn{2}{c}{\textbf{no\_ft, no\_ctx}} & \multicolumn{2}{c}{\textbf{ft, no\_ctx (MASK)}} & \multicolumn{2}{c}{\textbf{no\_ft, one\_ctx}} & \multicolumn{2}{c}{\textbf{ft, one\_ctx (MASK)}} \\
\cmidrule(rl){2-3} \cmidrule(rl){4-5} \cmidrule(rl){6-7} \cmidrule(rl){8-9}
     & \textbf{orig.} & \textbf{100} & \textbf{orig.} & \textbf{100} & \textbf{orig.} & \textbf{100} & \textbf{orig.} & \textbf{100} \\
\midrule
    \textit{uh} & 0.097 & 0.122 & 0.083 & 0.103 & 0.172 & 0.189 & 0.054 & 0.074 \\
    \textit{yeah} & 0.085 & 0.118 & 0.216 & 0.245 & 0.098 & 0.119 & 0.079 & 0.104 \\
    \textit{uh-huh} & 0.104 & 0.129 & 0.267 & 0.343 & 0.121 & 0.148 & 0.119 & 0.156 \\
    \textit{well} & 0.072 & 0.095 & 0.211 & 0.252 & 0.050 & 0.064 & 0.186 & 0.233 \\
    \textit{right} & 0.095 & 0.127 & 0.251 & 0.313 & 0.110 & 0.153 & 0.354 & 0.426 \\
    \textit{oh} & 0.079 & 0.097 & 0.225 & 0.256 & 0.090 & 0.113 & 0.075 & 0.093 \\
    \textit{um} & 0.091 & 0.117 & 0.144 & 0.178 & 0.192 & 0.206 & 0.078 & 0.112 \\
    \textit{okay} & 0.088 & 0.114 & 0.219 & 0.216 & 0.074 & 0.097 & 0.175 & 0.218 \\
    \textit{no} & 0.093 & 0.117 & 0.190 & 0.242 & 0.067 & 0.086 & 0.195 & 0.255 \\
    \textit{yes} & 0.096 & 0.122 & 0.206 & 0.227 & 0.071 & 0.089 & 0.122 & 0.146 \\
    \textit{so} & 0.144 & 0.180 & 0.177 & 0.220 & 0.069 & 0.093 & 0.219 & 0.291 \\
    \textit{oh yeah} & 0.115 & 0.142 & 0.162 & 0.178 & 0.119 & 0.139 & 0.114 & 0.137 \\
    \textit{huh} & 0.229 & 0.294 & 0.242 & 0.283 & 0.103 & 0.126 & 0.189 & 0.194 \\
    \textit{mmhmm} & 0.205 & 0.241 & 0.270 & 0.331 & 0.103 & 0.109 & 0.093 & 0.132 \\
    \textit{of course} & 0.069 & 0.096 & 0.088 & 0.111 & 0.050 & 0.073 & 0.088 & 0.103 \\ 
\midrule
    \begin{CJK}{UTF8}{min} \bf Average \textuparrow  \end{CJK} & \textbf{0.111} & \textbf{0.141} & \textbf{0.197} & \textbf{0.233} & \textbf{0.099} & \textbf{0.120} & \textbf{0.143} & \textbf{0.178} \\
\bottomrule
\end{tabularx}
\label{tab:en_sil4}
\end{table*}

\begin{table*}[ht]
\scriptsize
\centering
\caption{Maximum silhouette scores value on the $k$-means clustering of the top 15 selected \textbf{Japanese} backchannels/fillers when using the \textbf{LLaMA-3} model in both the original dimensional space and after dimensionality reduction to 100 dimensions using PCA.}
    
\begin{tabularx}{\textwidth}{lXXXXXXXXXXXX}
\toprule
    \textbf{Backchannel/Filler} & \multicolumn{2}{c}{\textbf{no\_ft, no\_ctx}} & \multicolumn{2}{c}{\textbf{ft, no\_ctx (NTP)}} & \multicolumn{2}{c}{\textbf{no\_ft, one\_ctx}} & \multicolumn{2}{c}{\textbf{ft, one\_ctx (NTP)}}  & \multicolumn{2}{c}{\textbf{no\_ft, full\_ctx}} & \multicolumn{2}{c}{\textbf{ft, full\_ctx (NTP)}}  \\
\cmidrule(lr){2-3} \cmidrule(lr){4-5} \cmidrule(lr){6-7} \cmidrule(rl){8-9} \cmidrule(lr){10-11} \cmidrule(lr){12-13}
     & \textbf{orig.} & \textbf{100} & \textbf{orig.} & \textbf{100} & \textbf{orig.} & \textbf{100} & \textbf{orig.} & \textbf{100} & \textbf{orig.} & \textbf{100} & \textbf{orig.} & \textbf{100} \\
\midrule
    \begin{CJK}{UTF8}{min} うん (un) \end{CJK} &0.265 &0.255 &0.432 &0.442 &0.148 &0.192 &0.246 &0.350 &0.342 &0.348 &0.442 &0.224\\ 
    \begin{CJK}{UTF8}{min} あ (a) \end{CJK} &0.264 &0.213 &0.422 &0.427 &0.147 &0.177 &0.237 &0.341 &0.328 &0.300 &0.149 &0.419\\ 
    \begin{CJK}{UTF8}{min} はい (hai) \end{CJK} &0.261 &0.247 &0.440 &0.464 &0.182 &0.184 &0.305 &0.395 &0.331 &0.235 &0.238 &0.321\\ 
    \begin{CJK}{UTF8}{min} え (e) \end{CJK} &0.290 &0.262 &0.427 &0.430 &0.144 &0.186 &0.251 &0.378 &0.311 &0.233 &0.402 &0.261\\ 
    \begin{CJK}{UTF8}{min} そう (sou) \end{CJK} &0.287 &0.270 &0.429 &0.428 &0.153 &0.184 &0.240 &0.386 &0.324 &0.344 &0.427 &0.291\\ 
    \begin{CJK}{UTF8}{min} ま (ma) \end{CJK} &0.282 &0.242 &0.400 &0.414 &0.114 &0.131 &0.206 &0.316 &0.336 &0.361 &0.139 &0.450\\ 
    \begin{CJK}{UTF8}{min} なんか (nanka) \end{CJK} &0.330 &0.309 &0.496 &0.529 &0.111 &0.163 &0.191 &0.286 &0.345 &0.389 &0.472 &0.532\\ 
    \begin{CJK}{UTF8}{min} あの (ano) \end{CJK} &0.262 &0.252 &0.412 &0.427 &0.147 &0.115 &0.162 &0.281 &0.240 &0.247 &0.123 &0.201\\ 
    \begin{CJK}{UTF8}{min} ん (n) \end{CJK} &0.291 &0.261 &0.429 &0.428 &0.129 &0.162 &0.307 &0.412 &0.142 &0.339 &0.198 &0.269\\ 
    \begin{CJK}{UTF8}{min} そうです (soudesu) \end{CJK} &0.316 &0.283 &0.410 &0.434 &0.133 &0.228 &0.169 &0.280 &0.349 &0.356 &0.464 &0.175\\ 
    \begin{CJK}{UTF8}{min} は (ha) \end{CJK} &0.228 &0.210 &0.352 &0.355 &0.116 &0.164 &0.242 &0.343 &0.265 &0.316 &0.359 &0.315\\ 
    \begin{CJK}{UTF8}{min} ね (ne) \end{CJK} &0.249 &0.246 &0.369 &0.380 &0.115 &0.174 &0.332 &0.436 &0.141 &0.322 &0.311 &0.400\\ 
    \begin{CJK}{UTF8}{min} いや (iya) \end{CJK} &0.283 &0.205 &0.418 &0.474 &0.166 &0.181 &0.246 &0.328 &0.319 &0.358 &0.423 &0.465\\ 
    \begin{CJK}{UTF8}{min} へー (he-) \end{CJK} &0.336 &0.253 &0.533 &0.591 &0.185 &0.221 &0.154 &0.232 &0.405 &0.436 &0.560 &0.605\\ 
    \begin{CJK}{UTF8}{min} そうか (souka) \end{CJK} &0.319 &0.252 &0.442 &0.455 &0.158 &0.192 &0.271 &0.373 &0.328 &0.369 &0.203 &0.225\\ 
\midrule
    \begin{CJK}{UTF8}{min} \bf Average \textuparrow \end{CJK} &\textbf{0.284} &\textbf{0.256} &\textbf{0.427} &\textbf{0.445} &\textbf{0.143} &\textbf{0.177} &\textbf{0.237} &\textbf{0.343} &\textbf{0.300} &\textbf{0.330} &\textbf{0.327} &\textbf{0.344} \\ 
\bottomrule
\end{tabularx}
\label{tab:jap_sil1}
\end{table*}

\begin{table*}[ht]
\scriptsize
\centering
\caption{Maximum silhouette scores value on the $k$-means clustering of the top 15 selected \textbf{Japanese} backchannels/fillers when using the \textbf{Qwen-3} model in both the original dimensional space and after dimensionality reduction to 100 dimensions using PCA.}
    
\begin{tabularx}{\textwidth}{lXXXXXXXXXXXX}
\toprule
    \textbf{Backchannel/Filler} & \multicolumn{2}{c}{\textbf{no\_ft, no\_ctx}} & \multicolumn{2}{c}{\textbf{ft, no\_ctx (NTP)}} & \multicolumn{2}{c}{\textbf{no\_ft, one\_ctx}} & \multicolumn{2}{c}{\textbf{ft, one\_ctx (NTP)}} & \multicolumn{2}{c}{\textbf{no\_ft, full\_ctx}} & \multicolumn{2}{c}{\textbf{ft, full\_ctx (NTP)}} \\
\cmidrule(lr){2-3} \cmidrule(lr){4-5} \cmidrule(lr){6-7} \cmidrule(lr){8-9} \cmidrule(lr){10-11} \cmidrule(lr){12-13}
     & \textbf{orig.} & \textbf{100} & \textbf{orig.} & \textbf{100} & \textbf{orig.} & \textbf{100} & \textbf{orig.} & \textbf{100} & \textbf{orig.} & \textbf{100} & \textbf{orig.} & \textbf{100} \\
\midrule
    \begin{CJK}{UTF8}{min} うん (un)\end{CJK} &0.292 &0.183 &0.666 &0.564 &0.363 &0.186 &0.275 &0.235 &0.189 &0.130 &0.208 &0.154\\ 
    \begin{CJK}{UTF8}{min} あ (a)\end{CJK} &0.272 &0.172 &0.157 &0.535 &0.344 &0.151 &0.225 &0.208 &0.291 &0.188 &0.165 &0.168\\ 
    \begin{CJK}{UTF8}{min} はい (hai)\end{CJK} &0.322 &0.242 &0.359 &0.496 &0.356 &0.222 &0.302 &0.240 &0.632 &0.242 &0.168 &0.200\\ 
    \begin{CJK}{UTF8}{min} え (e)\end{CJK} &0.291 &0.115 &0.740 &0.578 &0.371 &0.134 &0.198 &0.211 &0.213 &0.126 &0.179 &0.154\\ 
    \begin{CJK}{UTF8}{min} そう (sou)\end{CJK} &0.295 &0.191 &0.706 &0.557 &0.353 &0.149 &0.266 &0.289 &0.299 &0.187 &0.175 &0.169\\ 
    \begin{CJK}{UTF8}{min} ま (ma)\end{CJK} &0.295 &0.168 &0.318 &0.290 &0.328 &0.122 &0.335 &0.330 &0.152 &0.149 &0.130 &0.124\\
    \begin{CJK}{UTF8}{min} なんか (nanka)\end{CJK} &0.243 &0.164 &0.792 &0.153 &0.304 &0.115 &0.155 &0.171 &0.182 &0.172 &0.156 &0.171\\ 
    \begin{CJK}{UTF8}{min} あの (ano)\end{CJK} &0.292 &0.178 &0.770 &0.223 &0.295 &0.131 &0.184 &0.209 &0.201 &0.137 &0.140 &0.161\\ 
    \begin{CJK}{UTF8}{min} ん (n)\end{CJK} &0.286 &0.151 &0.739 &0.167 &0.360 &0.155 &0.286 &0.241 &0.252 &0.174 &0.190 &0.144\\ 
    \begin{CJK}{UTF8}{min} そうです (soudesu)\end{CJK} &0.306 &0.122 &0.702 &0.558 &0.343 &0.133 &0.203 &0.267 &0.256 &0.165 &0.150 &0.176\\ 
    \begin{CJK}{UTF8}{min} は (ha)\end{CJK} &0.337 &0.227 &0.335 &0.299 &0.365 &0.219 &0.340 &0.313 &0.320 &0.183 &0.204 &0.224\\ 
    \begin{CJK}{UTF8}{min} ね (ne)\end{CJK} &0.298 &0.140 &0.346 &0.405 &0.357 &0.122 &0.450 &0.493 &0.205 &0.114 &0.370 &0.336\\ 
    \begin{CJK}{UTF8}{min} いや (iya)\end{CJK} &0.294 &0.144 &0.742 &0.609 &0.339 &0.419 &0.215 &0.219 &0.198 &0.109 &0.152 &0.162\\ 
    \begin{CJK}{UTF8}{min} へー (he-)\end{CJK} &0.321 &0.151 &0.843 &0.723 &0.406 &0.183 &0.256 &0.223 &0.170 &0.196 &0.182 &0.177\\ 
    \begin{CJK}{UTF8}{min} そうか (souka)\end{CJK} &0.304 &0.128 &0.732 &0.587 &0.298 &0.106 &0.224 &0.250 &0.178 &0.261 &0.176 &0.154\\
\midrule
    \begin{CJK}{UTF8}{min} \bf Average \textuparrow \end{CJK} &\textbf{0.296} &\textbf{0.165} &\textbf{0.596} &\textbf{0.450} &\textbf{0.345} &\textbf{0.152} &\textbf{0.261} &\textbf{0.260} &\textbf{0.249} &\textbf{0.169} &\textbf{0.183} &\textbf{0.178} \\ 
\bottomrule
\end{tabularx}
\label{tab:jap_sil2}
\end{table*}

\begin{table*}[ht]
\scriptsize
\centering
\caption{Maximum silhouette scores value on the $k$-means clustering of the top 15 selected \textbf{Japanese} backchannels/fillers when using the \textbf{GPT-2} model in both the original dimensional space and after dimensionality reduction to 100 dimensions using PCA.}
    
\begin{tabularx}{\textwidth}{lXXXXXXXXXXXX}
\toprule
    \textbf{Backchannel/Filler} & \multicolumn{2}{c}{\textbf{no\_ft, no\_ctx}} & \multicolumn{2}{c}{\textbf{ft, no\_ctx (NTP)}} & \multicolumn{2}{c}{\textbf{ft, no\_ctx (TTP)}}   & \multicolumn{2}{c}{\textbf{no\_ft, one\_ctx}} & \multicolumn{2}{c}{\textbf{}\textbf{ft, one\_ctx (NTP)}} & \multicolumn{2}{c}{\textbf{ft, one\_ctx (TTP)}} \\
\cmidrule(lr){2-3} \cmidrule(lr){4-5} \cmidrule(lr){6-7} \cmidrule(lr){8-9} \cmidrule(lr){10-11} \cmidrule(lr){12-13}
     & \textbf{orig.} & \textbf{100} & \textbf{orig.} & \textbf{100} & \textbf{orig.} & \textbf{100} & \textbf{orig.} & \textbf{100} & \textbf{orig.} & \textbf{100} & \textbf{orig.} & \textbf{100} \\
\midrule
    \begin{CJK}{UTF8}{min} うん (un)\end{CJK} &0.122 &0.153 &0.241 &0.261 &0.302 &0.311 &0.099 &0.103 &0.235 &0.262 &0.316 &0.321\\ 
    \begin{CJK}{UTF8}{min} あ (a)\end{CJK} &0.114 &0.119 &0.193 &0.229 &0.256 &0.249 &0.083 &0.081 &0.238 &0.241 &0.292 &0.283\\ 
    \begin{CJK}{UTF8}{min} はい (hai)\end{CJK} &0.186 &0.201 &0.338 &0.390 &0.240 &0.258 &0.107 &0.120 &0.362 &0.388 &0.187 &0.211\\ 
    \begin{CJK}{UTF8}{min} え (e)\end{CJK} &0.156 &0.150 &0.265 &0.269 &0.293 &0.295 &0.097 &0.087 &0.258 &0.293 &0.319 &0.320\\ 
    \begin{CJK}{UTF8}{min} そう (sou)\end{CJK} &0.146 &0.140 &0.264 &0.305 &0.327 &0.335 &0.122 &0.104 &0.274 &0.312 &0.337 &0.342\\ 
    \begin{CJK}{UTF8}{min} ま (ma)\end{CJK} &0.118 &0.146 &0.204 &0.252 &0.267 &0.278 &0.095 &0.093 &0.194 &0.249 &0.294 &0.303\\
    \begin{CJK}{UTF8}{min} なんか (nanka)\end{CJK} &0.118 &0.155 &0.103 &0.134 &0.185 &0.195 &0.102 &0.095 &0.119 &0.148 &0.150 &0.174\\ 
    \begin{CJK}{UTF8}{min} あの (ano)\end{CJK} &0.116 &0.160 &0.299 &0.325 &0.268 &0.279 &0.092 &0.093 &0.289 &0.340 &0.290 &0.304\\ 
    \begin{CJK}{UTF8}{min} ん (n)\end{CJK} &0.107 &0.133 &0.237 &0.278 &0.275 &0.286 &0.082 &0.079 &0.263 &0.296 &0.299 &0.296\\ 
    \begin{CJK}{UTF8}{min} そうです (soudesu)\end{CJK} &0.136 &0.165 &0.345 &0.369 &0.363 &0.309 &0.119 &0.116 &0.159 &0.178 &0.103 &0.136\\ 
    \begin{CJK}{UTF8}{min} は (ha)\end{CJK} &0.135 &0.160 &0.274 &0.313 &0.199 &0.227 &0.121 &0.128 &0.292 &0.330 &0.258 &0.214\\ 
    \begin{CJK}{UTF8}{min} ね (ne)\end{CJK} &0.091 &0.128 &0.284 &0.347 &0.161 &0.208 &0.074 &0.091 &0.288 &0.340 &0.249 &0.250\\ 
    \begin{CJK}{UTF8}{min} いや (iya)\end{CJK} &0.154 &0.163 &0.178 &0.310 &0.226 &0.348 &0.123 &0.100 &0.172 &0.197 &0.164 &0.182\\ 
    \begin{CJK}{UTF8}{min} へー (he-)\end{CJK} &0.187 &0.210 &0.341 &0.375 &0.284 &0.365 &0.104 &0.089 &0.258 &0.284 &0.155 &0.183\\ 
    \begin{CJK}{UTF8}{min} そうか (souka)\end{CJK} &0.158 &0.152 &0.254 &0.278 &0.249 &0.302 &0.134 &0.110 &0.204 &0.246 &0.208 &0.243\\
\midrule
    \begin{CJK}{UTF8}{min} \bf Average \textuparrow \end{CJK} &\textbf{0.136} &\textbf{0.156} &\textbf{0.255} &\textbf{0.296} &\textbf{0.260} &\textbf{0.283} &\textbf{0.104} &\textbf{0.099} &\textbf{0.240} &\textbf{0.273} &\textbf{0.241} &\textbf{0.251} \\
\bottomrule
\end{tabularx}
\label{tab:jap_sil3}
\end{table*}

\begin{table*}[ht]
\scriptsize
\centering
\caption{Maximum silhouette scores value on the $k$-means clustering of the top 15 selected \textbf{Japanese} backchannels/fillers when using the \textbf{BERT} model in both the original dimensional space and after dimensionality reduction to 100 dimensions using PCA.}
    
\begin{tabularx}{\textwidth}{lXXXXXXXX}
\toprule
    \textbf{Backchannel/Filler} & \multicolumn{2}{c}{\textbf{no\_ft, no\_ctx}} & \multicolumn{2}{c}{\textbf{ft, no\_ctx (MASK)}} & \multicolumn{2}{c}{\textbf{no\_ft, one\_ctx}} & \multicolumn{2}{c}{\textbf{ft, one\_ctx (MASK)}} \\
\cmidrule(lr){2-3} \cmidrule(lr){4-5} \cmidrule(lr){6-7} \cmidrule(lr){8-9}
    & \textbf{orig.} & \textbf{100} & \textbf{orig.} & \textbf{100} & \textbf{orig.} & \textbf{100} & \textbf{orig.} & \textbf{100} \\
\midrule
    \begin{CJK}{UTF8}{min} うん (un)\end{CJK} & 0.232 & 0.265 & 0.385 & 0.459 & 0.211 & 0.242 & 0.463 & 0.487 \\ 
    \begin{CJK}{UTF8}{min} あ (a)\end{CJK} & 0.147 & 0.166 & 0.399 & 0.393 & 0.141 & 0.171 & 0.447 & 0.428 \\ 
    \begin{CJK}{UTF8}{min} はい (hai)\end{CJK} & 0.288 & 0.330 & 0.423 & 0.195 & 0.209 & 0.240 & 0.440 & 0.213 \\ 
    \begin{CJK}{UTF8}{min} え (e)\end{CJK} & 0.164 & 0.191 & 0.293 & 0.364 & 0.150 & 0.170 & 0.406 & 0.425 \\ 
    \begin{CJK}{UTF8}{min} そう (sou)\end{CJK} & 0.217 & 0.258 & 0.379 & 0.419 & 0.205 & 0.226 & 0.403 & 0.495 \\ 
    \begin{CJK}{UTF8}{min} ま (ma)\end{CJK} & 0.148 & 0.177 & 0.400 & 0.430 & 0.136 & 0.160 & 0.414 & 0.491 \\
    \begin{CJK}{UTF8}{min} なんか (nanka)\end{CJK} & 0.105 & 0.134 & 0.329 & 0.372 & 0.115 & 0.143 & 0.406 & 0.455 \\ 
    \begin{CJK}{UTF8}{min} あの (ano)\end{CJK} & 0.117 & 0.157 & 0.342 & 0.386 & 0.114 & 0.145 & 0.372 & 0.425 \\ 
    \begin{CJK}{UTF8}{min} ん (n)\end{CJK} & 0.155 & 0.192 & 0.429 & 0.488 & 0.161 & 0.190 & 0.523 & 0.569 \\ 
    \begin{CJK}{UTF8}{min} そうです (soudesu)\end{CJK} & 0.136 & 0.170 & 0.227 & 0.303 & 0.119 & 0.140 & 0.351 & 0.393 \\ 
    \begin{CJK}{UTF8}{min} は (ha)\end{CJK} & 0.208 & 0.250 & 0.403 & 0.437 & 0.204 & 0.247 & 0.431 & 0.484 \\ 
    \begin{CJK}{UTF8}{min} ね (ne)\end{CJK} & 0.206 & 0.264 & 0.324 & 0.371 & 0.169 & 0.210 & 0.377 & 0.424 \\ 
    \begin{CJK}{UTF8}{min} いや (iya)\end{CJK} & 0.162 & 0.185 & 0.408 & 0.461 & 0.165 & 0.185 & 0.462 & 0.511 \\ 
    \begin{CJK}{UTF8}{min} へー (he-)\end{CJK} & 0.160 & 0.176 & 0.249 & 0.287 & 0.152 & 0.180 & 0.081 & 0.104 \\ 
    \begin{CJK}{UTF8}{min} そうか (souka)\end{CJK} & 0.203 & 0.237 & 0.346 & 0.375 & 0.258 & 0.296 & 0.410 & 0.436 \\
\midrule
    \begin{CJK}{UTF8}{min} \bf Average \textuparrow \end{CJK} & \textbf{0.177} & \textbf{0.210} & \textbf{0.356} & \textbf{0.383} & \textbf{0.167} & \textbf{0.196} & \textbf{0.399} & \textbf{0.423} \\ 
\bottomrule
\end{tabularx}
\label{tab:jap_sil4}
\end{table*}

\FloatBarrier 

\begin{table*}[ht]
\scriptsize
\caption{%
    $k$-values of optimal results on the $k$-means clustering of the top 15 selected \textbf{English} backchannels/fillers when using the \textbf{LLaMA-3} model in both the original dimensional space and after dimensionality reduction to 100 dimensions using PCA.}
\begin{tabularx}{\textwidth}{lXXXXXXXXXXXX}
\toprule
    \textbf{Backchannel/Filler} & \multicolumn{2}{c}{\textbf{no\_ft, no\_ctx}} & \multicolumn{2}{c}{\textbf{ft, no\_ctx (NTP)}}  & \multicolumn{2}{c}{\textbf{no\_ft, one\_ctx}} & \multicolumn{2}{c}{\textbf{ft, one\_ctx (NTP)}}  & \multicolumn{2}{c}{\textbf{no\_ft, full\_ctx}} & \multicolumn{2}{c}{\textbf{ft, full\_ctx (NTP)}}  \\
    \cmidrule(lr){2-3} \cmidrule(lr){4-5} \cmidrule(lr){6-7} \cmidrule(lr){8-9} \cmidrule(lr){10-11} \cmidrule(lr){12-13}
     & \textbf{orig.} & \textbf{100} & \textbf{orig.} & \textbf{100} & \textbf{orig.} & \textbf{100} & \textbf{orig.} & \textbf{100} & \textbf{orig.} & \textbf{100} & \textbf{orig.} & \textbf{100} \\
\midrule
    \textit{uh} &2 &2 &2 &3 &2 &2 &2 &2 &2 &2 &2 &2\\
    \textit{yeah} &2 &2 &3 &3 &2 &2 &2 &2 &3 &2 &2 &2\\
    \textit{uh-huh} &2 &2 &3 &3 &2 &2 &2 &2 &2 &2 &2 &2\\
    \textit{well} &2 &2 &2 &2 &2 &2 &3 &2 &2 &2 &2 &2\\
    \textit{right} &2 &2 &2 &2 &2 &2 &2 &2 &3 &2 &2 &2\\
    \textit{oh} &2 &2 &2 &3 &2 &2 &2 &2 &2 &2 &3 &3\\
    \textit{um} &2 &2 &3 &3 &2 &2 &2 &2 &3 &2 &2 &2\\
    \textit{okay} &2 &2 &2 &3 &2 &2 &2 &2 &3 &2 &2 &2\\
    \textit{no} &2 &2 &2 &3 &2 &2 &2 &2 &3 &3 &2 &2\\
    \textit{yes} &2 &2 &2 &3 &2 &2 &2 &2 &2 &2 &2 &2\\
    \textit{so} &2 &2 &2 &3 &2 &2 &2 &2 &4 &4 &2 &2\\
    \textit{oh yeah} &3 &9 &3 &11 &2 &2 &4 &4 &2 &2 &3 &3\\
    \textit{huh} &2 &2 &3 &3 &2 &2 &2 &2 &2 &2 &2 &2\\
    \textit{mmhmm} &2 &15 &3 &9 &2 &2 &3 &3 &2 &2 &3 &3\\
    \textit{of course} &2 &2 &2 &3 &2 &2 &3 &2 &2 &2 &2 &2\\
\midrule
\textbf{Average} &\textbf{2.07} &\textbf{3.33} &\textbf{2.40} &\textbf{3.80} &\textbf{2.00} &\textbf{2.00} &\textbf{2.33} &\textbf{2.20} &\textbf{2.47} &\textbf{2.20} &\textbf{2.07} &\textbf{2.20} \\ 
\bottomrule
\end{tabularx}
\label{tab:en_k1}
\end{table*}

\begin{table*}[ht]
\scriptsize
\centering
\caption{$k$-values of optimal results on the $k$-means clustering of the top 15 selected \textbf{English} backchannels/fillers when using the \textbf{Qwen-3} model in both the original dimensional space and after dimensionality reduction to 100 dimensions using PCA.}

\begin{tabularx}{\textwidth}{lXXXXXXXXXXXX}
\toprule
    \textbf{Backchannel/Filler} & \multicolumn{2}{c}{\textbf{no\_ft, no\_ctx}} & \multicolumn{2}{c}{\textbf{ft, no\_ctx (NTP)}} & \multicolumn{2}{c}{\textbf{no\_ft, one\_ctx}} & \multicolumn{2}{c}{\textbf{ft, one\_ctx (NTP)}} & \multicolumn{2}{c}{\textbf{no\_ft, full\_ctx}} & \multicolumn{2}{c}{\textbf{ft, full\_ctx (NTP)}} \\
\cmidrule(lr){2-3} \cmidrule(lr){4-5} \cmidrule(lr){6-7} \cmidrule(lr){8-9} \cmidrule(lr){10-11} \cmidrule(lr){12-13}
     & \textbf{orig.} & \textbf{100} & \textbf{orig.} & \textbf{100} & \textbf{orig.} & \textbf{100} & \textbf{orig.} & \textbf{100} & \textbf{orig.} & \textbf{100} & \textbf{orig.} & \textbf{100} \\
\midrule
    \textit{uh} & 2 & 2 & 2 & 2 & 2 & 2 & 2 & 2 & 2 & 2 & 2 & 2 \\ 
    \textit{yeah} & 2 & 2 & 4 & 3 & 2 & 2 & 2 & 3 & 2 & 3 & 2 & 3 \\ 
    \textit{uh-huh} & 2 & 2 & 2 & 3 & 2 & 2 & 2 & 2 & 2 & 2 & 2 & 2 \\ 
    \textit{well} & 2 & 2 & 2 & 3 & 2 & 2 & 2 & 3 & 2 & 2 & 2 & 3 \\ 
    \textit{right} & 2 & 5 & 2 & 3 & 2 & 2 & 2 & 2 & 2 & 2 & 2 & 2 \\ 
    \textit{oh} & 2 & 4 & 2 & 4 & 2 & 2 & 3 & 3 & 2 & 2 & 2 & 3 \\ 
    \textit{um} & 2 & 2 & 2 & 2 & 2 & 2 & 2 & 2 & 2 & 2 & 2 & 2 \\ 
    \textit{okay} & 2 & 2 & 3 & 3 & 2 & 2 & 2 & 2 & 2 & 3 & 2 & 2 \\ 
    \textit{no} & 2 & 14 & 5 & 3 & 2 & 2 & 2 & 2 & 2 & 3 & 2 & 2 \\ 
    \textit{yes} & 2 & 10 & 2 & 3 & 2 & 3 & 3 & 2 & 2 & 2 & 2 & 3 \\ 
    \textit{so} & 2 & 2 & 2 & 3 & 2 & 2 & 3 & 3 & 2 & 2 & 2 & 2 \\ 
    \textit{oh} yeah & 14 & 15 & 2 & 2 & 2 & 2 & 5 & 3 & 2 & 2 & 4 & 3 \\ 
    \textit{huh} & 2 & 2 & 2 & 2 & 2 & 2 & 2 & 2 & 2 & 2 & 2 & 2 \\ 
    \textit{mmhmm} & 2 & 12 & 2 & 2 & 2 & 2 & 4 & 3 & 2 & 2 & 2 & 2 \\ 
    \textit{of course} & 3 & 2 & 2 & 3 & 2 & 2 & 2 & 2 & 2 & 2 & 2 & 2 \\ 
\midrule
    \textbf{Average} & \textbf{2.87} & \textbf{5.20} & \textbf{2.40} & \textbf{2.73} & \textbf{2.00} & \textbf{2.07} & \textbf{2.53} & \textbf{2.40} & \textbf{2.00} & \textbf{2.20} & \textbf{2.13} & \textbf{2.33} \\ 
    \bottomrule
    \end{tabularx}
    \label{tab:eng_k}
\end{table*}

\begin{table*}[ht]
\scriptsize
\caption{$k$-values of optimal results on the $k$-means clustering of the top 15 selected \textbf{English} backchannels/fillers when using the \textbf{GPT-2} model in both the original dimensional space and after dimensionality reduction to 100 dimensions using PCA.}
    
\begin{tabularx}{\textwidth}{lXXXXXXXXXXXX}
\toprule
    \textbf{Backchannel/Filler} & \multicolumn{2}{c}{\textbf{no\_ft, no\_ctx}} & \multicolumn{2}{c}{\textbf{ft, no\_ctx (NTP)}} & \multicolumn{2}{c}{\textbf{ft, no\_ctx (TTP)}}   & \multicolumn{2}{c}{\textbf{no\_ft, one\_ctx}} & \multicolumn{2}{c}{\textbf{ft, one\_ctx (NTP)}} & \multicolumn{2}{c}{\textbf{ft, one\_ctx (TTP)}} \\
\cmidrule(lr){2-3} \cmidrule(lr){4-5} \cmidrule(lr){6-7} \cmidrule(lr){8-9} \cmidrule(lr){10-11} \cmidrule(lr){12-13}
     & \textbf{orig.} & \textbf{100} & \textbf{orig.} & \textbf{100} & \textbf{orig.} & \textbf{100} & \textbf{orig.} & \textbf{100} & \textbf{orig.} & \textbf{100} & \textbf{orig.} & \textbf{100} \\
\midrule
    \textit{uh} &2 &2 &2 &2 &2 &2 &2 &3 &2 &2 &2 &2\\
    \textit{yeah} &2 &2 &2 &2 &2 &2 &2 &2 &2 &2 &2 &2\\
    \textit{uh-huh} &2 &4 &2 &3 &2 &2 &2 &2 &2 &2 &2 &2\\ 
    \textit{well} &2 &7 &2 &2 &2 &2 &2 &2 &3 &3 &2 &4\\ 
    \textit{right} &2 &15 &2 &3 &2 &2 &2 &2 &2 &2 &2 &2\\ 
    \textit{oh} &2 &2 &2 &2 &2 &4 &2 &2 &2 &3 &2 &2\\ 
    \textit{um} &2 &2 &2 &2 &2 &2 &2 &2 &2 &2 &2 &2\\ 
    \textit{okay} &2 &15 &2 &3 &2 &2 &2 &2 &2 &2 &3 &2\\ 
    \textit{no} &2 &8 &2 &3 &2 &2 &2 &2 &2 &2 &2 &2\\ 
    \textit{yes} &3 &13 &2 &2 &2 &5 &2 &2 &2 &2 &2 &2\\ 
    \textit{so} &2 &7 &2 &3 &2 &5 &2 &2 &2 &2 &2 &4\\ 
    \textit{oh yeah} &3 &15 &2 &4 &2 &3 &2 &2 &2 &2 &2 &2\\
    \textit{huh} &2 &2 &2 &2 &2 &4 &3 &2 &2 &2 &2 &2\\
    \textit{mmhmm} &2 &15 &2 &2 &4 &7 &2 &2 &2 &2 &2 &2\\
    \textit{of course} &2 &2 &2 &5 &2 &2 &2 &2 &2 &2 &2 &2\\ 
\midrule
    \textbf{Average} &\textbf{2.13} &\textbf{7.40} &\textbf{2.00} &\textbf{2.87} &\textbf{2.13} &\textbf{3.07} &\textbf{2.07} &\textbf{2.07} &\textbf{2.13} &\textbf{2.13} &\textbf{2.07} &\textbf{2.27} \\ 
\bottomrule
\end{tabularx}
\label{tab:en_k2}
\end{table*}

\begin{table*}[ht]
\scriptsize
\centering
\caption{$k$-values of optimal results on the $k$-means clustering of the top 15 selected \textbf{English} backchannels/fillers when using the \textbf{BERT} model in both the original dimensional space and after dimensionality reduction to 100 dimensions using PCA.}
    
\begin{tabularx}{\textwidth}{lXXXXXXXX}
\toprule
    \textbf{Backchannel/Filler} & \multicolumn{2}{c}{\textbf{no\_ft, no\_ctx}} & \multicolumn{2}{c}{\textbf{ft, no\_ctx (MASK)}} & \multicolumn{2}{c}{\textbf{no\_ft, one\_ctx}} & \multicolumn{2}{c}{\textbf{ft, one\_ctx (MASK)}} \\
\cmidrule(rl){2-3} \cmidrule(rl){4-5} \cmidrule(rl){6-7} \cmidrule(rl){8-9}
     & \textbf{orig.} & \textbf{100} & \textbf{orig.} & \textbf{100} & \textbf{orig.} & \textbf{100} & \textbf{orig.} & \textbf{100} \\
\midrule
    \textit{uh} & 2 & 2 & 2 & 2 & 2 & 2 & 2 & 2 \\
    \textit{yeah} & 2 & 3 & 2 & 2 & 2 & 2 & 3 & 3 \\
    \textit{uh-huh} & 3 & 3 & 3 & 2 & 2 & 2 & 2 & 2 \\
    \textit{well} & 2 & 2 & 2 & 3 & 2 & 2 & 2 & 2 \\
    \textit{right} & 2 & 2 & 2 & 2 & 2 & 2 & 2 & 2 \\
    \textit{oh} & 4 & 3 & 2 & 2 & 2 & 2 & 3 & 5 \\
    \textit{um} & 2 & 2 & 2 & 2 & 2 & 2 & 2 & 2 \\
    \textit{okay} & 9 & 9 & 2 & 4 & 2 & 2 & 2 & 2 \\
    \textit{no} & 2 & 2 & 2 & 2 & 2 & 2 & 2 & 2 \\
    \textit{yes} & 5 & 2 & 2 & 2 & 2 & 2 & 2 & 2 \\
    \textit{so} & 2 & 2 & 3 & 3 & 2 & 3 & 3 & 2 \\
    \textit{oh yeah} & 4 & 3 & 3 & 3 & 2 & 2 & 2 & 2 \\
    \textit{huh} & 2 & 2 & 2 & 2 & 2 & 2 & 2 & 2 \\
    \textit{mmhmm} & 2 & 3 & 2 & 2 & 3 & 4 & 2 & 2 \\
    \textit{of course} & 2 & 2 & 2 & 2 & 2 & 2 & 2 & 2 \\ 
\midrule
\textbf{Average} & \textbf{3.00} & \textbf{2.80} & \textbf{2.20} & \textbf{2.33} & \textbf{2.06} & \textbf{2.20} & \textbf{2.20} & \textbf{2.27} \\ \bottomrule
    \end{tabularx}
    \label{tab:en_k3}
\end{table*}

\begin{table*}[ht]
\scriptsize
\centering
\caption{$k$-values of optimal results on the $k$-means clustering of the top 15 selected \textbf{Japanese} backchannels/fillers when using the \textbf{LLaMA-3} model in both the original dimensional space and after dimensionality reduction to 100 dimensions using PCA.}
    
\begin{tabularx}{\textwidth}{lXXXXXXXXXXXX}
\toprule
    \textbf{Backchannel/Filler} & \multicolumn{2}{c}{\textbf{no\_ft, no\_ctx}} & \multicolumn{2}{c}{\textbf{ft, no\_ctx (NTP)}} & \multicolumn{2}{c}{\textbf{no\_ft, one\_ctx}} & \multicolumn{2}{c}{\textbf{ft, one\_ctx (NTP)}}  & \multicolumn{2}{c}{\textbf{no\_ft, full\_ctx}} & \multicolumn{2}{c}{\textbf{ft, full\_ctx (NTP)}}  \\
\cmidrule(lr){2-3} \cmidrule(lr){4-5} \cmidrule(lr){6-7} \cmidrule(rl){8-9} \cmidrule(lr){10-11} \cmidrule(lr){12-13}
     & \textbf{orig.} & \textbf{100} & \textbf{orig.} & \textbf{100} & \textbf{orig.} & \textbf{100} & \textbf{orig.} & \textbf{100} & \textbf{orig.} & \textbf{100} & \textbf{orig.} & \textbf{100} \\
\midrule
    \begin{CJK}{UTF8}{min} うん (un) \end{CJK} &3 &3 &2 &3 &2 &2 &2 &4 &2 &2 &2 &3\\ 
    \begin{CJK}{UTF8}{min} あ (a) \end{CJK} &3 &3 &2 &3 &2 &2 &4 &3 &2 &2 &2 &2\\ 
    \begin{CJK}{UTF8}{min} はい (hai) \end{CJK} &3 &5 &2 &3 &2 &2 &2 &2 &2 &2 &3 &2\\ 
    \begin{CJK}{UTF8}{min} え (e) \end{CJK} &3 &3 &2 &3 &2 &2 &2 &4 &2 &2 &2 &6\\ 
    \begin{CJK}{UTF8}{min} そう (sou) \end{CJK} &3 &3 &2 &3 &2 &3 &3 &5 &2 &2 &2 &5\\ 
    \begin{CJK}{UTF8}{min} ま (ma) \end{CJK} &3 &3 &2 &3 &2 &2 &2 &3 &2 &2 &3 &2\\ 
    \begin{CJK}{UTF8}{min} なんか (nanka) \end{CJK} &2 &3 &3 &3 &2 &2 &2 &2 &2 &2 &2 &2\\ 
    \begin{CJK}{UTF8}{min} あの (ano) \end{CJK} &2 &3 &2 &2 &2 &3 &2 &2 &2 &2 &3 &3\\ 
    \begin{CJK}{UTF8}{min} ん (n) \end{CJK} &3 &3 &2 &2 &2 &2 &2 &2 &2 &2 &3 &3\\ 
    \begin{CJK}{UTF8}{min} そうです (soudesu) \end{CJK} &3 &3 &4 &4 &2 &2 &3 &2 &2 &2 &2 &4\\ 
    \begin{CJK}{UTF8}{min} は (ha) \end{CJK} &3 &4 &2 &3 &2 &2 &3 &3 &2 &2 &2 &3\\ 
    \begin{CJK}{UTF8}{min} ね (ne) \end{CJK} &3 &3 &2 &2 &2 &2 &2 &2 &3 &3 &3 &3\\ 
    \begin{CJK}{UTF8}{min} いや (iya) \end{CJK} &2 &3 &2 &3 &2 &3 &2 &2 &2 &2 &2 &2\\ 
    \begin{CJK}{UTF8}{min} へー (he-) \end{CJK} &2 &3 &3 &3 &2 &2 &2 &2 &2 &2 &2 &2\\ 
    \begin{CJK}{UTF8}{min} そうか (souka) \end{CJK} &3 &3 &3 &3 &2 &2 &4 &4 &2 &2 &5 &4\\
\midrule
    \begin{CJK}{UTF8}{min} \bf Average \end{CJK} &\textbf{2.73} &\textbf{3.20} &\textbf{2.33} &\textbf{2.87} &\textbf{2.00} &\textbf{2.13} &\textbf{2.53} &\textbf{2.80} &\textbf{2.13} &\textbf{2.00} &\textbf{2.53} &\textbf{3.07} \\ 
\bottomrule
\end{tabularx}
\label{tab:jap_k1}
\end{table*}

\begin{table*}[ht]
\scriptsize
\centering
\caption{%
    $k$-values of optimal results on the $k$-means clustering of the top 15 selected \textbf{Japanese} backchannels/fillers when using the \textbf{Qwen-3} model in both the original dimensional space and after dimensionality reduction to 100 dimensions using PCA.}
\begin{tabularx}{\textwidth}{lXXXXXXXXXXXX}
\toprule
    \textbf{Backchannel/Filler} & \multicolumn{2}{c}{\textbf{no\_ft, no\_ctx}} & \multicolumn{2}{c}{\textbf{ft, no\_ctx (NTP)}} & \multicolumn{2}{c}{\textbf{no\_ft, one\_ctx}} & \multicolumn{2}{c}{\textbf{ft, one\_ctx (NTP)}} & \multicolumn{2}{c}{\textbf{no\_ft, full\_ctx}} & \multicolumn{2}{c}{\textbf{ft, full\_ctx (NTP)}} \\
\cmidrule(lr){2-3} \cmidrule(lr){4-5} \cmidrule(lr){6-7} \cmidrule(lr){8-9} \cmidrule(lr){10-11} \cmidrule(lr){12-13}
     & \textbf{orig.} & \textbf{100} & \textbf{orig.} & \textbf{100} & \textbf{orig.} & \textbf{100} & \textbf{orig.} & \textbf{100} & \textbf{orig.} & \textbf{100} & \textbf{orig.} & \textbf{100} \\
\midrule
    \begin{CJK}{UTF8}{min} うん (un)\end{CJK} &2 &3 &2 &2 &2 &2 &2 &3 &2 &4 &2 &2\\ 
    \begin{CJK}{UTF8}{min} あ (a)\end{CJK} &2 &2 &2 &2 &2 &2 &2 &3 &2 &2 &2 &2\\ 
    \begin{CJK}{UTF8}{min} はい (hai)\end{CJK} &2 &2 &4 &2 &2 &2 &2 &3 &2 &2 &2 &2\\ 
    \begin{CJK}{UTF8}{min} え (e)\end{CJK} &2 &5 &2 &2 &2 &2 &2 &3 &3 &2 &2 &2\\ 
    \begin{CJK}{UTF8}{min} そう (sou)\end{CJK} &2 &2 &2 &2 &2 &2 &2 &2 &2 &2 &2 &2\\ 
    \begin{CJK}{UTF8}{min} ま (ma)\end{CJK} &2 &2 &2 &2 &2 &3 &2 &2 &2 &2 &2 &4\\
    \begin{CJK}{UTF8}{min} なんか (nanka)\end{CJK} &2 &2 &2 &5 &2 &2 &2 &4 &2 &2 &2 &2\\ 
    \begin{CJK}{UTF8}{min} あの (ano)\end{CJK} &2 &2 &2 &2 &2 &2 &2 &2 &2 &2 &2 &2\\ 
    \begin{CJK}{UTF8}{min} ん (n)\end{CJK} &2 &2 &2 &2 &2 &2 &2 &2 &2 &2 &2 &2\\ 
    \begin{CJK}{UTF8}{min} そうです (soudesu)\end{CJK} &2 &15 &2 &2 &2 &2 &2 &2 &2 &2 &2 &2\\ 
    \begin{CJK}{UTF8}{min} は (ha)\end{CJK} &2 &2 &4 &3 &2 &2 &2 &2 &2 &2 &2 &2\\ 
    \begin{CJK}{UTF8}{min} ね (ne)\end{CJK} &2 &2 &2 &2 &2 &2 &2 &2 &2 &2 &2 &2\\ 
    \begin{CJK}{UTF8}{min} いや (iya)\end{CJK} &2 &3 &2 &2 &2 &2 &3 &3 &3 &3 &2 &2\\ 
    \begin{CJK}{UTF8}{min} へー (he-)\end{CJK} &2 &7 &2 &2 &2 &2 &2 &2 &2 &2 &2 &2\\ 
    \begin{CJK}{UTF8}{min} そうか (souka)\end{CJK} &2 &2 &2 &2 &2 &5 &3 &2 &2 &2 &2 &2\\
\midrule
    \begin{CJK}{UTF8}{min} \bf Average \end{CJK} &\textbf{2.00} &\textbf{3.53} &\textbf{2.27} &\textbf{2.27} &\textbf{2.00} &\textbf{2.27} &\textbf{2.07} &\textbf{2.47} &\textbf{2.07} &\textbf{2.20} &\textbf{2.00} &\textbf{2.13} \\ 
\bottomrule
\end{tabularx}
\label{tab:jap_k2}
\end{table*}

\begin{table*}[ht]
\scriptsize
\centering
\caption{$k$-values of optimal results on the $k$-means clustering of the top 15 selected \textbf{Japanese} backchannels/fillers when using the \textbf{GPT-2} model in both the original dimensional space and after dimensionality reduction to 100 dimensions using PCA.}
    
\begin{tabularx}{\textwidth}{lXXXXXXXXXXXX}
\toprule
    \textbf{Backchannel/Filler} & \multicolumn{2}{c}{\textbf{no\_ft, no\_ctx}} & \multicolumn{2}{c}{\textbf{ft, no\_ctx (NTP)}} & \multicolumn{2}{c}{\textbf{ft, no\_ctx (TTP)}}   & \multicolumn{2}{c}{\textbf{no\_ft, one\_ctx}} & \multicolumn{2}{c}{\textbf{ft, one\_ctx (NTP)}} & \multicolumn{2}{c}{\textbf{ft, one\_ctx (TTP)}} \\
\cmidrule(lr){2-3} \cmidrule(lr){4-5} \cmidrule(lr){6-7} \cmidrule(lr){8-9} \cmidrule(lr){10-11} \cmidrule(lr){12-13}
     & \textbf{orig.} & \textbf{100} & \textbf{orig.} & \textbf{100} & \textbf{orig.} & \textbf{100} & \textbf{orig.} & \textbf{100} & \textbf{orig.} & \textbf{100} & \textbf{orig.} & \textbf{100} \\
\midrule
    \begin{CJK}{UTF8}{min} うん (un)\end{CJK} &2 &4 &2 &2 &2 &2 &2 &4 &4 &4 &2 &2\\ 
    \begin{CJK}{UTF8}{min} あ (a)\end{CJK} &2 &2 &3 &5 &2 &2 &2 &2 &3 &6 &2 &2\\ 
    \begin{CJK}{UTF8}{min} はい (hai)\end{CJK} &2 &2 &2 &2 &3 &2 &2 &2 &2 &2 &2 &2\\ 
    \begin{CJK}{UTF8}{min} え (e)\end{CJK} &2 &2 &6 &3 &2 &2 &2 &2 &3 &3 &2 &2\\ 
    \begin{CJK}{UTF8}{min} そう (sou)\end{CJK} &2 &2 &2 &3 &2 &2 &2 &4 &2 &3 &2 &2\\ 
    \begin{CJK}{UTF8}{min} ま (ma)\end{CJK} &2 &3 &2 &2 &2 &2 &2 &2 &3 &2 &2 &2\\
    \begin{CJK}{UTF8}{min} なんか (nanka)\end{CJK} &2 &2 &2 &2 &2 &2 &2 &2 &2 &2 &2 &2\\ 
    \begin{CJK}{UTF8}{min} あの (ano)\end{CJK} &2 &2 &2 &2 &2 &2 &2 &2 &2 &2 &2 &2\\ 
    \begin{CJK}{UTF8}{min} ん (n)\end{CJK} &2 &3 &2 &3 &2 &2 &2 &3 &2 &2 &2 &2\\ 
    \begin{CJK}{UTF8}{min} そうです (soudesu)\end{CJK} &3 &3 &2 &2 &3 &3 &2 &2 &5 &4 &2 &3\\ 
    \begin{CJK}{UTF8}{min} は (ha)\end{CJK} &2 &2 &2 &2 &3 &3 &2 &2 &2 &2 &2 &3\\ 
    \begin{CJK}{UTF8}{min} ね (ne)\end{CJK} &3 &2 &2 &2 &4 &3 &2 &2 &3 &3 &2 &3\\ 
    \begin{CJK}{UTF8}{min} いや (iya)\end{CJK} &2 &5 &6 &2 &6 &2 &2 &2 &2 &2 &2 &2\\ 
    \begin{CJK}{UTF8}{min} へー (he-)\end{CJK} &2 &3 &2 &2 &5 &2 &2 &2 &2 &2 &2 &3\\ 
    \begin{CJK}{UTF8}{min} そうか (souka)\end{CJK} &2 &2 &2 &6 &5 &11 &2 &2 &3 &3 &3 &3\\
\midrule
    \begin{CJK}{UTF8}{min} \bf Average \end{CJK} &\textbf{2.13} &\textbf{2.60} &\textbf{2.60} &\textbf{2.67} &\textbf{3.00} &\textbf{2.80} &\textbf{2.00} &\textbf{2.33} &\textbf{2.67} &\textbf{2.80} &\textbf{2.07} &\textbf{2.33} \\ 
\bottomrule
\end{tabularx}
\label{tab:jap_k3}
\end{table*}

\begin{table*}[ht]
\scriptsize
\caption{%
    $k$-values of optimal results on the $k$-means clustering of the top 15 selected \textbf{Japanese} backchannels/fillers when using the \textbf{BERT} model in both the original dimensional space and after dimensionality reduction to 100 dimensions using PCA.}
\begin{tabularx}{\textwidth}{lXXXXXXXX}
\toprule
    \textbf{Backchannel/Filler} & \multicolumn{2}{c}{\textbf{no\_ft, no\_ctx}} & \multicolumn{2}{c}{\textbf{ft, no\_ctx (MASK)}} & \multicolumn{2}{c}{\textbf{no\_ft, one\_ctx}} & \multicolumn{2}{c}{\textbf{ft, one\_ctx (MASK)}} \\
\cmidrule(lr){2-3} \cmidrule(lr){4-5} \cmidrule(lr){6-7} \cmidrule(lr){8-9}
     & \textbf{orig.} & \textbf{100} & \textbf{orig.} & \textbf{100} & \textbf{orig.} & \textbf{100} & \textbf{orig.} & \textbf{100} \\
\midrule
    \begin{CJK}{UTF8}{min} うん (un)\end{CJK} & 2 & 2 & 2 & 3 & 4 & 2 & 3 & 3 \\ 
    \begin{CJK}{UTF8}{min} あ (a)\end{CJK} & 3 & 4 & 4 & 2 & 3 & 4 & 4 & 2 \\ 
    \begin{CJK}{UTF8}{min} はい (hai)\end{CJK} & 2 & 2 & 2 & 4 & 2 & 2 & 2 & 2 \\ 
    \begin{CJK}{UTF8}{min} え (e)\end{CJK} & 4 & 4 & 2 & 3 & 5 & 6 & 2 & 3 \\ 
    \begin{CJK}{UTF8}{min} そう (sou)\end{CJK} & 2 & 2 & 2 & 2 & 2 & 4 & 2 & 3 \\ 
    \begin{CJK}{UTF8}{min} ま (ma)\end{CJK} & 4 & 4 & 2 & 2 & 4 & 4 & 2 & 3 \\
    \begin{CJK}{UTF8}{min} なんか (nanka)\end{CJK} & 2 & 2 & 2 & 2 & 2 & 2 & 2 & 2 \\ 
    \begin{CJK}{UTF8}{min} あの (ano)\end{CJK} & 2 & 2 & 2 & 2 & 3 & 3 & 2 & 2 \\ 
    \begin{CJK}{UTF8}{min} ん (n)\end{CJK} & 3 & 3 & 2 & 2 & 2 & 2 & 2 & 2 \\ 
    \begin{CJK}{UTF8}{min} そうです (soudesu)\end{CJK} & 2 & 2 & 2 & 3 & 2 & 2 & 3 & 4 \\ 
    \begin{CJK}{UTF8}{min} は (ha)\end{CJK} & 2 & 2 & 3 & 4 & 2 & 2 & 5 & 5 \\ 
    \begin{CJK}{UTF8}{min} ね (ne)\end{CJK} & 2 & 2 & 2 & 2 & 2 & 2 & 2 & 2 \\ 
    \begin{CJK}{UTF8}{min} いや (iya)\end{CJK} & 4 & 2 & 2 & 2 & 3 & 2 & 2 & 2 \\ 
    \begin{CJK}{UTF8}{min} へー (he-)\end{CJK} & 2 & 2 & 2 & 2 & 2 & 2 & 2 & 4 \\ 
    \begin{CJK}{UTF8}{min} そうか (souka)\end{CJK} & 3 & 3 & 2 & 2 & 2 & 2 & 2 & 2 \\
\midrule
    \begin{CJK}{UTF8}{min} \bf Average \end{CJK} & \textbf{2.60} & \textbf{2.53} & \textbf{2.20} & \textbf{2.47} & \textbf{2.66} & \textbf{2.73} & \textbf{2.46} & \textbf{2.73} \\
\bottomrule
\end{tabularx}
\label{tab:jap_k4}
\end{table*}


\end{document}